%% file: main.tex
\documentclass{article}

\usepackage{iclr2027_conference,times}
\input{math_commands.tex}

\usepackage{amsmath,amssymb}
\usepackage{booktabs}
\usepackage{calc}
\usepackage{graphicx}
\usepackage{longtable}
\usepackage{multirow}
\usepackage{tabularx}
\usepackage{xcolor}
\usepackage{fvextra}
\usepackage{mdframed}
\usepackage{tikz}
\usepackage{pgfplots}
\usepgfplotslibrary{groupplots}
\usetikzlibrary{positioning,arrows.meta}
\usepackage{hyperref}
\usepackage{url}
\usepackage{placeins}
\usepackage{pifont}
\usepackage{needspace}

\pgfplotsset{
  compat=1.17,
  every axis/.append style={font=\scriptsize, tick align=outside, tick pos=left,
    grid=major, grid style={gray!20}, line width=0.5pt},
  legend style={font=\scriptsize, draw=none, fill=none},
}

\graphicspath{{assets/}}
\newmdenv[backgroundcolor=gray!5,linecolor=gray,linewidth=0.4pt,
  innertopmargin=4pt,innerbottommargin=4pt,
  innerleftmargin=5pt,innerrightmargin=5pt,
  skipabove=5pt,skipbelow=5pt]{paperprompt}
\fvset{fontsize=\small,breaklines=true,breakindent=1.2em}
\makeatletter
\newenvironment{paperfigure}{%
  \par\medskip\noindent\begin{minipage}{\linewidth}\def\@captype{figure}\centering%
}{%
  \end{minipage}\par\medskip%
}
\newenvironment{papertable}{%
  \par\medskip\noindent\begin{minipage}{\linewidth}\def\@captype{table}\centering%
}{%
  \end{minipage}\par\medskip%
}
\makeatother
\input{figures/main_typography}

\title{TRIAD: Multimodal Retrieval across\\ Abstraction, Constraints, and State}

\author{%
\textbf{Xiangyang Wu\textsuperscript{1,2,3},
Chen Ju\textsuperscript{1,\ding{41}},
Xu Chen\textsuperscript{1},
Likang Peng\textsuperscript{1},}\\
\textbf{Jinsong Lan\textsuperscript{1},
Haihong Tang\textsuperscript{1},
Yang Hu\textsuperscript{2,3,\ding{41}}}\\[3pt]
\textsuperscript{1}Alibaba Group\\
\textsuperscript{2}Hangzhou International Innovation Institute, Beihang University\\
\textsuperscript{3}Sino-Italian Joint Research Laboratory on Prognostics and Health Management\\
for Intelligent Maintenance, Beihang University\\[3pt]
\href{mailto:wuxiangyang@buaa.edu.cn}{\{wuxiangyang}, \href{mailto:yang_hu@buaa.edu.cn}{yang\_hu\}}@buaa.edu.cn\quad
\href{mailto:cju.void@gmail.com}{cju.void@gmail.com}\\[4pt]
\begingroup
\setlength{\fboxsep}{3pt}%
\fcolorbox{black!55}{gray!8}{\href{https://github.com/simonmonmonn/TRIAD}{\textbf{Code}}}\quad
\fcolorbox{black!55}{gray!8}{\href{https://huggingface.co/datasets/sisisisis/TRIAD-data}{\textbf{Data}}}\quad
\fcolorbox{black!55}{gray!8}{\href{https://huggingface.co/spaces/sisisisis/PROBE-Leaderboard}{\textbf{Leaderboard}}}%
\endgroup
}

\iclrfinalcopy
\makeatletter
\def\@maketitle{%
  \vbox{\hsize\textwidth
    \centering
    {\LARGE\scshape \@title\par}%
    \vskip 0.12in
    {\normalsize \@author\par}%
    \vskip 0.30in minus 0.10in}}
\makeatother

\begin{document}
\maketitle
\fancyhead{}
\begingroup
\renewcommand{\thefootnote}{}
\footnotetext{\ding{41} Corresponding Authors.}
\addtocounter{footnote}{-1}
\endgroup

\begin{abstract}
\input{sections/en/00_abstract}
\end{abstract}

\input{sections/en/01_introduction}
\input{sections/en/02_related_work}
\input{sections/en/03_benchmark}
\input{sections/en/04_experimental_evaluation}
\input{sections/en/06_diagnostic_analysis}

\input{sections/en/07_conclusion}

\mainfiguretypographyfalse
\bibliography{references}
\bibliographystyle{iclr2027_conference}

\appendix
\input{figures/appendix_layout}
\input{sections/en/A_benchmark_construction}
\input{sections/en/B_evaluation_details}
\input{sections/en/C_diagnostic_details}

\input{sections/en/D_supplementary_experiments}
\input{sections/en/E_quality_audit}
\input{sections/en/F_benchmark_comparison}

\end{document}

%% file: math_commands.tex
\usepackage{amsmath,amsfonts,bm}

\def\eqref#1{equation~\ref{#1}}

\def\1{\bm{1}}

\DeclareMathAlphabet{\mathsfit}{\encodingdefault}{\sfdefault}{m}{sl}
\SetMathAlphabet{\mathsfit}{bold}{\encodingdefault}{\sfdefault}{bx}{n}



%% file: figures/main_typography.tex
\newcommand{\paperplotfont}{\normalfont\rmfamily\fontsize{8}{9.5}\selectfont}
\newcommand{\paperpanelfont}{\normalfont\rmfamily\fontsize{9}{10.5}\selectfont}
\pgfplotsset{paper main axis/.style={
  font=\paperplotfont,
  tick label style={font=\paperplotfont},
  label style={font=\paperplotfont},
  title style={font=\paperpanelfont},
  legend style={font=\paperplotfont,draw=none,fill=none},
}}
\newif\ifmainfiguretypography
\mainfiguretypographytrue
\makeatletter
\let\paper@originalmakecaption\@makecaption
\long\def\@makecaption#1#2{%
  \begingroup
  \def\paper@figuretype{figure}%
  \ifmainfiguretypography
    \ifx\@captype\paper@figuretype
      \normalfont\rmfamily\fontsize{10}{11}\selectfont
      \setlength{\abovecaptionskip}{5pt}%
      \setlength{\belowcaptionskip}{0pt}%
    \fi
  \fi
  \paper@originalmakecaption{#1}{#2}%
  \endgroup
}
\makeatother

%% file: sections/en/00_abstract.tex
Multimodal retrieval increasingly supports agentic systems, where user requests can involve cross-level relations, combine multiple constraints, and evolve across dialogue turns. In these settings, a highly similar result may still violate the user's requirements, while retrieved content may directly inform downstream actions without human screening. We introduce TRIAD, a multimodal retrieval benchmark for evaluating how well retrieval systems satisfy these complex, evolving requests. TRIAD comprises three tracks: abstraction, testing bidirectional retrieval between visual instances and concepts at different hierarchical levels; constraints, testing the joint satisfaction of positive and negative conditions; and state, testing the integration of requirements across dialogue turns and the updating of earlier conditions when users revise their requests. The benchmark contains six subsets, each with 1,000 queries and a fixed pool of 1,000 candidates per query, evaluated using a common multi-positive mean average precision (mAP) protocol. Simpler reference tasks and hard negatives grouped by the requirements they violate help characterize specific retrieval difficulties. We evaluate 20 configurations of 17 systems and find that no system leads across all six subsets. Performance declines with hierarchical distance, and negative-only queries generally underperform positive-only queries with the same number of constraints. Across paired tasks expressing the same final intent, multi-turn retrieval yields lower mAP in 39 of 40 configuration--domain comparisons. The benefits of explicit reasoning also vary across models and tasks. These findings highlight the gap between retrieving similar content and satisfying a complete request.

%% file: sections/en/01_introduction.tex
\section{Introduction}

Traditional multimodal search presents users with a ranked list of results in response to a text, image, or combined query. Embedding-based retrievers produce this ranking by mapping queries and candidates into a shared space and comparing their similarity, supported by advances in holistic image--text alignment and diversified multimodal embeddings~\citep{wang2025advancing,wang2025explore}. Its traditional use, in which users inspect and filter results ranked primarily by approximate similarity, rests on three premises. \emph{The query is self-contained}: the current input already expresses the full active intent, without requiring the system to determine which earlier requirements still apply. \emph{Overall similarity can stand in for relevance}: candidates are compared mainly by degree of match, without hard requirements whose violation cannot be compensated for by similarity in other respects. \emph{Retrieval ends with human judgment}: users can skip results that appear relevant but miss their actual needs.

In agentic settings, user requests become more detailed and complex and evolve through interaction, so these premises no longer generally hold. A request may seek concepts in a superordinate or subordinate relation, rather than simply similar content~\citep{vendrov2016order}. For example, when retrieving a superordinate concept for a sparrow, biologically similar bird species may resemble the sparrow more closely in appearance or fine-grained semantics, yet cannot substitute for a superordinate concept such as \emph{bird} or even \emph{animal}. A request may also impose multiple constraints that must hold jointly: ``like this scarf, but blue and not patterned'' requires both inclusion and exclusion, together with preservation of attributes that were not requested to change. Similarity in other respects cannot compensate for violating these requirements. A request may further span multiple turns, requiring selective retention or updating of historical state: when a user first asks for orange and later changes to blue, the system must retain conditions from the history that remain valid while replacing the abandoned value with the new one, rather than assume that the latest turn is self-contained. Meanwhile, retrieved results may feed directly into subsequent reasoning or tool calls, without a user screening each result for errors. Relevance must therefore be judged against the user's full current requirements, rather than overall similarity alone. Although existing benchmarks partially cover multi-constraint and multi-turn retrieval, a unified framework for evaluating these requirements and diagnosing specific model capability gaps is still lacking~\citep{wu2019fashioniq,alhamoud2025negbench,guo2018dialog}. Our concern is not to replace similarity-based ranking, but to test whether the relevance it represents responds to the complete request: we define correct results from the request's requirements, measure whether a model ranks them highly, and examine which requirements high-ranked errors violate.

\begin{figure}[t]
  \centering
  \setlength{\abovecaptionskip}{2pt}
  \input{figures/fig01_motivation_en}
  \caption{\textbf{Overview of TRIAD.} Cross-level, multi-constraint, and multi-turn retrieval test hierarchical relations, constraint composition, and intent updates. Green/red indicate satisfied/violated requirements. Dashed boxes show simpler diagnostic references.}
  \label{fig:motivation-en}
\end{figure}
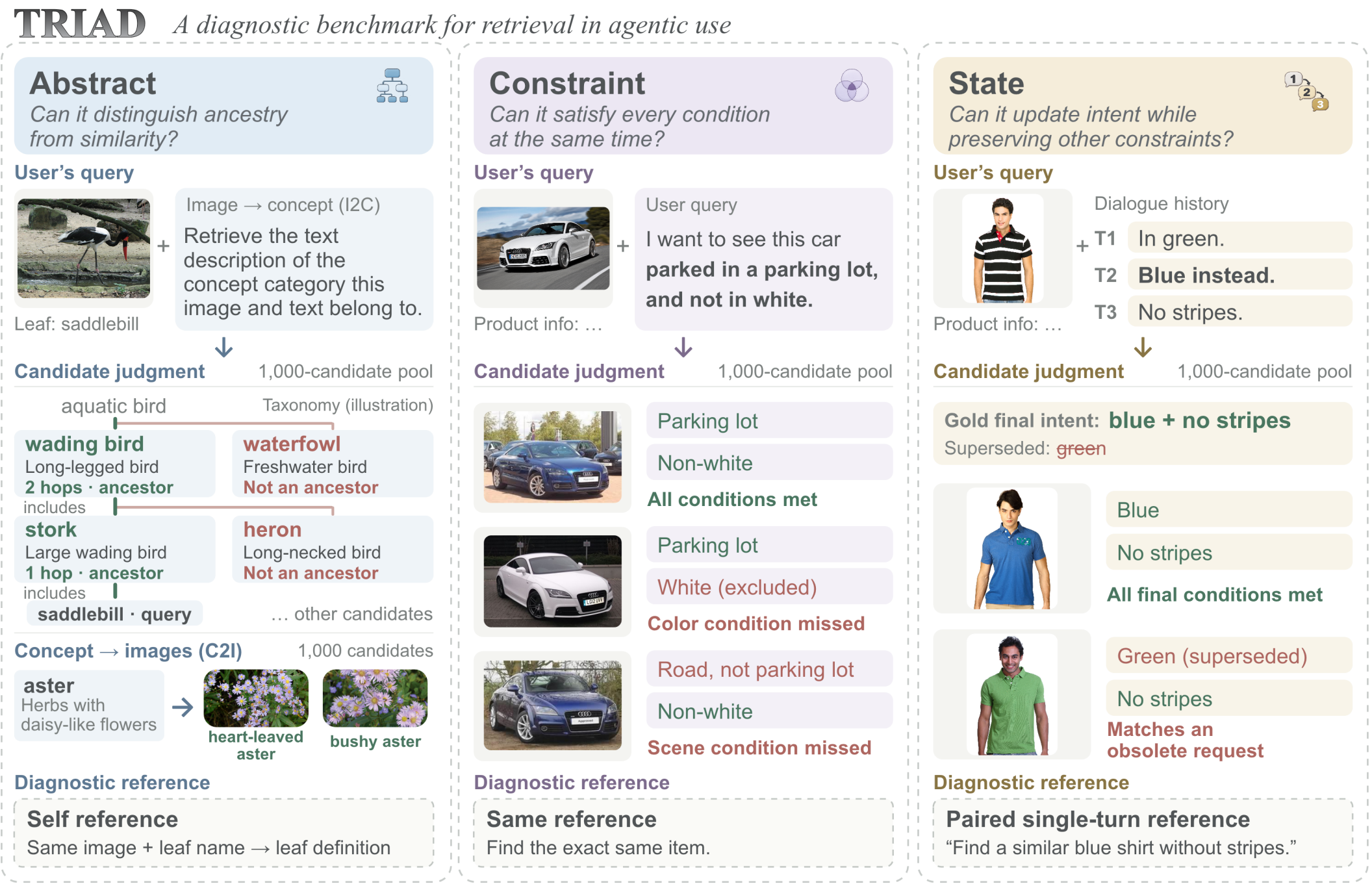

To this end, we introduce \textbf{TRIAD}, a multimodal retrieval benchmark designed to go beyond similarity matching and measure how well retrieval systems execute three types of specifications in user requests: cross-level retrieval, multiple constraints, and state updates (Figure~\ref{fig:motivation-en}). \textbf{\emph{The abstraction track}} evaluates bidirectional retrieval between visual instances and concepts at different levels, testing correct hierarchical relations beyond direct category matching. \textbf{\emph{The constraints track}} combines positive and negative requirements around a reference product, testing whether multiple conditions and attributes that should be preserved are jointly satisfied. \textbf{\emph{The state track}} distributes requirements across dialogue turns, testing integration of new conditions and retrieval according to the final intent after a revision. Each track also includes a lower-complexity reference and distractors grouped by the requirements they violate, enabling comparisons between basic matching and complex requests and analysis of high-ranked errors involving hierarchical confusion, constraint violations, or stale values. The abstraction track is split by retrieval direction, and the constraints and state tracks are each instantiated in fashion and cars, yielding six main subsets of 1,000 queries with 1,000 candidates per query. All subsets use a common multi-positive mAP protocol and are reported separately.

We evaluate 20 configurations of 17 systems to examine how representative embedding models from the retrieval literature perform on TRIAD. The results show that 1) instance-to-concept retrieval quality decreases with hierarchical distance; 2) query groups containing negative constraints generally score lower than positive-only groups, with candidates violating only negative requirements often becoming high-ranked distractors; and 3) multi-turn versions typically underperform their paired single-turn counterparts with the same final intent. Within-model comparisons with and without reasoning show no uniform gain across models, tasks, and integration methods.

%% file: figures/fig01_motivation_en.tex
\includegraphics[width=\linewidth]{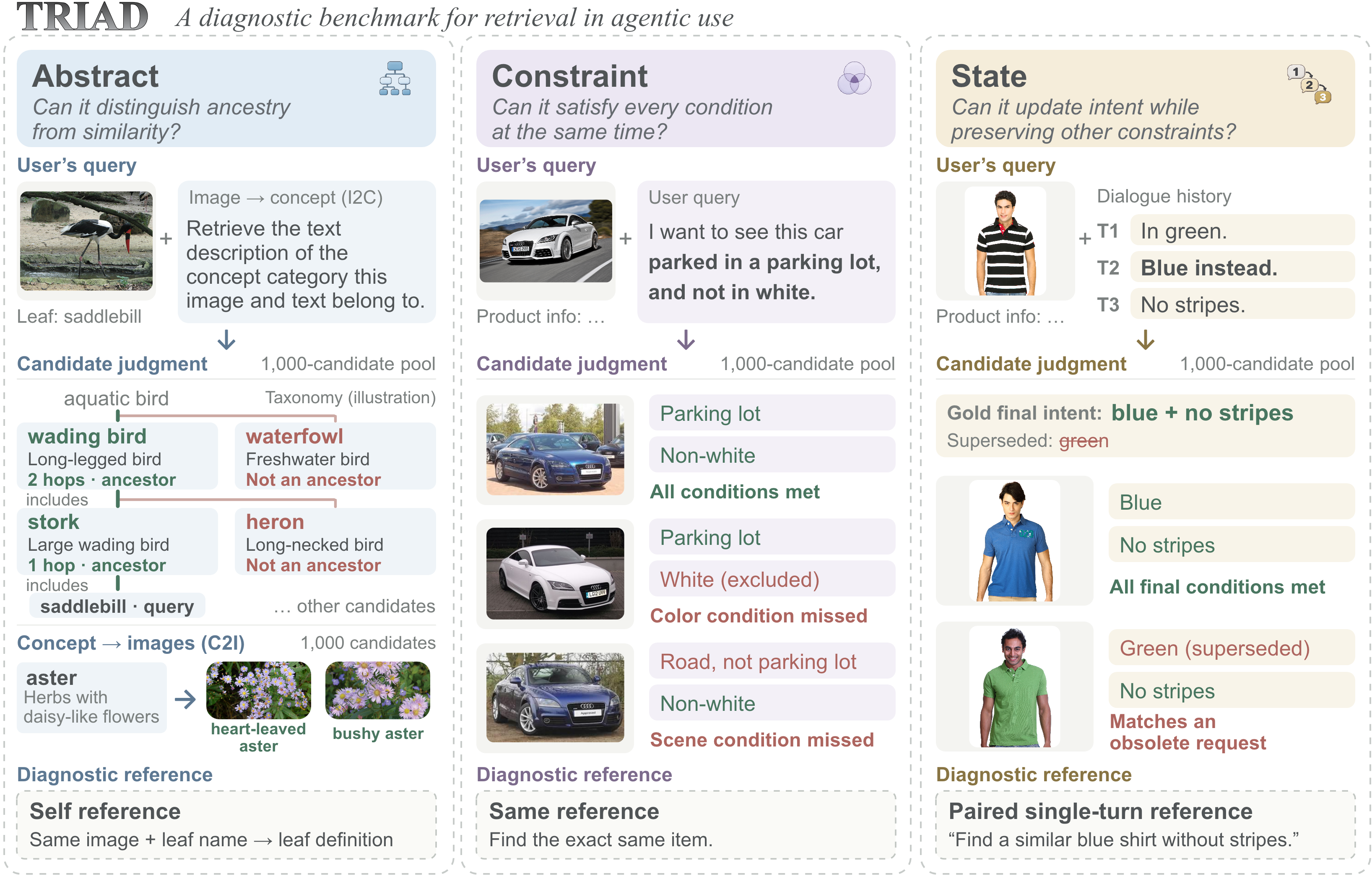}

%% file: sections/en/02_related_work.tex
\section{Related Work}
\label{sec:related-en}

\paragraph{Broad-Coverage and Instruction-Following Evaluation.}
Broad-coverage benchmarks bring together data across tasks, domains, and modalities to assess general capabilities across retrieval settings. BEIR and MTEB assemble heterogeneous tasks under a shared evaluation interface with per-task reporting~\citep{thakur2021beir,muennighoff2023mteb}. M-BEIR and the MMEB series extend unified evaluation to instruction-conditioned multimodal retrieval, including video, visual documents, and omni-modal settings~\citep{wei2024uniir,jiang2024vlm2vec,meng2025vlm2vecv2,huang2026mmebv3}; related representation work adapts image--language models to video and audio--visual learning~\citep{ju2022prompting,liu2022exploiting,liu2023audio,liu2024audio}. MMEB-V3 additionally introduces OmniSET for controlled diagnosis with semantic content aligned across modalities. FollowIR and InstructIR focus on instruction following in text retrieval, examining whether relevance judgments respond to the instruction~\citep{weller2025followir,oh2024instructir}. Representation work validated on such benchmarks benefits a range of downstream tasks~\citep{chen2025wave,cheng2023category,chen2024wear,ju2024turbo,wang2025folder}.

\paragraph{Focused Capability Evaluation.}
Specialized studies examine hierarchical relations, condition composition, or dialogue in greater depth. Hierarchy-aware methods use category or visual part--whole hierarchies to structure representations for classification and retrieval~\citep{vendrov2016order,barz2018hierarchy,bertinetto2020making,wang2025visualhierarchies}. Composed image retrieval specifies a target through a reference image and modification text~\citep{vo2019tirg,wu2019fashioniq,liu2021cirr}. Related evaluations examine multiple positives~\citep{baldrati2023circo}, conditional similarity~\citep{vaze2023genecis}, instance identity~\citep{psomas2025icir}, hard negatives~\citep{byun2025macir,mahadev2026pinpoint}, input robustness~\citep{sun2023benchcir}, and multi-turn interaction~\citep{liu2026sanitycheck}. NegBench specifically tests inclusion and exclusion requirements~\citep{alhamoud2025negbench}. Image--text compositionality benchmarks examine compositional generalization, attribute and relation understanding, and fine-grained image--text distinctions~\citep{ma2023crepe,yuksekgonul2023aro,hsieh2023sugarcrepe,thrush2022winoground}. Annotation-light settings organize visual concepts through prototypes, attribute decomposition, and unsupervised discovery~\citep{ma2023attrseg,yang2024multi,ma2023diffusionseg,ma2025freesegdiff}. Weak- and partial-supervision studies cut annotation cost across recognition and grounding tasks~\citep{ju2021divide,ju2022adaptive,ju2023distilling,ju2023multi,ju2023constraint,wang2025contrast,ju2026r}. Dialogue retrieval, retrieval-oriented rewriting, multi-turn RAG, and agentic evaluations study retrieval behavior over interaction histories~\citep{guo2018dialog,tan2019drilldown,zhao2024chatsearch,hu2026recqr,katsis2025mtrag,cheng2025coral,ali2026recor,you2026agenticragtracer}.

TRIAD complements these studies by bringing cross-level relations, joint constraints, and state updates into a unified multimodal retrieval evaluation framework. Disaggregated results, low-complexity references, and typed hard negatives reveal difficulties in following specific requirements. A dimension-by-dimension comparison with existing benchmarks appears in Appendix~\ref{app:compare-en}.

%% file: sections/en/03_benchmark.tex
\section{The TRIAD Benchmark}

\subsection{Three Capability Axes and Design Principles}
\label{sec:axes-en}

TRIAD defines relevance through the relations, conditions, and active intent specified in a request. Its abstraction, constraints, and state tracks evaluate cross-level retrieval, joint constraint satisfaction, and state updates, respectively. Four design principles help locate ranking difficulties: (P1) provide low-complexity references; (P2) screen potential false negatives using hierarchical or attribute rules; (P3) categorize distractors by the requirements they violate; and (P4) audit input shortcuts that could bypass the target capability.

\subsection{Tasks and Data}
\label{sec:tasks-en}

The three tracks yield six main subsets: I2C and C2I for abstraction, and Fashion and Cars for each of the constraints and state tracks. Each subset contains 1,000 queries, each with 1,000 fixed candidates and a labeled positive set. Figure~\ref{fig:motivation-en} illustrates the inputs and retrieval targets.

\subsubsection{Main-Task Data}
\label{sec:main-task-data-en}

\paragraph{Abstraction: Retrieval Across Levels.} Using the WordNet 3.0 noun IS-A hierarchy~\citep{miller1995wordnet} and ImageNet-21K images~\citep{deng2009imagenet}, we construct bidirectional cross-level retrieval between instances and concepts: Instance-to-Concept (I2C) and Concept-to-Instance (C2I). I2C asks for an image's ancestor concepts one or two levels above its leaf class; C2I asks for images from descendant classes one or two levels below a concept. Image inputs contain the image and leaf-class name; concept inputs contain the name and definition. Positives follow WordNet ancestor--descendant relations. The pools contain 1,317 concepts and 15,981 instance images.

\paragraph{Constraints: Multi-Condition Retrieval.} We retain 12,946 items from Fashion Product Images~\citep{aggarwal2019fashion} and 8,007 entries from Stanford Cars~\citep{krause2013cars}, constructing queries that seek similar items to a reference while satisfying one to three conditions. Queries include a reference image, product text, and requested changes; candidates include an image and product text. Positives must satisfy all explicit conditions and preserve attributes designated by the construction rules. Positive conditions (P) require an attribute, and negative conditions (N) exclude one; D1--D7 denote 1P, 1N, 2P, 2N, 1P1N, 2P1N, and 1P2N.

\paragraph{State: Multi-Turn Retrieval.} We rewrite constraints-track queries as two- or three-turn conversations, preserving the reference item, final intent, and complete positive set. Natural sessions progressively add conditions; conflict sessions introduce earlier values that are explicitly replaced later. Queries use the reference image and text with the full dialogue, while candidate inputs remain unchanged. Positives must satisfy the final active conditions, retaining unrevised requirements and discarding replaced values.

\subsubsection{Diagnostic and Reference Data}
\label{sec:diagnostic-data-en}

We use three references to distinguish basic matching from additional requirements: Self retrieves between images and their own leaf concepts in both directions; Same retrieves other images of the same item from an image--text reference; and the paired single turn expresses the complete final intent of a multi-turn query at once, rather than retaining only its last turn. These references help locate difficulties, but their candidate composition need not match the main tasks, so they do not provide a common upper bound.

Hard negatives are typed by the violated requirements. Abstraction uses semantically related candidates with incorrect hierarchical relations. Constraints distinguishes violations of positive conditions, negative conditions, preserved attributes, and both positive and negative conditions, alongside generic visually similar negatives. State introduces revision traps that satisfy an obsolete value but violate the final conditions. Complete construction procedures and statistics appear in Appendix~\ref{app:benchmark-construction-en}.

\subsection{Data Quality and Validity}
\label{sec:data-quality-en}

Complementary to finer-grained data-governance methods~\citep{lin2026squeeze}, we screen benchmark data through image filtering, attribute normalization, and checks of conditions and final intent. Fashion uses two-model agreement on visual attributes; Cars uses single-model annotation with programmatic checks (Appendix~\ref{app:quality-assurance-en}). Human sample-based reviews cover 80 queries or sessions and 80 images, distinguishing query issues by their nature and impact on understanding the requirements. Among the sampled images, 507 of 513 determinate visual attributes are annotated correctly, with eight additional attributes marked not visible/undetermined, yielding 98.83\% accuracy. Query-review and attribute-verification results appear in Appendix~\ref{app:e6a-audit-en}. Some models were trained on source datasets or related category knowledge, so we do not claim complete decontamination; per-system exposure audits appear in Appendix~\ref{app:reliability-checks-en}.

\input{sections/en/03_design_comparison}

\subsection{Evaluation Protocol and Metrics}
\label{sec:protocol-en}

We L2-normalize query and candidate vectors and rank by cosine similarity. Across the six main subsets, 65.3\%--100.0\% of queries have multiple positives; MRR and Hit@K consider only the first positive or whether any is retrieved. We use mAP, averaging query AP equally over full candidate rankings with the complete labeled-positive count as denominator. Metric definitions appear in Appendix~\ref{app:metric-definitions-en}; supplementary Top-1 labeled-positive hit rates appear in Appendix~\ref{app:main-results-details-en}.

%% file: sections/en/03_design_comparison.tex
\subsection{Design Differences from Existing Benchmarks}
\label{sec:bench-design-en}

Table~\ref{tab:bench-design-en} summarizes the design differences between TRIAD and three representative benchmarks. MMEB-V2 integrates heterogeneous tasks under a unified evaluation protocol, but score differences across tasks alone do not directly identify specific capability gaps~\citep{meng2025vlm2vecv2}; TRIAD connects ranking performance to specific difficulties through reference tasks and hard negatives grouped by the requirements they violate. NegBench primarily evaluates negation understanding through the presence or absence of particular semantic entities~\citep{alhamoud2025negbench}, whereas TRIAD considers entities with multiple attributes, jointly organizing positive attribute requirements, negative exclusions, and implicit preservation constraints from a reference entity to test whether models satisfy requirements across multiple dimensions. ChatSearch focuses on recovering retrieval intent from multi-turn image--text contexts~\citep{zhao2024chatsearch}; its publicly documented construction does not treat multidimensional constraints distributed across turns, replacement of earlier conditions, and trap interference as explicitly controlled evaluation factors. TRIAD distributes known constraints across turns, explicitly constructs condition accumulation and the replacement of old values by new requirements, and introduces trap candidates that satisfy obsolete conditions but violate the final request, to diagnose whether models retain valid conditions, update superseded requirements, and exclude interference from earlier states.

\begin{table}[!htbp]
\small\setlength{\tabcolsep}{3pt}\renewcommand{\arraystretch}{1.00}
\caption{Design comparison with representative benchmarks. Extended criteria and comparisons appear in Appendix~\ref{app:bench-comparison-en}.}
\label{tab:bench-design-en}
\begin{center}
\begin{tabularx}{\linewidth}{@{}>{\raggedright\arraybackslash}p{43pt}>{\raggedright\arraybackslash}p{70pt}>{\raggedright\arraybackslash}p{84pt}>{\raggedright\arraybackslash}p{70pt}>{\raggedright\arraybackslash}X@{}}
\toprule
\textbf{Dimension} & \textbf{MMEB-V2} & \textbf{NegBench} & \textbf{ChatSearch} & \textbf{TRIAD (ours)} \\
\midrule
Targets
& Classification, QA, retrieval
& Object presence / absence
& Dialogue-specified images
& Instances $\leftrightarrow$ higher/lower concepts; products \\
\addlinespace[3pt]
Conditions
& Task-specific; composed retrieval
& Positive, negative, hybrid
& Descriptions and feedback
& 1--3 conditions; 7 polarity mixes; preserved attributes \\
\addlinespace[3pt]
State
& Context tasks (e.g., VisDial)
& Single-turn
& Single/multi-turn; text/image--text
& Accumulate/replace; matched single turn \\
\addlinespace[3pt]
Diagnostics
& Task-specific
& Negation distractors; presence/absence pairs
& Context-format comparisons
& Typed violations and traps; Self/Same/single-turn \\
\bottomrule
\end{tabularx}
\end{center}
\end{table}

%% file: sections/en/04_experimental_evaluation.tex
\section{Experimental Evaluation}
\label{sec:results-en}
\label{sec:experimental-evaluation-en}

We compare existing embedding models on the six subsets, using cross-benchmark rankings and a reranking reference to examine performance and headroom under specific retrieval requirements; unified embedding--reranking systems are also increasingly used in real-time industrial multimodal search~\citep{chen2026pailitao}.

\subsection{Models and Experimental Setup}
\label{sec:setup-en}

We evaluate 20 configurations of 17 independent systems spanning dual-tower encoders, discriminative embedders, and reasoning embedders (Table~\ref{tab:main-results-en}). Configurations include different sizes within a series, UME-R1 Disc/Gen, and Embed-RL-2B/4B with and without CoT to examine scaling and reasoning effects. All models are evaluated zero-shot without fine-tuning on TRIAD.
Models use fixed per-query candidate sets and checkpoint-specific input adaptations. We report mAP separately on the six main subsets, without an aggregate score. Encoding, fusion, and reasoning configurations appear in Appendix~\ref{app:evaluation-details-en}; confidence intervals and statistical tests for paired comparisons appear in Appendices~\ref{app:main-results-details-en} and~\ref{app:supplementary-experiments-en}. We additionally use Qwen3-VL-Reranker-8B to rerank candidates from Qwen3-VL-Embedding-8B as a two-stage retrieval reference (Section~\ref{sec:ceiling-en}).

\subsection{Overall Results and Cross-Task Performance}
\label{sec:overall-en}
\label{sec:families-en}

Table~\ref{tab:main-results-en} summarizes the six main subsets; complete Self and Same reference-task results appear in Appendix~\ref{app:self-same-results-en}. WeMM-Embedding-9B achieves the highest point estimates on C2I and both domains of the constraints and state tracks, while RzenEmbed-v2 leads I2C. The two leading C2I scores differ by only 0.001 mAP.

\begin{table}[!t]
\caption{mAP on six main subsets (higher is better). Bold/underlined values mark the best/second-best embedding configurations before rounding. The final reranking reference is excluded from these marks and family comparisons (Section~\ref{sec:ceiling-en}).}
\label{tab:main-results-en}
\footnotesize
\setlength{\tabcolsep}{3pt}
\begin{center}
\begin{tabularx}{\linewidth}{@{}>{\raggedright\arraybackslash}Xrrrrrr@{}}
\toprule
& \multicolumn{2}{c}{Abstraction} & \multicolumn{2}{c}{Constraints} & \multicolumn{2}{c}{State} \\
\cmidrule(lr){2-3}\cmidrule(lr){4-5}\cmidrule(l){6-7}
Model & I2C & C2I & Fashion & Cars & Fashion & Cars \\
\midrule
\multicolumn{7}{@{}l}{\textit{Dual-tower encoders}} \\
CLIP ViT-L/14~\citep{radford2021clip} & 0.318 & 0.224 & 0.185 & 0.065 & 0.180 & 0.062 \\
SigLIP-2-so400m~\citep{tschannen2025siglip2multilingualvisionlanguage} & 0.232 & 0.278 & 0.173 & 0.082 & 0.144 & 0.069 \\
jina-clip-v2~\citep{koukounas2024jinaclipv2multilingualmultimodalembeddings} & 0.354 & 0.401 & 0.123 & 0.052 & 0.117 & 0.050 \\
\midrule
\multicolumn{7}{@{}l}{\textit{Discriminative embedders}} \\
WeMM-Embedding-9B~\citep{zhou2026wemm} & 0.524 & \textbf{0.638} & \textbf{0.547} & \textbf{0.367} & \textbf{0.498} & \textbf{0.291} \\
E5-Omni-7B~\citep{chen2026e5omni} & 0.333 & 0.557 & 0.364 & 0.186 & 0.223 & 0.101 \\
RzenEmbed-v2~\citep{jian2025rzenembed} & \textbf{0.569} & 0.610 & 0.482 & \underline{0.298} & 0.441 & \underline{0.272} \\
Qwen3-VL-Emb.-8B~\citep{qwen3vlembedding} & 0.494 & \underline{0.637} & \underline{0.508} & 0.230 & \underline{0.454} & 0.177 \\
Qwen3-VL-Emb.-2B~\citep{qwen3vlembedding} & 0.463 & 0.507 & 0.442 & 0.157 & 0.212 & 0.085 \\
OmniEmbed-3B~\citep{xu2025omni} & 0.384 & 0.496 & 0.184 & 0.084 & 0.167 & 0.076 \\
VLM2Vec-V2.0-2B~\citep{meng2025vlm2vecv2} & 0.333 & 0.370 & 0.318 & 0.182 & 0.265 & 0.142 \\
VLM2Vec-2B~\citep{jiang2024vlm2vec} & 0.306 & 0.348 & 0.213 & 0.101 & 0.191 & 0.092 \\
VLM2Vec-7B~\citep{jiang2024vlm2vec} & 0.403 & 0.392 & 0.251 & 0.128 & 0.221 & 0.107 \\
Embed-RL-2B (no CoT)~\citep{jiang2026embed} & 0.448 & 0.465 & 0.355 & 0.198 & 0.308 & 0.170 \\
Embed-RL-4B (no CoT)~\citep{jiang2026embed} & 0.518 & 0.527 & 0.356 & 0.227 & 0.321 & 0.194 \\
B-Tokens-2B~\citep{sun2026btoks} & 0.453 & 0.467 & 0.368 & 0.169 & 0.300 & 0.146 \\
UME-R1-7B (Disc)~\citep{lan2025ume} & 0.538 & 0.577 & 0.436 & 0.249 & 0.321 & 0.195 \\
\midrule
\multicolumn{7}{@{}l}{\textit{Reasoning embedders}} \\
UME-R1-7B (Gen)~\citep{lan2025ume} & 0.537 & 0.589 & 0.359 & 0.174 & 0.235 & 0.130 \\
TWN-4B~\citep{zhang2026twn} & 0.527 & 0.521 & 0.264 & 0.140 & 0.201 & 0.092 \\
Embed-RL-2B + T-CoT~\citep{jiang2026embed} & 0.504 & 0.519 & 0.403 & 0.221 & 0.374 & 0.195 \\
Embed-RL-4B + T-CoT~\citep{jiang2026embed} & \underline{0.538} & 0.577 & 0.360 & 0.281 & 0.376 & 0.259 \\
\midrule
Top-30 reranking (ceiling probe)~\citep{qwen3vlembedding} & 0.607 & 0.674 & 0.566 & 0.327 & 0.516 & 0.258 \\
\bottomrule
\end{tabularx}
\end{center}
\end{table}

\begin{figure}[!t]
\centering
\includegraphics[width=\linewidth,trim=0 14bp 0 3bp,clip]{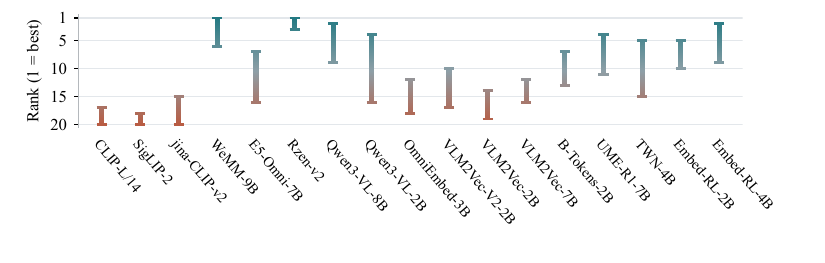}
\caption{Rank ranges across six subsets; vertical segments connect the best and worst ranks, with lower being better. The 17 displayed configurations follow main-table order, using Gen for UME-R1 and +T-CoT for Embed-RL. Ranks use scores before display rounding among all 20 configurations, with average ranks for ties.}
\label{fig:profiles-en}
\end{figure}

Relative strengths vary with task and domain: the median span between a configuration's best and worst ranks across the six subsets is 5.5. Figure~\ref{fig:profiles-en} illustrates why a single overall ranking cannot capture adaptation to different retrieval requirements; complete rank-agreement analyses appear in Appendix~\ref{app:main-results-details-en}.

Under the current input adaptations, the best dual-tower score is below the best discriminative-model score on every subset. Within the Qwen3-VL-Embedding series, 8B outperforms 2B on all six subsets. Across series or generations, however, parameter count does not directly predict performance: VLM2Vec-V2.0-2B exceeds VLM2Vec-7B on constraints and state but falls behind on abstraction. Paired reasoning analyses appear in Section~\ref{sec:reasoning-en}.

\input{sections/en/04_external_ranking}

\subsection{Reranking Reference and Performance Headroom}
\label{sec:ceiling-en}

We use Qwen3-VL-Reranker-8B to jointly score and reorder the Top-30 returned by Qwen3-VL-Embedding-8B, leaving the remaining candidates in place and computing mAP over the full ranking. Relative to the first-stage baseline from the same run, all six subsets improve by 0.037--0.113 mAP. Reranking exceeds the best embedding configuration on four subsets, while WeMM still leads both Cars subsets. Joint scoring improves ranking, but reaches only 41\%--77\% of oracle@30, which ideally places positives first within the retrieved Top-30. This reference reveals remaining headroom within the fixed retrieved set; positives outside that set still require a better first stage. Complete results and statistics appear in Appendix~\ref{app:e5-reranking-en}; Appendix~\ref{app:e6b-ratings-en} additionally reports a small-scale comparison of human relevance ratings and reranker scores on identical ten-candidate sets.

%% file: sections/en/04_external_ranking.tex
\subsection{Ranking Comparison with MMEB-V2}
\label{sec:external-ranking-en}

We compare the ranks of the 13 systems shared between the main table and MMEB-V2 on its Image subset and on the six TRIAD subsets (Table~\ref{tab:bench-ranks-en}). The leading order on general evaluation does not fully carry over: RzenEmbed-v2 rises from third on MMEB-V2 Image to first on I2C; UME-R1 (Gen) rises from seventh to third, and Embed-RL-4B/2B (+T-CoT) rise from eighth/ninth to second/fifth. Reasoning embedders' relative strength on I2C may reflect richer query information from reasoning, helping alignment with more distant ancestor concepts.

\begin{table}[!htbp]
\footnotesize\setlength{\tabcolsep}{2pt}\renewcommand{\arraystretch}{1.00}
\caption{Rank comparison between MMEB-V2 Image and TRIAD (selected systems). Ranks are computed among the 13 systems in Table~\ref{tab:main-results-en} shared with MMEB-V2; parentheses show positions gained or lost relative to MMEB-V2, and --- denotes no change.}
\label{tab:bench-ranks-en}
\begin{center}
\begin{tabular*}{\linewidth}{@{}l@{\extracolsep{\fill}}rrrrrrr@{}}
\toprule
Model & MMEB-V2 Image & I2C & C2I & B-Fashion & B-Cars & C-Fashion & C-Cars \\
\midrule
WeMM-9B & 1 & \textcolor[HTML]{B85C43}{4 ($\downarrow$3)} & \textcolor[HTML]{69747D}{1 (---)} & \textcolor[HTML]{69747D}{1 (---)} & \textcolor[HTML]{69747D}{1 (---)} & \textcolor[HTML]{69747D}{1 (---)} & \textcolor[HTML]{69747D}{1 (---)} \\
Qwen3-VL-8B & 2 & \textcolor[HTML]{B85C43}{6 ($\downarrow$4)} & \textcolor[HTML]{69747D}{2 (---)} & \textcolor[HTML]{69747D}{2 (---)} & \textcolor[HTML]{B85C43}{4 ($\downarrow$2)} & \textcolor[HTML]{69747D}{2 (---)} & \textcolor[HTML]{B85C43}{5 ($\downarrow$3)} \\
Rzen-v2 & 3 & \textcolor[HTML]{247A83}{1 ($\uparrow$2)} & \textcolor[HTML]{69747D}{3 (---)} & \textcolor[HTML]{69747D}{3 (---)} & \textcolor[HTML]{247A83}{2 ($\uparrow$1)} & \textcolor[HTML]{69747D}{3 (---)} & \textcolor[HTML]{247A83}{2 ($\uparrow$1)} \\
Qwen3-VL-2B & 4 & \textcolor[HTML]{B85C43}{7 ($\downarrow$3)} & \textcolor[HTML]{B85C43}{8 ($\downarrow$4)} & \textcolor[HTML]{69747D}{4 (---)} & \textcolor[HTML]{B85C43}{10 ($\downarrow$6)} & \textcolor[HTML]{B85C43}{11 ($\downarrow$7)} & \textcolor[HTML]{B85C43}{12 ($\downarrow$8)} \\
B-Tokens-2B & 5 & \textcolor[HTML]{B85C43}{8 ($\downarrow$3)} & \textcolor[HTML]{B85C43}{10 ($\downarrow$5)} & \textcolor[HTML]{B85C43}{6 ($\downarrow$1)} & \textcolor[HTML]{B85C43}{9 ($\downarrow$4)} & \textcolor[HTML]{B85C43}{6 ($\downarrow$1)} & \textcolor[HTML]{B85C43}{6 ($\downarrow$1)} \\
\bottomrule
\end{tabular*}
\end{center}
\end{table}

Multi-turn tasks also change relative order: Qwen3-VL-Embedding-2B leads B-Tokens-2B on MMEB-V2 Image and Fashion constraints but falls behind on Fashion state. Strength on general image tasks therefore cannot substitute for evaluation of cross-level and multi-turn capabilities. Complete comparisons appear in Appendix~\ref{app:bench-comparison-en}.

%% file: sections/en/06_diagnostic_analysis.tex
\raggedbottom
\section{Diagnostic Analysis}

\label{sec:diagnosis-en}

Chapter~\ref{sec:results-en} reports model performance across retrieval demands, but aggregate scores alone do not explain which requirements are harder to satisfy. We combine low-complexity references, typed negatives, and input ablations to examine difficulties in cross-level relations, joint constraints, and multi-turn state, then assess the effect of adding reasoning within the same checkpoint. All analyses follow the scoring protocol in Section~\ref{sec:protocol-en}; qualitative examples from all three tracks appear in Appendix~\ref{app:qualitative-cases-en}.

\subsection{Distance and Direction in Cross-Level Retrieval}
\label{sec:hierarchy-en}

\begin{figure}[!t]
  \centering
  \includegraphics[width=\linewidth,trim=0 6bp 0 7bp,clip]{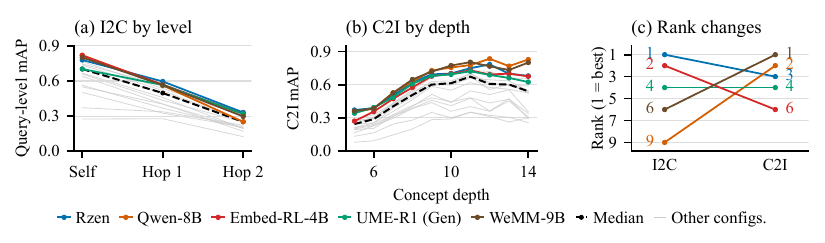}
  \caption{\textbf{Hierarchical distance and retrieval direction.} (a) mAP for Self and ancestors one and two levels away; (b) C2I mAP by concept depth; (c) I2C/C2I ranks of five representative configurations. Panels (a,b) show 17 configurations; the black dashed line is the median. Configuration selection and ranking follow Figure~\ref{fig:profiles-en}.}
  \label{fig:hierarchy-en}
\end{figure}

Cross-level retrieval becomes harder at greater hierarchical distances. Median mAP across 17 configurations is 0.702 for Self, 0.497 for one-level ancestors, and 0.251 for two-level ancestors (Figure~\ref{fig:hierarchy-en}a). On shared queries, two-level retrieval remains worse in all 17 configurations after removing labeled positives at the other distance. Thus, the distance gap is not solely an occupancy effect (Appendix~\ref{app:hierarchy-scoring-sensitivity-en}).

Retrieval direction also changes models' relative strengths. C2I performance varies with concept depth (Figure~\ref{fig:hierarchy-en}b), while leading I2C systems do not necessarily lead C2I: Rzen moves from rank 1 to 3, whereas WeMM rises from rank 6 to 1 (Figure~\ref{fig:hierarchy-en}c). Retrieving ancestor concepts from instances and retrieving instances from concepts thus place different demands on models.

\subsection{Joint Constraints and Negation}
\label{sec:constraints-en}

\begin{figure}[!t]
  \centering
  \includegraphics[width=\linewidth,trim=0 6bp 0 7bp,clip]{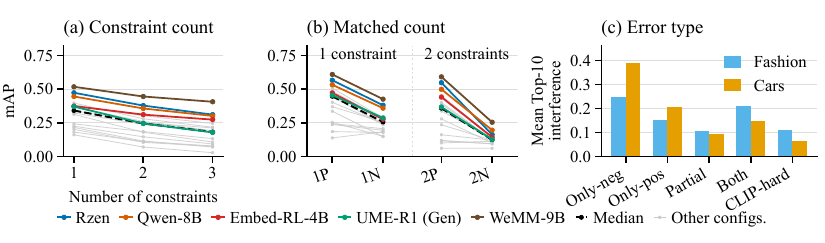}
  \caption{(a) Query-weighted mAP by constraint count; (b) matched-count positive-only and negative-only queries (P/N: positive/negative conditions). Panels (a,b) average Fashion and Cars equally per configuration; dashed black lines mark configuration medians. (c) Top-10 entry rates by negative type, normalized by that type's candidate count and averaged across configurations. The 17 configurations follow Figure~\ref{fig:profiles-en}.}
  \label{fig:constraints-en}
\end{figure}

Averaging mAP equally over Fashion and Cars, all 17 configurations decline as constraints increase from one to three (Figure~\ref{fig:constraints-en}a). With counts matched, negative-only queries score lower for 16 of 17 configurations in both the 1P--1N and 2P--2N comparisons (Figure~\ref{fig:constraints-en}b). The difficulty thus persists at matched counts; domain-specific and category/attribute-standardized results appear in Appendix~\ref{app:track-b-statistics-en}.

Typed negatives reveal how this difficulty manifests in ranking: candidates satisfying all positive conditions but violating one negative requirement have the highest Top-10 entry rate in 30 of 34 configuration--domain combinations (Figure~\ref{fig:constraints-en}c). Models still tend to prioritize candidates that resemble the reference but satisfy only part of the request.

In modality ablations, both text-only and image-only inputs underperform the full input, indicating the importance of combining both modalities to satisfy joint requirements; complete results appear in Appendix~\ref{app:modal-ablation-en}.

\subsection{Multi-Turn Information Integration and State Updates}
\label{sec:dialogue-en}
\label{sec:dialogue-cost-en}

We compare single- and multi-turn requests with the same reference item and final intent to examine performance when requirements are distributed across a conversation. mAP decreases in 39 of the 40 comparisons spanning 20 configurations and two domains (Table~\ref{tab:dialogue-selected-en}). The overall degradation persists when single-turn requests are rescored on the same state-task candidate sets (Appendix~\ref{app:state-paired-changes-en}). A breakdown by session type shows that degradation is concentrated in revision sessions, while multi-turn expression without revisions has a smaller overall effect, suggesting that updating superseded conditions is a major difficulty in multi-turn retrieval. Changes for revision and non-revision sessions appear in Table~\ref{tab:session-turn-en} of Appendix~\ref{app:state-paired-changes-en}.

\begin{table}[!t]
\small\setlength{\tabcolsep}{4pt}\renewcommand{\arraystretch}{1.00}
\caption{Single-to-multi-turn mAP changes for selected configurations. $\Delta$ mAP is multi-turn minus single-turn, computed from unrounded means and rounded to three decimals. UME-R1 uses Gen and Embed-RL uses +T-CoT; all 20 configurations appear in Appendix~\ref{app:state-paired-changes-en}.}
\label{tab:dialogue-selected-en}
\begin{center}
\input{tables/tab_dialogue_selected_en.tex}
\end{center}
\end{table}

Keeping only the last turn does not resolve this difficulty: last-turn-only inputs underperform full history in all 22 configuration--domain comparisons, showing that earlier conditions remain important for recovering the current request. Yet obsolete conditions in revision sessions can also interfere with ranking. Multi-turn retrieval therefore requires not just using history, but retaining valid conditions and updating superseded values. History-truncation and revision analyses appear in Appendix~\ref{app:history-revisions-en}.

\begin{table}[!t]
\small\setlength{\tabcolsep}{4pt}\renewcommand{\arraystretch}{1.00}
\caption{mAP before and after reasoning. Each pair reports the non-reasoning baseline followed by $\Delta$ mAP (reasoning minus non-reasoning). Full results and tests appear in Appendix~\ref{app:reasoning-results-en}.}
\label{tab:reasoning-changes-en}
\begin{center}
\input{tables/tab_reasoning_changes_en.tex}
\end{center}
\end{table}

\subsection{Effects and Limitations of Reasoning Configurations}

\label{sec:reasoning-en}

Can explicit reasoning improve retrieval? We compare two pairs with the encoder weights held fixed: UME-R1 switches query encoding from Disc to Gen while leaving the candidate side unchanged, whereas Embed-RL-4B adds external T-CoT on both the query and candidate sides (Table~\ref{tab:reasoning-changes-en}).

The evaluated UME-R1 Gen configuration does not yield consistent improvements: the mean I2C difference is near zero, C2I improves by 0.011 mAP, and all four constraints- and state-track subsets decline.
Two-sided T-CoT shows a different overall trend. Embed-RL-4B improves significantly on five subsets; its mean difference on the Fashion constraints task is +0.004, with an interval crossing zero.

%% file: tables/tab_dialogue_selected_en.tex
\begin{tabularx}{\linewidth}{@{}Xrrrrrr@{}}
\toprule
Configuration & \multicolumn{3}{c}{Fashion} & \multicolumn{3}{c}{Cars} \\
\cmidrule(lr){2-4}\cmidrule(lr){5-7}
 & Single & Multi & $\Delta$ mAP & Single & Multi & $\Delta$ mAP \\
\midrule
WeMM-Embedding-9B & 0.547 & 0.498 & \textcolor[HTML]{B85C43}{$-0.049$} & 0.367 & 0.291 & \textcolor[HTML]{B85C43}{$-0.076$} \\
RzenEmbed-v2 & 0.482 & 0.441 & \textcolor[HTML]{B85C43}{$-0.041$} & 0.298 & 0.272 & \textcolor[HTML]{B85C43}{$-0.026$} \\
Qwen3-VL-Embedding-8B & 0.508 & 0.454 & \textcolor[HTML]{B85C43}{$-0.054$} & 0.230 & 0.177 & \textcolor[HTML]{B85C43}{$-0.053$} \\
Qwen3-VL-Embedding-2B & 0.442 & 0.212 & \textcolor[HTML]{B85C43}{$-0.230$} & 0.157 & 0.085 & \textcolor[HTML]{B85C43}{$-0.072$} \\
Embed-RL-2B + T-CoT & 0.403 & 0.374 & \textcolor[HTML]{B85C43}{$-0.030$} & 0.221 & 0.195 & \textcolor[HTML]{B85C43}{$-0.026$} \\
Embed-RL-4B + T-CoT & 0.360 & 0.376 & \textcolor[HTML]{247A83}{$+0.016$} & 0.281 & 0.259 & \textcolor[HTML]{B85C43}{$-0.022$} \\
UME-R1-7B (Gen) & 0.359 & 0.235 & \textcolor[HTML]{B85C43}{$-0.124$} & 0.174 & 0.130 & \textcolor[HTML]{B85C43}{$-0.045$} \\
\bottomrule
\end{tabularx}

%% file: tables/tab_reasoning_changes_en.tex
\begin{tabularx}{\linewidth}{@{}Xrrrrrr@{}}
\toprule
Model & \multicolumn{2}{c}{Abstraction} & \multicolumn{2}{c}{Constraints} & \multicolumn{2}{c}{State} \\
\cmidrule(lr){2-3}\cmidrule(lr){4-5}\cmidrule(lr){6-7}
 & I2C & C2I & Fashion & Cars & Fashion & Cars \\
\midrule
UME-R1-7B (Disc) & 0.538 & 0.577 & 0.436 & 0.249 & 0.321 & 0.195 \\
\quad $\Delta$ mAP & \textcolor[HTML]{B85C43}{$-0.001$} & \textcolor[HTML]{247A83}{$+0.011$} & \textcolor[HTML]{B85C43}{$-0.077$} & \textcolor[HTML]{B85C43}{$-0.075$} & \textcolor[HTML]{B85C43}{$-0.086$} & \textcolor[HTML]{B85C43}{$-0.066$} \\
\addlinespace[3pt]
Embed-RL-4B (no CoT) & 0.518 & 0.527 & 0.356 & 0.227 & 0.321 & 0.194 \\
\quad $\Delta$ mAP & \textcolor[HTML]{247A83}{$+0.020$} & \textcolor[HTML]{247A83}{$+0.049$} & \textcolor[HTML]{247A83}{$+0.004$} & \textcolor[HTML]{247A83}{$+0.055$} & \textcolor[HTML]{247A83}{$+0.055$} & \textcolor[HTML]{247A83}{$+0.065$} \\
\bottomrule
\end{tabularx}

%% file: sections/en/07_conclusion.tex
\section{Conclusion}

We introduce TRIAD, a multimodal retrieval benchmark for evaluating systems' ability to follow requirements involving cross-level relations, joint constraints, and multi-turn state updates. Experiments reveal persistent difficulties with distant hierarchical relations, negative constraints, and multi-turn formulations, while the benefits of explicit reasoning vary across models and tasks. These findings show that semantic similarity does not imply satisfaction of the complete request; TRIAD provides a unified evaluation and diagnostic framework for studying how specific requirements can be reflected in retrieval rankings. The benchmark currently focuses on external retrieval for agentic systems; future work will extend it to retrieval and updates within their own memory systems. We have also yet to evaluate these capabilities within real agentic systems, where we plan to examine their impact on downstream task completion.

%% file: figures/appendix_layout.tex
\raggedbottom
\renewcommand{\floatpagefraction}{0.85}
\makeatletter
\let\paper@appendixsection\section
\renewcommand{\section}{\FloatBarrier\Needspace{8\baselineskip}\paper@appendixsection}
\let\paper@appendixsubsection\subsection
\renewcommand{\subsection}{\Needspace{5\baselineskip}\paper@appendixsubsection}
\setlength{\@fptop}{0pt}
\setlength{\@fpsep}{10pt plus 0fil}
\setlength{\@fpbot}{0pt plus 1fil}
\makeatother

%% file: sections/en/A_benchmark_construction.tex
\section{Benchmark Construction and Data}
\label{app:benchmark-construction-en}

This appendix supplements the benchmark construction described in Section~3. We detail the data sources, query and positive construction, candidate sampling, and diagnostic roles for the three tracks, followed by the coverage and quality checks of the final data. Although the tracks target different requests, they follow a common procedure: determine the results that satisfy the request, construct targeted distractors around that requirement, and then fix the candidate set and complete positive set for every query.

\begin{figure}[!htbp]
  \centering
  \input{figures/fig_pipeline}
  \caption{\textbf{Construction flow.} The abstraction, constraints, and state tracks start from hierarchical relations, joint conditions, and multi-turn state, respectively, and produce queries, positives, and typed negatives through filtering and annotation. They share potential-positive screening, a fixed 1{,}000-candidate evaluation protocol, and corresponding lower-complexity references.}
  \label{fig:pipeline-en}
\end{figure}
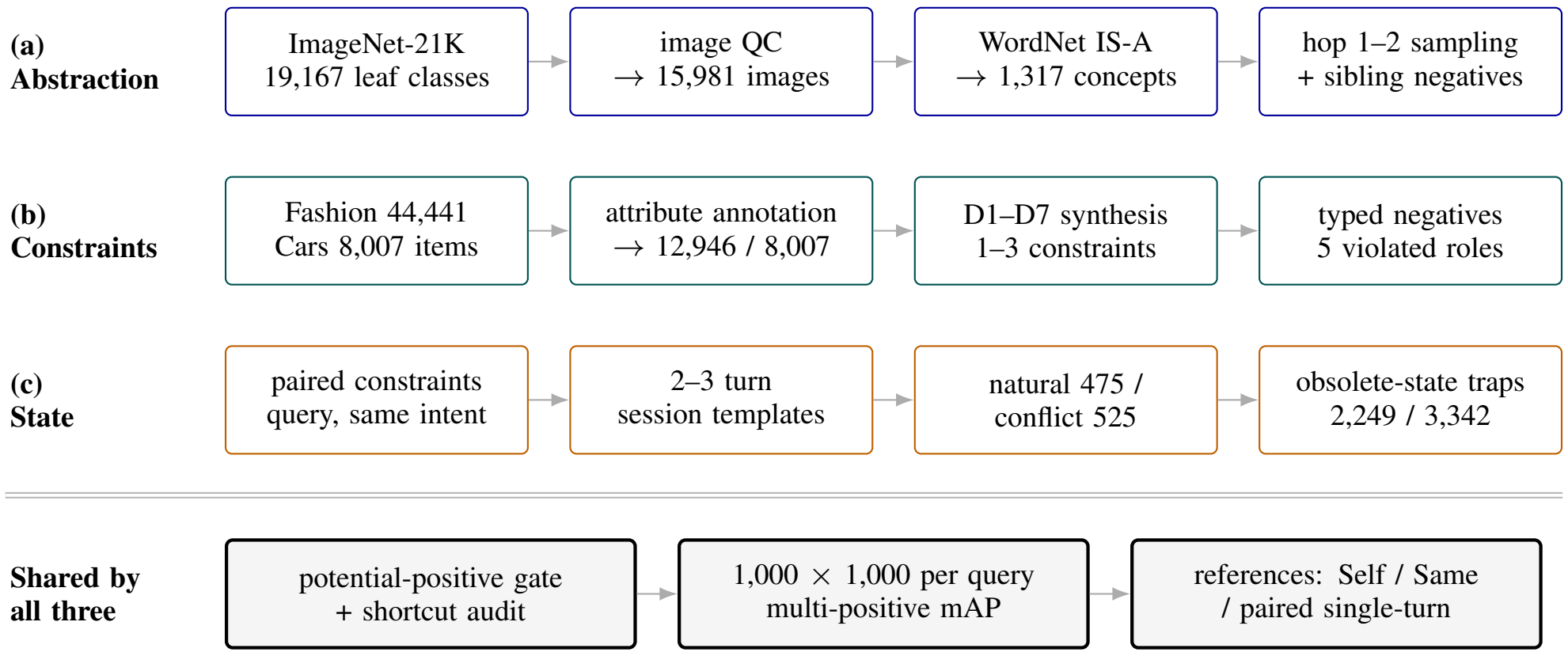

Figure~\ref{fig:pipeline-en} summarizes the process. The abstraction track builds bidirectional retrieval between instances and concepts from ImageNet-21K images and WordNet IS-A relations. The constraints track uses product images and attributes to construct queries containing positive and negative conditions. The state track retains the final intent of a constraints-track query and rewrites it as a multi-turn dialogue with accumulation or revision. All three tracks apply image or attribute checks, potential-positive screening, and typed-negative construction.

\subsection{Abstraction Track Construction}
\label{app:track-a-construction-en}

\paragraph{Image and concept pools.}
The abstraction track is built from ImageNet-21K and the WordNet 3.0 noun IS-A graph. For each of the 19,167 WordNet classes covered by ImageNet-21K, we select ten candidate images, compute image--class-text similarity with Qwen3-VL-Embedding, and retain the highest-scoring image. Deduplication and a minimum-short-side threshold of 224 pixels leave 15,981 instance images.

The WordNet noun graph contains 82,115 synsets and allows multiple hypernym paths. We define hierarchical distance by the shortest hypernym path. Candidate concepts must be one or two levels above a retained leaf class, have depth at least 5, and support at least five retained descendant images. These filters yield 1,317 internal concepts and remove 36 shallow concepts. A further 34 overly broad concepts are excluded from I2C level-2 positive selection but remain in the concept pool. Records contain synset names, definitions, paths, descendants, and lemmas; concept inputs use the format \texttt{"\{name\}. \{definition\}"}.

\paragraph{Bidirectional queries, positives, and candidates.}
I2C and C2I each contain 1,000 queries with 1,000 candidates per query. I2C takes an instance image and its leaf-class name as input, with concepts one or two levels above the instance as positives. C2I takes a concept name and definition as input, with images from classes one or two levels below it as positives. Each query retains at most five positives; the remaining candidates are taxonomy-derived sibling hard negatives and random negatives.

Construction first adds positives and taxonomy hard negatives, then fills the remaining positions with random samples. During random filling, I2C excludes the image's own class and its ancestors, while C2I excludes images from strict descendants of the query concept. In the final data, I2C contains 1.77 positives and 9.05 sibling negatives per query on average. C2I fixes five positives per query and contains 150.87 sibling-descendant negatives on average.

\paragraph{Self reference.}
Self measures basic matching when no cross-level relation is required. In I2C, the target is the instance's own leaf concept; in C2I, the targets are images belonging to that leaf concept. Self uses an independent 1,000-candidate leaf-concept pool. To avoid direct string matching between the concept name and the image leaf label, the official protocol provides only the WordNet definition on the concept side. In the construction-stage I2C Self class-name shortcut check, Qwen3-VL-Embedding-8B obtains an MRR of 0.978 when the concept input includes both its name and definition, compared with 0.815 for definition-only input. The formal protocol therefore uses definition-only concept inputs to avoid direct class-name matching. Each Self query has one positive, so per-query AP equals reciprocal rank and mAP across queries equals MRR.

\subsection{Constraints Track Construction}
\label{app:track-b-construction-en}

\paragraph{Domain corpora and attribute annotation.}
The constraints track uses Fashion Product Images and Stanford Cars, two product domains with distinct attribute systems. Fashion begins with 44,441 images. Qwen3.6-27B and Qwen3.5-397B-A17B independently annotate visual attributes under the v8.5 schema. The 13,345 items on which the two annotations agree enter further filtering, and 12,946 items across 21 match pools remain. Cars uses the Stanford Cars test set. Qwen3.5-27B annotates color, body type, view angle, setting, and related attributes; value normalization and programmatic checks leave 8,007 items from 196 model classes.

Attributes have three construction roles. A \texttt{constraint} attribute generates an explicitly requested modification; a \texttt{modifiable} attribute may vary within a fixed budget; and a \texttt{non\_constraint\_hard} attribute is not explicitly modified but specifies a property that must be preserved. A positive must satisfy all explicit conditions and the required preserved attributes. Typed negatives are generated according to whether they violate a positive condition, a negative condition, or an implicitly preserved attribute. Fashion colors and patterns additionally distinguish coarse families from fine-grained values to avoid treating near-synonyms as negatives.

\paragraph{Queries, positives, and difficulty profiles.}
Each domain contains 1,000 queries. A query provides a reference image, product text, and one to three requested changes; a candidate likewise contains an image and product attributes. The seven profiles D1--D7 correspond to 1P, 1N, 2P, 2N, 1P1N, 2P1N, and 1P2N, where P requires a value and N excludes one. Their sampling proportions are fixed during construction and are not adjusted to model results.

Query--target pairs are first generated from structured attributes and then naturalized by an LLM. A positive condition states only the target value and does not reveal the reference value; a negative condition may explicitly mention the excluded value. Ordered attributes such as length use comparative expressions such as \emph{longer} and \emph{shorter}. Programmatic checks verify that the query expresses all intended conditions, does not disclose the reference product's original value in positive modification requests, and preserves the requested negations.

\paragraph{Typed negatives and potential-positive screening.}
Each candidate set contains all validated positives, five diagnostic negative types, and random fillers:

\begin{center}
\small
\begin{tabularx}{\linewidth}{@{}lX@{}}
\toprule
Type & Operational definition \\
\midrule
\texttt{only\_neg\_violation} & violates only one requested negative condition \\
\texttt{only\_pos\_violation} & violates only one requested positive condition \\
\texttt{partial\_match} & satisfies explicit conditions but violates a required implicit property \\
\texttt{both\_violation} & violates both positive and negative conditions \\
\texttt{clip\_hard} & a visually similar general hard negative \\
\bottomrule
\end{tabularx}
\end{center}

The target quotas for the first four roles are 8/8/8/4 in Fashion and 5/5/3/3 in Cars; visually similar negatives and random samples fill the remaining positions. Every candidate passes a shared potential-positive gate, so only products that fail the complete positive rule can enter as negatives. Final scoring uses every valid positive for the query rather than only the target selected during construction.

\paragraph{Same reference.}
Same measures same-item retrieval without an attribute modification. A query contains a reference image, product text, and an instruction to retrieve the same item; candidates likewise contain product images and text. Other images of the same product may therefore be positive. Same remains multimodal, multi-positive retrieval, but removes the additional attribute changes. Fashion and Cars each contain 500 queries with independently constructed 1,000-candidate pools and average positive counts of 2.042 and 5.156, respectively.

\subsection{State Track Construction}
\label{app:track-c-construction-en}

\paragraph{Pairing with the constraints track.}
The state track rewrites constraints-track queries as two- or three-turn dialogues. Every multi-turn query retains the reference product, final active conditions, target product, and complete positive set of its source query, giving the two forms the same final retrieval intent. The constraints track states the complete request at once, whereas the state track requires the model to recover the currently active conditions from dialogue history.

The state track constructs a separate 1,000-candidate pool for each query. It preserves the source query's typed hard negatives and, for conflict sessions, adds revision traps that satisfy an old value but violate the final request. Mean candidate retention is 0.998565 in Fashion and 0.997076 in Cars. The tracks are therefore highly paired but not strictly same-pool, and the single-to-multi-turn difference measures the complete dialogue-history task change.

\paragraph{Session types and generation.}
The state track contains natural and conflict sessions. Natural sessions add active conditions over turns. Conflict sessions first introduce a value and then explicitly revise it. Conditions that are not revised remain active; superseded values no longer determine final relevance.

Each domain's 1,000 dialogues contain 525 conflict and 475 natural sessions: 300 two-turn conflict sessions originate from one-condition queries, 225 three-turn conflict sessions from two-condition queries, 225 two-turn natural sessions from two-condition queries, and 250 three-turn natural sessions from three-condition queries. Natural and conflict sessions with the same turn count mention the same number of attribute values, but one value in a conflict session is later superseded. All dialogues are generated by deterministic templates with seed 42 and are not paraphrased by an LLM.

\paragraph{Revision traps and state diagnostics.}
A \texttt{conflict\_trap} satisfies the superseded value, violates the final revision, and satisfies the remaining active conditions. Fashion contains 2,249 conflict traps covering 459 of 525 conflict queries; Cars contains 3,342 covering 517 of 525. An \texttt{implicit\_violation} satisfies explicit conditions but breaks an attribute that should remain unchanged. Other hard negatives retained from the constraints track continue to test general condition satisfaction.

For a fixed turn count, a conflict query has one fewer final active condition than a natural query. We report each session type's AP change relative to its source single-turn query to compare retrieval under history accumulation and revision; stratified results appear in Appendix~\ref{app:state-paired-changes-en}.

\subsection{Final Dataset Statistics}
\label{app:dataset-statistics-en}

\subsubsection{Data Distributions}
\label{sec:data-distribution-en}

\begin{figure}[!htbp]
\centering
\includegraphics[width=\linewidth]{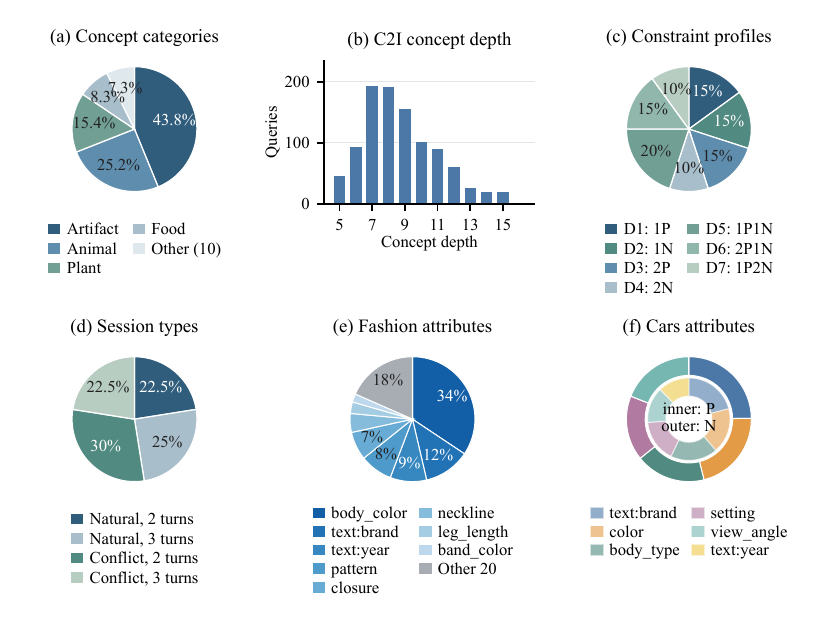}
\caption{\textbf{Data and attribute distributions.} (a) WordNet lexicographer categories of 1,317 concepts, with ten smaller categories grouped as Other; (b) concept depths of 1,000 C2I queries; (c) D1--D7 profiles, where P/N denote positive/negative conditions; (d) natural/conflict sessions crossed with two/three turns. Panels (c,d) each use 1,000 queries per domain, with identical proportions in Fashion and Cars. (e) The 1,950 Fashion conditions: eight fields shown separately and the remaining 20 grouped; shares below 5\% are unlabeled. (f) Cars: positive conditions on the inner ring (1,050), negative on the outer ring (900), with consistent field colors; text:year has no negative conditions.}
\label{fig:dataset-composition-en}
\label{fig:dataset-tracka-en}
\label{fig:attribute-coverage-en}
\end{figure}

Each of the six main subsets contains 1,000 queries with 1,000 candidates each. Figure~\ref{fig:dataset-composition-en} combines concept categories, C2I concept depths, constraint and session composition, and condition-attribute distributions in both domains. The 1,317 internal concepts span 14 WordNet lexicographer categories; artifact, animal, and plant account for 84.43\%, while C2I query concepts span depths 5--16. Fashion and Cars have identical constraint and session compositions, so each distribution is shown once as a pie chart.

\paragraph{Constraint-attribute coverage.}
The 12,946 Fashion items contain 56 category-specific fields, of which 28 generate explicit conditions; all six attributes of the 8,007 Cars items appear in queries. Each domain contains 1,950 explicit conditions: 1,050 positive and 900 negative, averaging 1.95 conditions per query. Attribute-use proportions appear in Figure~\ref{fig:dataset-composition-en}(e,f); Table~\ref{tab:attribute-schema-en} adds field cardinalities and applicable-item counts.

\begin{table}[!htbp]
\caption{Normalized value counts and applicable-item counts of condition attributes. Use proportions appear in Figure~\ref{fig:dataset-composition-en}(e,f), so use counts are not repeated here. Applicable-item counts reflect the category coverage of structured attributes, not their missingness; for example, heel\_type is defined only for footwear. Visual attributes in this table are used to construct and verify retrieval conditions, not supplied as additional product text. Model-visible text fields are listed in Appendix~\ref{app:serialization-en}.}
\label{tab:attribute-schema-en}
\setlength{\tabcolsep}{3pt}
\begin{center}
\input{tables/app_attribute_schema_details.tex}
\end{center}
\end{table}

\subsubsection{Candidate-Set Statistics}
\label{sec:candidate-statistics-en}

Table~\ref{tab:candidate-composition-en} reports the mean number of labeled positives and each hard-negative (HN) type per query in the six main subsets. Means are computed over all 1,000 queries in each subset, counting a type as zero for queries in which it is absent.

\input{tables/candidate_composition_en}

\FloatBarrier
\subsection{Candidate Similarity and Pool Scaling}
\label{app:pool-validity-en}
\label{app:e7-scaling-en}

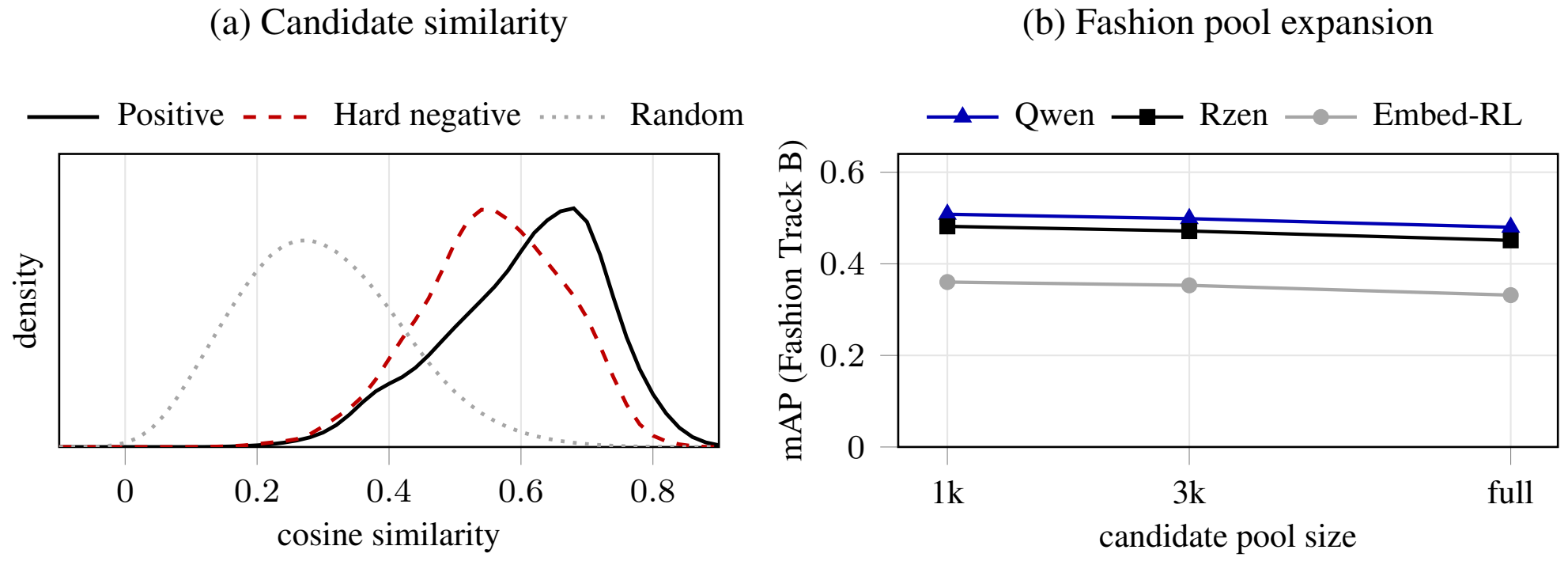
\begin{figure}[!htbp]
  \centering
  \centering
  \input{figures/fig_pool_validity}
  \caption{\textbf{Candidate similarity and pool scaling.} (a) Cosine-similarity densities pooled across Fashion/Cars under a fixed Qwen3-VL-Embedding-8B: positives, negation-only-violation negatives, and random fillers, with shared Gaussian KDE bandwidth 0.025. (b) Pool-scaling mAP for the same 1,000 Fashion constraints queries; full is the query-specific eligible corpus. Qwen, Rzen, and Embed-RL denote Qwen3-VL-Embedding-8B, RzenEmbed-v2, and Embed-RL-4B + T-CoT.}
  \label{fig:pool-validity-en}
  \label{fig:e7-en}
\end{figure}

\paragraph{Similarity confusability.}
Using a fixed Qwen3-VL-Embedding-8B encoder, we compare positives, negation-only-violation negatives, and random fillers across all Fashion/Cars query--candidate pairs (Figure~\ref{fig:pool-validity-en}a). The similarity distributions of negation-only violations and positives overlap substantially, making these negatives more confusable with positives than random fillers.

\paragraph{Pool expansion.}

For the same 1,000 Fashion constraints queries, we construct strictly nested 1k, 3k, and eligible-full pools, retaining all original positives and typed negatives. Eligible-full is the full-corpus pool after retaining the original candidates and excluding label-matching potential positives and the query image from added products. The 3k pool uses deterministic sampling with seed 2027; all three models share pool manifests. Eligible-full pools contain 11,956--12,945 candidates per query (mean 12,916).

For the three evaluated models on Fashion constraints, expanding the fixed pool from 1,000 to approximately 13,000 eligible candidates lowers mAP by about 0.03 without changing model ordering (Figure~\ref{fig:pool-validity-en}b). The main model comparison is therefore stable to this increase in candidate-set size under the present construction.

\subsection{Construction Validation and Quality Audit}
\label{app:quality-assurance-en}

Each track undergoes checks aligned with its task definition. The abstraction track verifies WordNet relations, shortest-path levels, the blacklist for I2C level-2 positives, image resolution, deduplication, and random-filler rules. The constraints track rechecks positive eligibility, typed-negative roles, and the potential-positive gate against the attribute tables. The state track further verifies that final active conditions match the source single-turn query and that revision traps satisfy the old value while violating the final one.

Attribute annotation uses domain-specific quality control. Fashion retains only items on which two visual-attribute annotations agree, followed by completeness and candidate-construction checks. Cars uses one model followed by value normalization and programmatic validation. We further audit query quality and visual-attribute quality separately: 20 queries from each of the four constraints and state subsets and 40 images from each of Fashion and Cars. Human review judges 507 of 513 assessable visual attributes correct (98.83\%), with eight additional attributes marked undetermined. Full procedures and results appear in Appendix~\ref{app:e6a-audit-en}.

These checks cover construction rules, query expression, and sampled visual attributes, providing evidence for task consistency and annotation quality in the finalized data.

Extended design and ranking comparisons with existing benchmarks are provided in Appendix~\ref{app:bench-comparison-en}.

%% file: figures/fig_pipeline.tex
\begin{tikzpicture}[
  font=\scriptsize,
  lab/.style={align=left, text width=15.5mm, inner sep=1pt, font=\scriptsize\bfseries},
  box/.style={draw, rounded corners=1.5pt, align=center, inner sep=2.5pt,
    minimum height=9mm, text width=23.5mm},
  abox/.style={box, draw=blue!60!black},
  bbox/.style={box, draw=teal!65!black},
  cbox/.style={box, draw=orange!75!black},
  gate/.style={box, draw=black, thick, fill=gray!8, text width=32.4mm},
  ar/.style={-{Latex[length=1.5mm]}, gray!65},
  node distance=3.4mm,
]

\node[abox] (a1) {ImageNet-21K\\19{,}167 leaf classes};
\node[lab, left=2mm of a1] (la) {(a)\\Abstraction};
\node[abox, right=of a1] (a2) {image QC\\$\rightarrow$ 15{,}981 images};
\node[abox, right=of a2] (a3) {WordNet IS-A\\$\rightarrow$ 1{,}317 concepts};
\node[abox, right=of a3] (a4) {hop 1--2 sampling\\+ sibling negatives};

\node[bbox, below=5mm of a1] (b1) {Fashion 44{,}441\\Cars 8{,}007 items};
\node[lab, left=2mm of b1] (lb) {(b)\\Constraints};
\node[bbox, right=of b1] (b2) {attribute annotation\\$\rightarrow$ 12{,}946 / 8{,}007};
\node[bbox, right=of b2] (b3) {D1--D7 synthesis\\1--3 constraints};
\node[bbox, right=of b3] (b4) {typed negatives\\5 violated roles};

\node[cbox, below=5mm of b1] (c1) {paired constraints\\query, same intent};
\node[lab, left=2mm of c1] (lc) {(c)\\State};
\node[cbox, right=of c1] (c2) {2--3 turn\\session templates};
\node[cbox, right=of c2] (c3) {natural 475 /\\conflict 525};
\node[cbox, right=of c3] (c4) {obsolete-state traps\\2{,}249 / 3{,}342};

\foreach \s/\t in {a1/a2, a2/a3, a3/a4, b1/b2, b2/b3, b3/b4, c1/c2, c2/c3, c3/c4}
  \draw[ar] (\s) -- (\t);

\draw[gray!55, double, double distance=0.7pt]
  ([yshift=-3.4mm]la.west |- c1.south) -- ([yshift=-3.4mm]a4.east |- c1.south);
\node[gate, below=7mm of c1.south west, anchor=north west] (g1)
  {potential-positive gate\\+ shortcut audit};
\node[lab, left=2mm of g1, text width=15.5mm] (ls) {Shared by\\all three};
\node[gate, right=of g1] (g2) {1{,}000 $\times$ 1{,}000 per query\\multi-positive mAP};
\node[gate, right=of g2] (g3) {references: Self / Same\\/ paired single-turn};
\draw[ar] (g1) -- (g2);
\draw[ar] (g2) -- (g3);
\end{tikzpicture}

%% file: tables/app_attribute_schema_details.tex
\begin{tabular}{@{}lrr@{\qquad}lrr@{}}
\toprule
Fashion field & Values & Items & Fashion field & Values & Items \\
\midrule
band\_color & 24 & 736 & heel\_type & 4 & 593 \\
band\_material & 4 & 736 & leg\_length & 4 & 879 \\
body\_color & 26 & 11,840 & lens\_color & 19 & 370 \\
bra\_type & 5 & 286 & lens\_tint & 3 & 370 \\
buckle\_type & 4 & 462 & neckline & 11 & 5,079 \\
cap\_type & 6 & 103 & pattern & 19 & 6,398 \\
closure & 13 & 3,974 & rise & 2 & 454 \\
collar & 4 & 144 & sleeve\_length & 3 & 4,937 \\
dial\_color & 19 & 736 & sleeve\_type & 3 & 286 \\
dial\_shape & 6 & 736 & sock\_length & 4 & 176 \\
dress\_length & 3 & 202 & strap & 3 & 624 \\
fold\_type & 4 & 458 & strap\_style & 8 & 876 \\
frame\_color & 19 & 370 & text:brand & 334 & 12,946 \\
frame\_shape & 6 & 370 & text:year & 12 & 12,946 \\
\bottomrule
\end{tabular}

\vspace{2mm}
\begin{tabular}{@{}lr@{\qquad}lr@{}}
\toprule
Cars field & Values & Cars field & Values \\
\midrule
text:brand & 49 & setting & 5 \\
color & 11 & view\_angle & 4 \\
body\_type & 7 & text:year & 2 \\
\bottomrule
\end{tabular}

%% file: tables/candidate_composition_en.tex
\begin{table}[!htb]
  \caption{\textbf{Mean candidate counts per query.} Counts follow mutually exclusive construction labels; ``--'' indicates that a type is not used in a subset. Pos.: positives; Sib.: hierarchical-neighbor HN; P-V/N-V: violations of positive/negative conditions only; PN-V: both polarities violated; Imp-V: preserved-attribute violations (\texttt{partial\_match}); Vis.: visually similar HN; Trap: the exclusive revision-trap type. Trap flags may also overlay existing HN types; including overlays, C--Fashion/Cars means are 2.249/3.342. The table retains original types without double counting.}
  \label{tab:candidate-composition-en}
  \small
  \setlength{\tabcolsep}{7pt}
  \renewcommand{\arraystretch}{1.08}
  \begin{center}
  \begin{tabular}{@{}lrrrrrr@{}}
    \toprule
    & \multicolumn{2}{c}{A: Abstraction} & \multicolumn{2}{c}{B: Constraints} & \multicolumn{2}{c}{C: State} \\
    \cmidrule(lr){2-3}\cmidrule(lr){4-5}\cmidrule(l){6-7}
    Type & I2C & C2I & Fashion & Cars & Fashion & Cars \\
    \midrule
    Pos.  & 1.768 & 5.000   & 1.653 & 1.982 & 1.653 & 1.982 \\
    Sib.  & 9.053 & 150.872 & --    & --    & --    & --    \\
    P-V   & --    & --      & 4.539 & 3.714 & 4.539 & 3.714 \\
    N-V   & --    & --      & 2.384 & 2.552 & 2.384 & 2.552 \\
    PN-V  & --    & --      & 1.401 & 1.350 & 1.401 & 1.350 \\
    Imp-V & --    & --      & 6.724 & 2.781 & 6.724 & 2.781 \\
    Vis.  & --    & --      & 5.100 & 4.603 & 5.100 & 4.603 \\
    Trap  & --    & --      & --    & --    & 1.146 & 2.195 \\
    \bottomrule
  \end{tabular}
  \end{center}
\end{table}

%% file: figures/fig_pool_validity.tex
\begin{tikzpicture}
\begin{groupplot}[
 paper main axis,
 group style={group size=2 by 1,horizontal sep=1.5cm},
 scale only axis, width=0.395\linewidth,height=2.45cm,ymin=0,
]
\nextgroupplot[
  title={(a) Candidate similarity},
  xlabel={cosine similarity}, xlabel style={yshift=2pt},
  ylabel={density}, ylabel style={yshift=-4pt},
  xmin=-0.10, xmax=0.90, ymax=4.4, ytick=\empty,
  title style={yshift=17pt},
  legend style={font=\paperplotfont, at={(0.5,1.015)}, anchor=south, legend columns=3,
    column sep=2pt, draw=none, fill=none},
]
\addplot[black, line width=0.9pt] coordinates {
  (-0.10,0.0000) (-0.08,0.0000) (-0.06,0.0000) (-0.04,0.0000) (-0.02,0.0000) (0.00,0.0000) (0.02,0.0000) (0.04,0.0000)
  (0.06,0.0000) (0.08,0.0000) (0.10,0.0001) (0.12,0.0008) (0.14,0.0033) (0.16,0.0082) (0.18,0.0154) (0.20,0.0253)
  (0.22,0.0407) (0.24,0.0645) (0.26,0.0976) (0.28,0.1445) (0.30,0.2171) (0.32,0.3285) (0.34,0.4862) (0.36,0.6723)
  (0.38,0.8367) (0.40,0.9496) (0.42,1.0499) (0.44,1.1895) (0.46,1.3764) (0.48,1.5892) (0.50,1.7994) (0.52,2.0003)
  (0.54,2.2009) (0.56,2.4062) (0.58,2.6453) (0.60,2.9367) (0.62,3.2140) (0.64,3.4045) (0.66,3.5383) (0.68,3.5832)
  (0.70,3.3803) (0.72,2.8850) (0.74,2.2479) (0.76,1.6497) (0.78,1.1680) (0.80,0.7938) (0.82,0.5046) (0.84,0.2896)
  (0.86,0.1494) (0.88,0.0694) (0.90,0.0263)};
\addlegendentry{Positive}
\addplot[red!75!black, dashed, line width=0.9pt] coordinates {
  (-0.10,0.0000) (-0.08,0.0000) (-0.06,0.0000) (-0.04,0.0000) (-0.02,0.0000) (0.00,0.0000) (0.02,0.0000) (0.04,0.0000)
  (0.06,0.0000) (0.08,0.0001) (0.10,0.0006) (0.12,0.0026) (0.14,0.0067) (0.16,0.0129) (0.18,0.0245) (0.20,0.0444)
  (0.22,0.0662) (0.24,0.0862) (0.26,0.1192) (0.28,0.1916) (0.30,0.3094) (0.32,0.4428) (0.34,0.5795) (0.36,0.7598)
  (0.38,1.0152) (0.40,1.3227) (0.42,1.6272) (0.44,1.9095) (0.46,2.2307) (0.48,2.6388) (0.50,3.0709) (0.52,3.4036)
  (0.54,3.5630) (0.56,3.5507) (0.58,3.4182) (0.60,3.2327) (0.62,3.0146) (0.64,2.7668) (0.66,2.5224) (0.68,2.2686)
  (0.70,1.9387) (0.72,1.5137) (0.74,1.0499) (0.76,0.6373) (0.78,0.3404) (0.80,0.1703) (0.82,0.0864) (0.84,0.0405)
  (0.86,0.0142) (0.88,0.0032) (0.90,0.0004)};
\addlegendentry{Hard negative}
\addplot[gray!70, dotted, line width=0.9pt] coordinates {
  (-0.10,0.0001) (-0.08,0.0004) (-0.06,0.0018) (-0.04,0.0072) (-0.02,0.0231) (0.00,0.0616) (0.02,0.1408) (0.04,0.2793)
  (0.06,0.4868) (0.08,0.7576) (0.10,1.0731) (0.12,1.4092) (0.14,1.7460) (0.16,2.0724) (0.18,2.3786) (0.20,2.6505)
  (0.22,2.8698) (0.24,3.0204) (0.26,3.0968) (0.28,3.1022) (0.30,3.0429) (0.32,2.9286) (0.34,2.7698) (0.36,2.5715)
  (0.38,2.3375) (0.40,2.0774) (0.42,1.8035) (0.44,1.5285) (0.46,1.2667) (0.48,1.0307) (0.50,0.8261) (0.52,0.6527)
  (0.54,0.5087) (0.56,0.3913) (0.58,0.2979) (0.60,0.2253) (0.62,0.1687) (0.64,0.1244) (0.66,0.0901) (0.68,0.0640)
  (0.70,0.0439) (0.72,0.0289) (0.74,0.0182) (0.76,0.0109) (0.78,0.0061) (0.80,0.0031) (0.82,0.0014) (0.84,0.0005)
  (0.86,0.0002) (0.88,0.0000) (0.90,0.0000)};
\addlegendentry{Random}
\nextgroupplot[
  ymin=0,
  title={(b) Fashion pool expansion},
  xmode=log, log basis x=10,
  xmin=800, xmax=16000, xtick={1000,3000,12916}, xticklabels={1k,3k,full},
  xlabel={candidate pool size}, xlabel style={yshift=2pt},
  ylabel={mAP (Fashion Track B)}, ylabel style={yshift=-4pt},
  ymax=0.64, ytick={0,0.2,0.4,0.6},
  title style={yshift=17pt},
  legend style={font=\paperplotfont, at={(0.5,1.015)}, anchor=south, legend columns=3,
    column sep=1.5pt, draw=none, fill=none},
]
\addplot[blue!70!black, mark=triangle*, mark size=1.8pt, line width=0.8pt] coordinates {
  (1000,0.5082)(3000,0.4986)(12916,0.4798)};
\addlegendentry{Qwen}
\addplot[black, mark=square*, mark size=1.5pt, line width=0.8pt] coordinates {
  (1000,0.4818)(3000,0.4716)(12916,0.4512)};
\addlegendentry{Rzen}
\addplot[gray!70, mark=*, mark size=1.5pt, line width=0.8pt] coordinates {
  (1000,0.3602)(3000,0.3530)(12916,0.3316)};
\addlegendentry{Embed-RL}
\end{groupplot}
\end{tikzpicture}%

%% file: sections/en/B_evaluation_details.tex
\section{Evaluation Protocol and Implementation Details}
\label{app:evaluation-details-en}

This appendix describes the metrics, model adaptations, input organization, and runtime environments used for the main evaluation and diagnostic experiments. Per-model software versions appear in the runtime tables, and task input formats in Appendix~\ref{app:serialization-en}. Supplementary analyses of the main evaluation appear in Appendix~\ref{app:main-results-details-en}, supplementary experiments in Appendix~\ref{app:supplementary-experiments-en}, and human quality review and relevance ratings in Appendix~\ref{app:quality-audit-en}. Three interpretive limits that apply throughout are stated once here. The evaluated panel is restricted to publicly available checkpoints, and its comparisons do not establish a performance gap between open models and closed APIs. Cross-system comparisons describe behaviour only and cannot be read causally, since the systems differ in backbone, scale, training data, and embedding interface. The strength of a family-level claim is limited by member count: only families with at least three members including a same-source control get family-level conclusions, while single-member families yield single-point observations.

\subsection{Formal Retrieval and Metric Definitions}
\label{app:metric-definitions-en}

This paper uses mean Average Precision (mAP) as the only primary metric. We define the notation as follows: \(C_q\) is the query-specific candidate set, \(P_q\) is the complete labeled positive set used for scoring, and \(Q\) is the evaluated query set; the indicator \(r_{q,k}\) equals \(1\) if the candidate at rank \(k\) belongs to \(P_q\), and \(0\) otherwise. mAP is computed as
\begin{equation}
\label{eq:map-appendix-en}
\begin{gathered}
\mathrm{Precision}_q@k = \frac{1}{k}\sum_{j=1}^{k}r_{q,j},\qquad
\mathrm{AP}_q = \frac{1}{|P_q|}\sum_{k=1}^{|C_q|}\mathrm{Precision}_q@k\,r_{q,k},\\[4pt]
\mathrm{mAP} = \frac{1}{|Q|}\sum_{q\in Q}\mathrm{AP}_q.
\end{gathered}
\end{equation}
The divisor of \(\mathrm{AP}_q\) is \(|P_q|\), not the ranking length or the number of positives retrieved in a Top-\(K\) prefix; multiple positives are scored together within a query rather than converted into independent single-positive trials. AP equals reciprocal rank only for genuinely single-positive controls such as Self; Same remains multi-positive and is scored over its complete positive set.

mAP is our only primary metric. We aggregate at the highest available precision and compute ranks and differences before rounding displayed values. Hard-negative entry rates, dialogue AP changes, and positive coverage@30 serve the corresponding diagnostic analyses and are not combined into an additional aggregate score. Evaluation saves complete per-query rankings for recomputation; each main-task ranking contains 1,000 distinct candidates, excludes the query image itself, and retains the complete annotated positive set.

One convention behind the main table's I2C column deserves a note: that column counts hop-1 and hop-2 ancestor concepts together in \(P_q\) over one shared candidate pool, so it is a single number that merges two abstraction distances rather than an average of per-hop results. Per-hop scoring preserves the original candidate pool and full ranking, retaining only the current hop's positives in \(P_q\); candidates from the other hop are not removed and ranks are not compressed. The descriptive values in Section~\ref{sec:hierarchy-en} aggregate the eligible queries separately for each hop: all 1,000 for hop-1 and the 711 with hop-2 positives for hop-2. Paired hop-1 versus hop-2 tests instead use only the 711 queries shared by both. The descriptive per-hop means and paired differences therefore use different query populations. The main-table I2C value cannot be recovered as a weighted combination of the per-hop values, and the two should not be substituted for each other in comparisons. Appendix~\ref{app:hierarchy-scoring-sensitivity-en} additionally removes labeled positives at the other distance and compresses ranks to assess occupancy effects; this sensitivity analysis does not replace the main-task or original distance-specific scoring.

The benchmark assigns role labels to hard negatives: the five constraints-track types are defined in Appendix~\ref{app:track-b-construction-en}, and obsolete-value traps in Appendix~\ref{app:track-c-construction-en}. These labels support statistics on each type's entry into top-ranked results: for each type of constraints-track negative, top-10 interference rate is the proportion of negatives of that type ranked at most 10, and the conflict-trap top-10 rate shows whether obsolete instructions reach top-ranked results. Because the conflict-trap rate can show a floor effect once the ranking degrades, it is read across groups of one model rather than across subsets of differing overall difficulty, and it does not enter the primary metric.

The constraints and state tracks are paired by \texttt{query\_id}. Each query's dialogue cost is its state-track AP minus its corresponding constraints-track AP, averaged within each session-type and turn-count group; negative values indicate degradation under multi-turn expression. AP/mAP changes are reported as signed absolute differences, without division by the baseline score. The two tracks use highly overlapping but non-identical candidate pools, so the difference describes the overall task change from a single-turn request to a multi-turn dialogue.

For scale, random-ranking mAP on the six main subsets ranges from 0.008 to 0.011.

\subsection{Evaluated Models and Reasoning Configurations}
\label{app:model-adapters-en}

\input{sections/en/B_model_configuration}

\subsection{Query and Candidate Inputs}
\label{app:serialization-en}

This section gives the query and doc (candidate) input organization for each track directly. Textual attributes and instructions retain their original English wording; images are supplied separately as visual inputs and are not embedded in the text strings below.

\paragraph{Abstraction.}
For I2C, the query is an instance image plus its leaf-class name and the doc is the concept text; for C2I, the query is the concept text and the doc is an instance image plus its leaf-class name. Concept text is written as \texttt{\{name\}. \{definition\}}. The archived shared and UME-R1 interfaces use the following query-side directional instructions; candidates do not receive them:

\begin{paperprompt}[nobreak=true]
\begin{Verbatim}
I2C: Retrieve the text description of the concept category this image and text belong to.
C2I: Retrieve images that are instances of the described concept category.
\end{Verbatim}
\end{paperprompt}

These directional instructions are supplied in two ways depending on the model interface: models whose interface exposes a dedicated instruction field (e.g.\ Qwen3-VL-Embedding and RzenEmbed) receive them through that field rather than inside the content text, whereas models without such a field (the dual-tower encoders and Embed-RL) prepend the same instruction to the query content as a plain-text prefix. Both routes use identical instruction strings. The constraints and state tracks carry no benchmark-level directional instruction, and each model keeps only its own outer wrapper there.

Self uses definition-only concept text: the concept name is omitted from I2C candidates and C2I queries. Its construction is described in Appendix~\ref{app:track-a-construction-en}.

\paragraph{Constraints.}
The query is a reference product image, its reference product text, and a modification request; the doc is a product image and its text attributes. The query text is organized as:

\begin{paperprompt}[nobreak=true]
\begin{Verbatim}
Reference product info: {reference_attributes}. User query: {query_text}
\end{Verbatim}
\end{paperprompt}

The doc text is organized as:

\begin{paperprompt}[nobreak=true]
\begin{Verbatim}
Product info: {candidate_attributes}
\end{Verbatim}
\end{paperprompt}

Fashion attributes are ordered as gender, brand, year, and season; Cars uses brand and the actual year. Visual attribute labels used to define relevance are not supplied as additional product text. Qwen3-VL retains the outer instruction \texttt{Represent the user's input.}, and UME-R1 retains its Gen/Disc wrapper.

\paragraph{State.}
The doc input is identical to that in the constraints track; the query appends the complete turn-ordered history to the reference product information, organized as:

\begin{paperprompt}[nobreak=true]
\begin{Verbatim}
Reference product info: {reference_attributes}. User conversation:
[Round 1] User: {turn_1}
...
[Round T] User: {turn_T}
\end{Verbatim}
\end{paperprompt}

Dialogues with revisions retain both old values and updated requests instead of being replaced with the final state; the model must identify the conditions that remain active from the full input.

\paragraph{Same.}
Same retains the constraints track's image--text serialization. Fashion and Cars each use six fixed exact-same templates, assigned cyclically in construction order across their 500 queries:

\begin{paperprompt}[nobreak=true]
\begin{Verbatim}
Fashion:
Find the exact same item as shown in the picture.
I want this exact product — find it for me.
Looking for this identical piece, the very same one.
Show me this same item, not a similar one — the same.
Need to find the exact match of what's in this photo.
Find me this one — identical product, same everything.
\end{Verbatim}
\end{paperprompt}

\begin{paperprompt}[nobreak=true]
\begin{Verbatim}
Cars:
Find this exact same car as shown in the picture.
I want this very car — the identical one.
Looking for exactly this car, the same one.
Show me this same vehicle, not a similar one.
Need to find the exact match of the car in this photo.
Find me this one — the identical car.
\end{Verbatim}
\end{paperprompt}

\paragraph{The same input on each family's encoding path.}
The task inputs above are identical for every model, but each family encodes them differently. Dual-tower encoders have no native joint image--text input: image and text pass through separate towers and are combined by the pre-fixed 1:1 addition with L2 normalization, so composed queries rely on this fixed fusion rule rather than any weight tuned on TRIAD. In UME-R1's Disc mode both the query and candidate sides read the representation at \texttt{<disc\_emb>}, whereas Gen mode generates reasoning text on the query side only and then takes the last-layer representation of the generation step aligned to \texttt{<gen\_emb>}; candidates use Disc encoding in both configurations. Modality fusion, reasoning integration, and interface adaptations used here are described in Appendix~\ref{app:model-adapters-en}.

\subsection{Code and Data Availability}
\label{app:resource-availability-en}

The evaluation code, complete dataset, and leaderboard will be made publicly available after the double-blind review process concludes.

%% file: sections/en/B_model_configuration.tex
This section describes the reasoning integration of the evaluated systems and their inference runtime environment. Task fields of queries and candidates are in Appendix~\ref{app:serialization-en}.

\subsubsection{Reasoning Integration and Configuration Changes}

All models encode inputs through their official code and interfaces. B-Tokens uses generative supervision during training but forms its inference-time embedding in a single forward pass, so it is listed as a discriminative rather than a reasoning embedder~\citep{sun2026btoks}. Beyond the official implementations, configuration changes were made only for the following four items:

\paragraph{Modality fusion for dual-tower encoders.}
CLIP, SigLIP-2, and jina-clip-v2 encode images and text separately. Following MMEB's equal-weight feature addition~\citep{jiang2024vlm2vec}, our implementation adds the two towers' unnormalized projected features with 1:1 weights and then applies L2 normalization; queries and candidates use the same fusion function, and no fusion weight is learned on TRIAD. This is a protocol concession: the cross-family gap in the body describes the behavior of ``the dual towers as evaluated plus this fusion protocol'' rather than a causal effect of the architecture.

\paragraph{Embedding-token fallback in UME-R1's Gen mode.}
Gen mode asks the model to generate reasoning text on the query side and uses the last-layer representation of the generation step aligned to the \texttt{<gen\_emb>} marker as the embedding. A fraction of inputs did not emit this marker, or the marker was rejected by the end-of-output check; in those cases the run falls back to the hidden state of the last usable generation step. Disc mode is unaffected and reads the representation at the \texttt{<disc\_emb>} position on both query and candidate sides. In addition, UME-R1 receives a $28\times28$ white placeholder image on pure-text abstraction-track inputs; this is an interface adaptation and introduces no additional dataset imagery.

\paragraph{Substitute external reasoner for Embed-RL.}
The Embed-RL paper does not release the weights of its reasoner. We therefore generate T-CoT offline with the reasoner's training backbone, Qwen3-VL-8B-Instruct, and then encode with Embed-RL-2B/4B. In the original paper's Embed-RL-2B ablation, using Qwen3-VL-8B-Instruct as the reasoner scores 1.2 percentage points lower on MMEB-V2 image tasks than using the trained reasoner~\citep[Table~3]{jiang2026embed}.

\paragraph{E5-Omni's encoding interfaces.}
Its official query interface supports only pure text, so image and image--text inputs are all encoded through the document interface, while pure-text abstraction-track items still use the query interface. These interface names do not map one-to-one to the benchmark's query and candidate roles.

UME-R1's Gen mode retains the checkpoint's sampling settings rather than deterministic decoding; inputs include the multimodal token-type metadata required by the interface. TWN loads reasoning and embedding adapters plus a routing gate on Qwen3-VL-4B-Instruct and decides per input whether to generate on either side; because it changes backbone, training, routing, and wrapper together, its difference from other configurations is not a controlled ablation.

\subsubsection{Inference Runtime Environment}

Embedding encoding uses one NVIDIA A100-SXM4-80GB GPU per job. External T-CoT generation jobs covered by available logs use four GPUs with four-way tensor parallelism. Table~\ref{tab:inference-env-en} consolidates model dependencies and compute precision.

\input{tables/tab_inference_env_en}

\input{sections/en/B_release_environment}

%% file: tables/tab_inference_env_en.tex
\begin{table}[!htbp]
\caption{Software dependencies and experimental precision by model. NI: not installed; --: unrecorded.}
\label{tab:inference-env-en}
\begin{center}
\begingroup\footnotesize
\setlength{\tabcolsep}{3pt}
\renewcommand{\arraystretch}{1.15}
\begin{tabularx}{\linewidth}{@{}>{\raggedright\arraybackslash}Xcccccc@{}}
\toprule
Model & PyTorch & Transformers & Flash-Attn & PEFT & NumPy & Precision \\
\midrule
CLIP / SigLIP-2 & 2.6.0 & 4.52.4 & 2.7.4 & NI & 1.26.4 & -- \\
jina-clip-v2 & 2.6.0 & 4.46.3 & 2.7.4 & NI & 1.26.4 & -- \\
WeMM-Embedding-9B & 2.6.0 & 5.14.1 & 2.7.2.post1 & NI & 1.26.4 & BF16 \\
E5-Omni-7B & 2.6.0 & 5.14.1 & 2.7.2.post1 & NI & 1.26.4 & BF16 \\
RzenEmbed-v2 & 2.5.1 & 4.49.0 & 2.7.4.post1 & NI & 2.2.6 & -- \\
Qwen3-VL-Embedding-8B / 2B & 2.6.0 & 5.14.1 & 2.7.2.post1 & NI & 1.26.4 & BF16 \\
OmniEmbed-3B & 2.6.0 & 5.14.1 & 2.7.2.post1 & NI & 1.26.4 & -- \\
VLM2Vec-2B / 7B / V2.0-2B & 2.6.0 & 4.52.3 & 2.7.2.post1 & 0.20.0 & 1.26.4 & BF16 \\
B-Tokens-2B & 2.6.0 & 5.14.1 & 2.7.2.post1 & NI & 1.26.4 & -- \\
Embed-RL-2B / 4B & 2.6.0 & 5.15.0.dev0 & NI & 0.20.0 & 2.2.6 & BF16 \\
UME-R1-7B (Gen/Disc) & 2.6.0 & 4.52.4 & 2.7.4 & NI & 1.26.4 & BF16 \\
TWN-4B & 2.6.0 & 4.57.1 & 2.7.4 & 0.18.0 & 1.26.4 & BF16 \\
\midrule
CoT for Embed-RL (Qwen3-VL-8B-Instruct) & 2.10.0 & 4.57.6 & 2.8.3 & NI & 2.2.6 & BF16 \\
Qwen3-VL-Reranker-8B & 2.6.0 & 5.14.1 & 2.7.2.post1 & NI & 1.26.4 & BF16 \\
\bottomrule
\end{tabularx}
\endgroup
\end{center}
\end{table}

%% file: sections/en/B_release_environment.tex
Release validation covers small-sample I2C and Cars constraints tasks for 17 encoding configurations. Jina-CLIP-v2 additionally requires timm 1.0.27; E5-Omni's import chain requires audio libraries including librosa 0.11.0 and lazy\_loader 0.5.

%% file: sections/en/C_diagnostic_details.tex
\section{Supplementary Analyses of the Main Evaluation}
\label{app:main-results-details-en}

This appendix supplements the main results with Top-1 labeled-positive hit rates, Self and Same reference tasks, rank agreement, constraint composition, single-to-multi-turn changes and session strata, complete reasoning comparisons, and hierarchy-scoring sensitivity. Input ablations, reranking, and reliability analyses are in Appendix~\ref{app:supplementary-experiments-en}; pool scaling appears in Appendix~\ref{app:pool-validity-en}. Rzen denotes RzenEmbed-v2, Qwen-8B/2B denotes Qwen3-VL-Embedding-8B/2B, and Omni denotes OmniEmbed-3B; other abbreviations and settings follow the row labels.

\input{sections/en/C_top1_results}

For query-paired comparisons, we first compute AP differences between the two settings and then use 3,000 query-level bootstrap resamples to report 95\% percentile intervals for the mean difference. Paired tests use the two-sided Wilcoxon signed-rank test with Holm correction within each analysis family. The rank-biserial effect size is the difference between positive and negative rank sums divided by their sum; it ranges from $-1$ to $1$, with the sign indicating direction. Each figure or table specifies its reference setting and marker definitions; $^\dagger$ in the session-stratified table only indicates that the bootstrap interval excludes zero.

\input{sections/en/C_self_same_results}

\subsection{Rank Agreement Across Subsets}
\label{app:rank-agreement-en}

\begin{figure}[!htbp]
  \centering
  \includegraphics[width=0.50\linewidth]{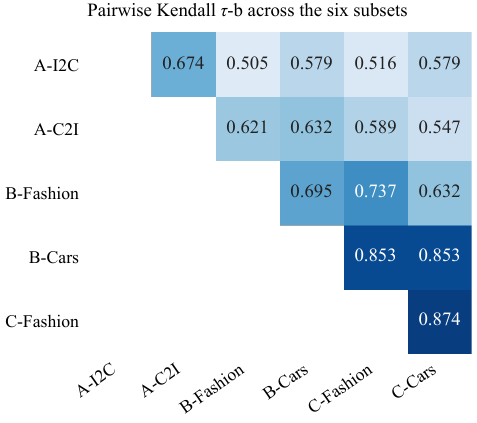}
  \caption{\textbf{Rank agreement across the six subsets (Kendall $\tau$-b, 20 configurations, all measured).} Colour depth and cell values are proportional; a lower $\tau$ means the model order produced by one subset is less substitutable by the other. The 15 pairs span 0.505--0.874 (mean 0.659), with within- and cross-track means of 0.747 and 0.637; ranks are computed before display rounding with ties averaged.}
  \label{fig:tau-en}
\end{figure}

Figure~\ref{fig:tau-en} shows model-rank agreement across the six subsets. Computed from scores before display rounding for 20 configurations, Kendall $\tau$-b ranges from 0.505 to 0.874 across 15 subset pairs, with a mean of 0.659; within-track agreement averages 0.747, versus 0.637 across tracks. Rankings also vary by domain: Fashion--Cars agreement is 0.695 for constraints and 0.874 for state, while constraints--state agreement within Fashion and Cars is 0.737 and 0.853, respectively. Relative model strengths therefore vary with both retrieval demands and domains. We retain the six subset scores to expose differences that aggregation can hide; Figure~\ref{fig:profiles-en} shows each configuration's rank changes across subsets.

\FloatBarrier
\subsection{Constraint Composition and Retrieval Performance}
\label{app:track-b-statistics-en}

Table~\ref{tab:track-b-stats-en} reports mAP for 17 configurations across seven constraint compositions in both domains. The groups contain different queries; below, we examine constraint count, polarity, and category/attribute composition without query-paired tests between groups.

\begin{table}[!htbp]
\caption{mAP across the seven difficulty tiers in both domains. Both Embed-RL configurations use +T-CoT.}
\label{tab:track-b-stats-en}
\setlength{\tabcolsep}{3pt}
\begin{center}
\input{tables/app_track_b_stats.tex}
\end{center}
\end{table}

\input{sections/en/C_constraint_controls}

\subsection{State Track Paired Changes}
\label{app:state-paired-changes-en}

\input{sections/en/C_dialogue_complete}

We use one representative configuration from each of 17 independent systems to compare the same query across the constraints and state tracks. Table~\ref{tab:session-turn-en} breaks down the 34 configuration--domain combinations by session type and turn count.

\begin{table}[!htbp]
\caption{Paired AP change by session type and turn count: state minus the corresponding constraints-track result. The 17 representatives use UME-R1 Gen and Embed-RL-2B/4B + T-CoT, giving 34 configuration--domain combinations with four session groups each. $^\dagger$ indicates a 95\% query-bootstrap interval excluding zero. Natural and conflict contain different queries and are not paired against each other. The main-text 39/40 count uses all 20 configurations and has a different denominator.}
\label{tab:session-turn-en}
\setlength{\tabcolsep}{3pt}
\begin{center}
\input{tables/app_session_turn_stats.tex}
\end{center}
\end{table}

Revision sessions show more consistent degradation: all 34 combinations fall below their paired single-turn results in both two- and three-turn conflict groups. Non-revision sessions vary by model. RzenEmbed-v2 improves slightly in both domains' natural groups, and Embed-RL-4B + T-CoT gains 0.040--0.044 mAP in Fashion; however, WeMM-9B still loses 0.026--0.029 in Cars, and VLM2Vec-V2 and TWN also show appreciable losses. Revisions are not the only source of multi-turn cost: models must also integrate valid conditions accumulated in history. More turns do not always increase the loss; RzenEmbed-v2's Fashion conflict change is $-0.106$ for two turns and $-0.060$ for three. History length alone therefore does not characterize multi-turn difficulty.

\begin{figure}[!htbp]
  \centering
  \includegraphics[width=\linewidth]{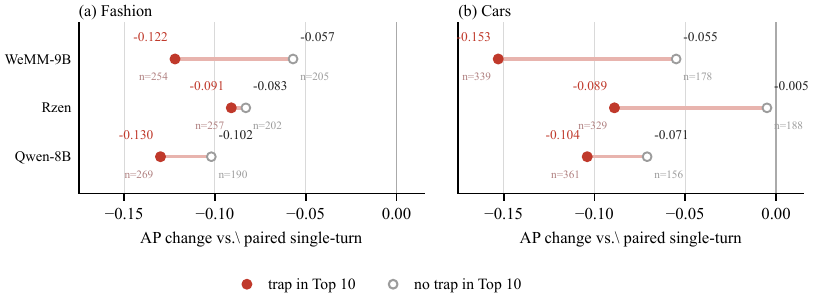}
  \caption{\textbf{Paired AP change in conflict sessions, stratified by whether the obsolete-value trap enters the Top 10 (relative to the paired constraints-track single-turn task).} Only the three strongest configurations are shown; red dots are the trap-in-Top-10 stratum and hollow grey dots the trap-free stratum, values are the stratum's mean AP change, and the small numbers below give the session count of each stratum (459 and 517 trap-constructed sessions in Fashion and Cars, respectively). The two strata are descriptive and not a randomized treatment comparison.}
  \label{fig:trap-strata-en}
\end{figure}

Figure~\ref{fig:trap-strata-en} relates obsolete-value interference to degradation. For the selected WeMM-9B, RzenEmbed-v2, and Qwen3-VL-Embedding-8B configurations, queries with a trap in the Top 10 degrade more in both domains than queries without one. On Cars, the two groups' AP changes are $-0.153$ and $-0.055$ for WeMM-9B, and $-0.089$ and $-0.005$ for RzenEmbed-v2. The gaps are smaller in Fashion, especially for RzenEmbed-v2. Beyond integrating valid conditions, models need to prevent superseded values from continuing to influence ranking.

The constraints- and state-track pools overlap highly but are not identical, so these paired differences describe overall single-to-multi-turn task changes; same-pool history-input interventions are in Appendix~\ref{app:history-revisions-en}.

\paragraph{Single- versus multi-turn retrieval with a shared candidate set.}
To examine the effect of candidate-set changes on multi-turn degradation, we rescore the complete single-turn requests for all 20 configurations against the corresponding state-task candidate sets and compare them with the multi-turn results from the main evaluation. Fashion and Cars each contain 1,000 paired queries, with unchanged positive sets and AP denominators. Rescored single-turn mAP is 0.0047 lower on average than in the main evaluation. With shared candidate sets, multi-turn retrieval remains worse in 38 of 40 configuration--domain comparisons, with an average difference of $-0.0410$ mAP. Thus, the overall multi-turn degradation persists after accounting for candidate-set changes.

\input{sections/en/C_reasoning_results}

\input{sections/en/C_hierarchy_sensitivity}

%% file: sections/en/C_top1_results.tex
\paragraph{Top-1 labeled-positive hit rate.} Top-1 hit rate is the fraction of queries whose first-ranked candidate belongs to the existing labeled positive set, with 1,000 queries per main subset. It measures first-result retrieval under the benchmark's relevance labels, not end-to-end agentic task success. Table~\ref{tab:top1-results-en} reports all 20 configurations. For example, WeMM scores 0.464 and 0.318 on the Fashion and Cars constraints tasks, and 0.411 and 0.249 on the corresponding state tasks.

\begin{table}[!htbp]
\small\setlength{\tabcolsep}{3pt}\renewcommand{\arraystretch}{1.08}
\caption{Top-1 labeled-positive hit rate on the six main subsets (higher is better). Each value is a proportion over 1,000 queries; configurations and row order follow Table~\ref{tab:main-results-en}.}
\label{tab:top1-results-en}
\begin{center}
\input{tables/app_top1_en.tex}
\end{center}
\end{table}

%% file: tables/app_top1_en.tex
\begin{tabular*}{\linewidth}{@{\extracolsep{\fill}}lrrrrrr@{}}
\toprule
Model & \multicolumn{2}{c}{Abstraction} & \multicolumn{2}{c}{Constraints} & \multicolumn{2}{c}{State} \\
\cmidrule(lr){2-3}\cmidrule(lr){4-5}\cmidrule(lr){6-7}
 & I2C & C2I & Fashion & Cars & Fashion & Cars \\
\midrule
CLIP ViT-L/14 & 0.318 & 0.281 & 0.090 & 0.016 & 0.092 & 0.014 \\
SigLIP-2-so400m & 0.233 & 0.381 & 0.100 & 0.035 & 0.068 & 0.022 \\
jina-clip-v2 & 0.328 & 0.467 & 0.071 & 0.017 & 0.063 & 0.017 \\
WeMM-Embedding-9B & 0.490 & 0.687 & 0.464 & 0.318 & 0.411 & 0.249 \\
E5-Omni-7B & 0.341 & 0.627 & 0.246 & 0.115 & 0.110 & 0.035 \\
RzenEmbed-v2 & 0.540 & 0.661 & 0.403 & 0.240 & 0.358 & 0.210 \\
Qwen3-VL-Emb.-8B & 0.475 & 0.686 & 0.427 & 0.171 & 0.341 & 0.108 \\
Qwen3-VL-Emb.-2B & 0.438 & 0.588 & 0.342 & 0.086 & 0.092 & 0.025 \\
OmniEmbed-3B & 0.385 & 0.584 & 0.095 & 0.023 & 0.082 & 0.020 \\
VLM2Vec-V2.0-2B & 0.296 & 0.468 & 0.210 & 0.118 & 0.155 & 0.088 \\
VLM2Vec-2B & 0.257 & 0.454 & 0.115 & 0.040 & 0.100 & 0.046 \\
VLM2Vec-7B & 0.360 & 0.493 & 0.172 & 0.083 & 0.150 & 0.072 \\
Embed-RL-2B (no CoT) & 0.416 & 0.541 & 0.254 & 0.124 & 0.206 & 0.101 \\
Embed-RL-4B (no CoT) & 0.494 & 0.617 & 0.281 & 0.176 & 0.233 & 0.128 \\
B-Tokens-2B & 0.430 & 0.545 & 0.263 & 0.083 & 0.183 & 0.068 \\
UME-R1-7B (Disc) & 0.510 & 0.636 & 0.347 & 0.168 & 0.213 & 0.114 \\
UME-R1-7B (Gen) & 0.509 & 0.649 & 0.259 & 0.089 & 0.132 & 0.051 \\
TWN-4B & 0.504 & 0.614 & 0.151 & 0.064 & 0.093 & 0.025 \\
Embed-RL-2B + T-CoT & 0.477 & 0.614 & 0.317 & 0.167 & 0.274 & 0.133 \\
Embed-RL-4B + T-CoT & 0.512 & 0.649 & 0.281 & 0.229 & 0.298 & 0.199 \\
\bottomrule
\end{tabular*}

%% file: sections/en/C_self_same_results.tex
\subsection{Results on the Self and Same Reference Tasks}
\label{app:self-same-results-en}

Self retrieves an instance's own leaf concept, or images of that concept in the reverse direction; Same retrieves images of the same reference item without requiring attribute changes. Table~\ref{tab:self-same-results-en} reports both tasks for the same 20 configurations as the main table. Their independently constructed candidate sets provide basic-matching references rather than performance upper bounds for the main tasks.

\begin{table}[!htbp]
\small\setlength{\tabcolsep}{7pt}\renewcommand{\arraystretch}{1.08}
\caption{mAP on Self and Same (higher is better). Self contains 1,000 queries per direction and Same contains 500 per domain, each with 1,000 candidates. Self has one positive per query, so its mAP equals MRR. Configurations and row order follow Table~\ref{tab:main-results-en}; bold marks the best result in each column.}
\label{tab:self-same-results-en}
\begin{center}
\input{tables/app_self_same_en.tex}
\end{center}
\end{table}

All 20 configurations score higher on both Self directions than on their corresponding cross-level main tasks. The leading model on Same varies by domain: jina-CLIP-v2 leads on Fashion, while RzenEmbed-v2 leads on Cars. Basic category matching and same-item retrieval thus yield different model rankings; further comparisons with cross-level and multi-constraint retrieval appear in Section~\ref{sec:diagnosis-en}.

%% file: tables/app_self_same_en.tex
\begin{tabularx}{\linewidth}{@{}Xrrrr@{}}
\toprule
Configuration & \multicolumn{2}{c}{Self} & \multicolumn{2}{c}{Same} \\
\cmidrule(lr){2-3}\cmidrule(lr){4-5}
 & I2C & C2I & Fashion & Cars \\
\midrule
CLIP-L/14 & 0.372 & 0.374 & 0.246 & 0.265 \\
SigLIP-2 & 0.275 & 0.429 & 0.184 & 0.311 \\
jina-clip-v2 & 0.550 & 0.506 & \textbf{0.426} & 0.381 \\
WeMM-9B & 0.801 & \textbf{0.809} & 0.353 & 0.560 \\
E5-Omni-7B & 0.641 & 0.738 & 0.322 & 0.465 \\
Rzen-v2 & 0.780 & 0.768 & 0.362 & \textbf{0.682} \\
Qwen3-VL-8B & \textbf{0.821} & 0.788 & 0.341 & 0.522 \\
Qwen3-VL-2B & 0.737 & 0.692 & 0.381 & 0.549 \\
OmniEmbed-3B & 0.667 & 0.655 & 0.333 & 0.342 \\
VLM2Vec-V2-2B & 0.564 & 0.552 & 0.212 & 0.488 \\
VLM2Vec-2B & 0.503 & 0.485 & 0.232 & 0.455 \\
VLM2Vec-7B & 0.660 & 0.593 & 0.057 & 0.236 \\
Embed-RL-2B (no CoT) & 0.681 & 0.642 & 0.391 & 0.458 \\
Embed-RL-4B (no CoT) & 0.783 & 0.690 & 0.398 & 0.394 \\
B-Tokens-2B & 0.710 & 0.665 & 0.339 & 0.515 \\
UME-R1-7B (Disc) & 0.692 & 0.737 & 0.385 & 0.559 \\
UME-R1-7B (Gen) & 0.702 & 0.741 & 0.393 & 0.561 \\
TWN-4B & 0.746 & 0.686 & 0.357 & 0.473 \\
Embed-RL-2B + T-CoT & 0.766 & 0.702 & 0.374 & 0.472 \\
Embed-RL-4B + T-CoT & 0.817 & 0.748 & 0.379 & 0.440 \\
\bottomrule
\end{tabularx}

%% file: tables/app_track_b_stats.tex
\begin{tabular}{@{}lrrrrrrrrrrrrrr@{}}
\toprule
& \multicolumn{7}{c}{Fashion} & \multicolumn{7}{c}{Cars} \\
\cmidrule(lr){2-8}\cmidrule(l){9-15}
Model & D1 & D2 & D3 & D4 & D5 & D6 & D7 & D1 & D2 & D3 & D4 & D5 & D6 & D7 \\
\midrule
Rzen & 0.68 & 0.49 & 0.62 & 0.25 & 0.45 & 0.50 & 0.23 & 0.45 & 0.27 & 0.47 & 0.08 & 0.27 & 0.30 & 0.12 \\
Qwen-8B & 0.69 & 0.50 & 0.66 & 0.31 & 0.46 & 0.51 & 0.32 & 0.38 & 0.22 & 0.34 & 0.08 & 0.20 & 0.20 & 0.11 \\
Qwen-2B & 0.70 & 0.41 & 0.62 & 0.19 & 0.41 & 0.42 & 0.19 & 0.32 & 0.16 & 0.18 & 0.06 & 0.13 & 0.13 & 0.07 \\
OmniEmbed-3B & 0.30 & 0.28 & 0.17 & 0.15 & 0.17 & 0.10 & 0.09 & 0.18 & 0.11 & 0.07 & 0.04 & 0.07 & 0.05 & 0.04 \\
VLM2Vec-V2 & 0.49 & 0.34 & 0.40 & 0.21 & 0.28 & 0.29 & 0.14 & 0.32 & 0.15 & 0.28 & 0.06 & 0.15 & 0.19 & 0.07 \\
VLM2Vec-2B & 0.30 & 0.28 & 0.20 & 0.19 & 0.21 & 0.16 & 0.10 & 0.19 & 0.13 & 0.12 & 0.06 & 0.08 & 0.07 & 0.03 \\
VLM2Vec-7B & 0.43 & 0.21 & 0.36 & 0.15 & 0.22 & 0.22 & 0.11 & 0.24 & 0.09 & 0.20 & 0.05 & 0.10 & 0.12 & 0.05 \\
Embed-RL-2B & 0.59 & 0.40 & 0.55 & 0.22 & 0.36 & 0.40 & 0.18 & 0.37 & 0.16 & 0.36 & 0.07 & 0.18 & 0.24 & 0.08 \\
Embed-RL-4B & 0.54 & 0.35 & 0.44 & 0.20 & 0.33 & 0.36 & 0.22 & 0.40 & 0.21 & 0.44 & 0.08 & 0.27 & 0.31 & 0.15 \\
UME-R1 Gen & 0.58 & 0.40 & 0.46 & 0.20 & 0.30 & 0.32 & 0.14 & 0.32 & 0.17 & 0.28 & 0.05 & 0.13 & 0.15 & 0.05 \\
B-Tokens & 0.58 & 0.33 & 0.49 & 0.18 & 0.33 & 0.39 & 0.18 & 0.32 & 0.14 & 0.23 & 0.06 & 0.15 & 0.16 & 0.06 \\
CLIP ViT-L/14 & 0.24 & 0.26 & 0.16 & 0.18 & 0.18 & 0.15 & 0.10 & 0.14 & 0.10 & 0.04 & 0.04 & 0.06 & 0.03 & 0.03 \\
SigLIP-2 & 0.32 & 0.20 & 0.16 & 0.14 & 0.14 & 0.14 & 0.07 & 0.19 & 0.10 & 0.07 & 0.04 & 0.07 & 0.04 & 0.03 \\
jina-clip-v2 & 0.17 & 0.28 & 0.09 & 0.09 & 0.12 & 0.04 & 0.04 & 0.11 & 0.09 & 0.03 & 0.03 & 0.04 & 0.02 & 0.02 \\
TWN-4B & 0.43 & 0.37 & 0.29 & 0.18 & 0.23 & 0.18 & 0.10 & 0.31 & 0.14 & 0.18 & 0.06 & 0.11 & 0.08 & 0.05 \\
WeMM-9B & 0.70 & 0.52 & 0.67 & 0.34 & 0.52 & 0.57 & 0.40 & 0.52 & 0.33 & 0.51 & 0.17 & 0.34 & 0.40 & 0.17 \\
E5-Omni-7B & 0.62 & 0.37 & 0.48 & 0.20 & 0.32 & 0.31 & 0.16 & 0.31 & 0.17 & 0.30 & 0.06 & 0.14 & 0.18 & 0.06 \\
\bottomrule
\end{tabular}

%% file: sections/en/C_constraint_controls.tex
\paragraph{Constraint count and cross-domain aggregation.} Figure~\ref{fig:constraints-en}a groups D1/D2, D3/D4/D5, and D6/D7 into one, two, and three constraints, comprising 300, 450, and 250 queries per domain, respectively. For each configuration, mAP is first query-weighted within each count group and then averaged equally over Fashion and Cars. Performance decreases at both successive count increases for all 17 configurations; the configuration medians are 0.341, 0.246, and 0.182. Figure~\ref{fig:constraints-en}b uses the same domain averaging for D1/D2 (1P/1N) and D3/D4 (2P/2N); negative-only groups score lower for 16 of 17 configurations in each comparison. Since polarity and attribute composition differ across count groups, the count curves describe difficulty across the constructed query groups rather than a paired intervention on count alone.

\paragraph{Count-matched and composition-standardized comparisons.}
To examine condition-count and attribute-composition differences, we compare D1/D2 (1P/1N), D3/D4 (2P/2N), and D3/D5 (2P/1P1N). In addition to the complete groups, we form shared strata by product category and explicit attribute-name combination, weighting each stratum by the smaller of its two group counts to equalize category--attribute composition. Negation-bearing groups score lower in 32/34, 32/34, and 28/34 combinations, respectively; after standardization, the counts are 31/34, 30/34, and 28/34. Reference items, attribute values, and candidate composition can still differ, so these comparisons assess the robustness of the observed pattern rather than isolate a causal effect of negation wording.

\begin{table}[!htbp]
\footnotesize\setlength{\tabcolsep}{3pt}\renewcommand{\arraystretch}{1.12}
\caption{Polarity comparisons with matched condition counts and category/attribute standardization. Differences are right minus left. We report the number of negative differences among 34 configuration--domain combinations, their median, and their range. No query-paired or significance claim is made.}
\label{tab:negation-controls-en}
\begin{center}
\input{tables/app_negation_controls_en.tex}
\end{center}
\end{table}

%% file: tables/app_negation_controls_en.tex
\begin{tabular*}{\linewidth}{@{}l@{\extracolsep{\fill}}lcrr@{}}
\toprule
Control & Left / right & Right lower & Median $\Delta$ & Range of $\Delta$ \\
\midrule
Count-matched & 1P / 1N & 32/34 & $-0.165$ & $[-0.285, 0.110]$ \\
 & 2P / 2N & 32/34 & $-0.213$ & $[-0.423, 0.021]$ \\
 & 2P / 1P1N & 28/34 & $-0.123$ & $[-0.211, 0.031]$ \\
\midrule
Count + category/attributes & 1P / 1N & 31/34 & $-0.206$ & $[-0.326, 0.066]$ \\
 & 2P / 2N & 30/34 & $-0.217$ & $[-0.457, 0.035]$ \\
 & 2P / 1P1N & 28/34 & $-0.131$ & $[-0.265, 0.033]$ \\
\bottomrule
\end{tabular*}

%% file: sections/en/C_dialogue_complete.tex
Table~\ref{tab:dialogue-complete-en} reports single-turn and multi-turn mAP and absolute differences for all 20 main-table configurations, covering 40 configuration--domain comparisons. $\Delta\mathrm{mAP}=\mathrm{mAP}_{\mathrm{multi}}-\mathrm{mAP}_{\mathrm{single}}$, computed from unrounded means.

\begin{table}[!htbp]
\footnotesize\setlength{\tabcolsep}{4pt}\renewcommand{\arraystretch}{1.08}
\caption{Single-to-multi-turn changes for all 20 configurations. Single and multi denote the constraints and state tracks, respectively, with 1,000 paired queries per configuration--domain comparison. $\Delta$ mAP is multi-turn minus single-turn, rounded to three decimals. B/C candidate pools overlap highly but are not identical.}
\label{tab:dialogue-complete-en}
\begin{center}
\input{tables/app_dialogue_complete_en.tex}
\end{center}
\end{table}

%% file: tables/app_dialogue_complete_en.tex
\begin{tabularx}{\linewidth}{@{}Xrrrrrr@{}}
\toprule
Configuration & \multicolumn{3}{c}{Fashion} & \multicolumn{3}{c}{Cars} \\
\cmidrule(lr){2-4}\cmidrule(lr){5-7}
 & Single & Multi & $\Delta$ mAP & Single & Multi & $\Delta$ mAP \\
\midrule
CLIP-L/14 & 0.185 & 0.180 & \textcolor[HTML]{B85C43}{$-0.005$} & 0.065 & 0.062 & \textcolor[HTML]{B85C43}{$-0.002$} \\
SigLIP-2 & 0.173 & 0.144 & \textcolor[HTML]{B85C43}{$-0.029$} & 0.082 & 0.069 & \textcolor[HTML]{B85C43}{$-0.013$} \\
jina-clip-v2 & 0.123 & 0.117 & \textcolor[HTML]{B85C43}{$-0.006$} & 0.052 & 0.050 & \textcolor[HTML]{B85C43}{$-0.002$} \\
WeMM-9B & 0.547 & 0.498 & \textcolor[HTML]{B85C43}{$-0.049$} & 0.367 & 0.291 & \textcolor[HTML]{B85C43}{$-0.076$} \\
E5-Omni-7B & 0.364 & 0.223 & \textcolor[HTML]{B85C43}{$-0.142$} & 0.186 & 0.101 & \textcolor[HTML]{B85C43}{$-0.084$} \\
Rzen-v2 & 0.482 & 0.441 & \textcolor[HTML]{B85C43}{$-0.041$} & 0.298 & 0.272 & \textcolor[HTML]{B85C43}{$-0.026$} \\
Qwen3-VL-8B & 0.508 & 0.454 & \textcolor[HTML]{B85C43}{$-0.054$} & 0.230 & 0.177 & \textcolor[HTML]{B85C43}{$-0.053$} \\
Qwen3-VL-2B & 0.442 & 0.212 & \textcolor[HTML]{B85C43}{$-0.230$} & 0.157 & 0.085 & \textcolor[HTML]{B85C43}{$-0.072$} \\
OmniEmbed-3B & 0.184 & 0.167 & \textcolor[HTML]{B85C43}{$-0.017$} & 0.084 & 0.076 & \textcolor[HTML]{B85C43}{$-0.008$} \\
VLM2Vec-V2-2B & 0.318 & 0.265 & \textcolor[HTML]{B85C43}{$-0.053$} & 0.182 & 0.142 & \textcolor[HTML]{B85C43}{$-0.040$} \\
VLM2Vec-2B & 0.213 & 0.191 & \textcolor[HTML]{B85C43}{$-0.022$} & 0.101 & 0.092 & \textcolor[HTML]{B85C43}{$-0.009$} \\
VLM2Vec-7B & 0.251 & 0.221 & \textcolor[HTML]{B85C43}{$-0.031$} & 0.128 & 0.107 & \textcolor[HTML]{B85C43}{$-0.020$} \\
Embed-RL-2B (no CoT) & 0.355 & 0.308 & \textcolor[HTML]{B85C43}{$-0.047$} & 0.198 & 0.170 & \textcolor[HTML]{B85C43}{$-0.028$} \\
Embed-RL-4B (no CoT) & 0.356 & 0.321 & \textcolor[HTML]{B85C43}{$-0.035$} & 0.227 & 0.194 & \textcolor[HTML]{B85C43}{$-0.032$} \\
B-Tokens-2B & 0.368 & 0.300 & \textcolor[HTML]{B85C43}{$-0.068$} & 0.169 & 0.146 & \textcolor[HTML]{B85C43}{$-0.024$} \\
UME-R1-7B (Disc) & 0.436 & 0.321 & \textcolor[HTML]{B85C43}{$-0.115$} & 0.249 & 0.195 & \textcolor[HTML]{B85C43}{$-0.054$} \\
UME-R1-7B (Gen) & 0.359 & 0.235 & \textcolor[HTML]{B85C43}{$-0.124$} & 0.174 & 0.130 & \textcolor[HTML]{B85C43}{$-0.045$} \\
TWN-4B & 0.264 & 0.201 & \textcolor[HTML]{B85C43}{$-0.063$} & 0.140 & 0.092 & \textcolor[HTML]{B85C43}{$-0.048$} \\
Embed-RL-2B + T-CoT & 0.403 & 0.374 & \textcolor[HTML]{B85C43}{$-0.030$} & 0.221 & 0.195 & \textcolor[HTML]{B85C43}{$-0.026$} \\
Embed-RL-4B + T-CoT & 0.360 & 0.376 & \textcolor[HTML]{247A83}{$+0.016$} & 0.281 & 0.259 & \textcolor[HTML]{B85C43}{$-0.022$} \\
\bottomrule
\end{tabularx}

%% file: tables/app_session_turn_stats.tex
\begin{tabular}{@{}llrrrr@{}}
\toprule
Model & Domain & Natural-2 & Natural-3 & Conflict-2 & Conflict-3 \\
\midrule
Rzen & Fashion & +0.016 & +0.003 & -0.106$^\dagger$ & -0.060$^\dagger$ \\
Rzen & Cars & +0.011 & +0.008 & -0.063$^\dagger$ & -0.051$^\dagger$ \\
Qwen-8B & Fashion & +0.012 & +0.002 & -0.101$^\dagger$ & -0.117$^\dagger$ \\
Qwen-8B & Cars & -0.011 & -0.007 & -0.091$^\dagger$ & -0.098$^\dagger$ \\
Qwen-2B & Fashion & -0.208$^\dagger$ & -0.204$^\dagger$ & -0.257$^\dagger$ & -0.245$^\dagger$ \\
Qwen-2B & Cars & -0.061$^\dagger$ & -0.061$^\dagger$ & -0.089$^\dagger$ & -0.075$^\dagger$ \\
OmniEmbed-3B & Fashion & -0.008$^\dagger$ & -0.006$^\dagger$ & -0.034$^\dagger$ & -0.016$^\dagger$ \\
OmniEmbed-3B & Cars & +0.004 & -0.001 & -0.020$^\dagger$ & -0.009$^\dagger$ \\
VLM2Vec-V2 & Fashion & -0.025 & -0.047$^\dagger$ & -0.088$^\dagger$ & -0.038$^\dagger$ \\
VLM2Vec-V2 & Cars & -0.011 & -0.049$^\dagger$ & -0.044$^\dagger$ & -0.054$^\dagger$ \\
VLM2Vec-2B & Fashion & -0.007 & +0.011 & -0.038$^\dagger$ & -0.050$^\dagger$ \\
VLM2Vec-2B & Cars & +0.004 & -0.005 & -0.023$^\dagger$ & -0.007 \\
VLM2Vec-7B & Fashion & -0.019 & -0.041$^\dagger$ & -0.027 & -0.035 \\
VLM2Vec-7B & Cars & -0.010 & -0.023$^\dagger$ & -0.024$^\dagger$ & -0.021 \\
Embed-RL-2B + T-CoT & Fashion & -0.002 & -0.006 & -0.056$^\dagger$ & -0.048$^\dagger$ \\
Embed-RL-2B + T-CoT & Cars & +0.002 & -0.011 & -0.032$^\dagger$ & -0.063$^\dagger$ \\
Embed-RL-4B + T-CoT & Fashion & +0.044$^\dagger$ & +0.040$^\dagger$ & -0.010 & -0.005 \\
Embed-RL-4B + T-CoT & Cars & +0.020$^\dagger$ & -0.016 & -0.029$^\dagger$ & -0.061$^\dagger$ \\
UME-R1 Gen & Fashion & -0.073$^\dagger$ & -0.091$^\dagger$ & -0.167$^\dagger$ & -0.154$^\dagger$ \\
UME-R1 Gen & Cars & -0.007 & -0.017$^\dagger$ & -0.071$^\dagger$ & -0.077$^\dagger$ \\
B-Tokens & Fashion & -0.046$^\dagger$ & -0.075$^\dagger$ & -0.077$^\dagger$ & -0.069$^\dagger$ \\
B-Tokens & Cars & +0.007 & -0.016$^\dagger$ & -0.037$^\dagger$ & -0.046$^\dagger$ \\
CLIP ViT-L/14 & Fashion & +0.011 & -0.009 & -0.005 & -0.017$^\dagger$ \\
CLIP ViT-L/14 & Cars & +0.001 & -0.001 & -0.007$^\dagger$ & -0.001 \\
SigLIP-2 & Fashion & -0.016 & -0.026$^\dagger$ & -0.048$^\dagger$ & -0.019 \\
SigLIP-2 & Cars & -0.006 & -0.004 & -0.021$^\dagger$ & -0.018$^\dagger$ \\
jina-clip-v2 & Fashion & +0.002 & -0.002$^\dagger$ & -0.016$^\dagger$ & -0.007$^\dagger$ \\
jina-clip-v2 & Cars & $-2.7\times 10^{-6}$ & -0.0001$^\dagger$ & -0.005$^\dagger$ & -0.001$^\dagger$ \\
TWN-4B & Fashion & -0.043$^\dagger$ & -0.030$^\dagger$ & -0.104$^\dagger$ & -0.065$^\dagger$ \\
TWN-4B & Cars & -0.031$^\dagger$ & -0.019$^\dagger$ & -0.081$^\dagger$ & -0.054$^\dagger$ \\
WeMM-9B & Fashion & +0.023$^\dagger$ & -0.028 & -0.082$^\dagger$ & -0.101$^\dagger$ \\
WeMM-9B & Cars & -0.026$^\dagger$ & -0.029$^\dagger$ & -0.134$^\dagger$ & -0.101$^\dagger$ \\
E5-Omni-7B & Fashion & -0.124$^\dagger$ & -0.115$^\dagger$ & -0.165$^\dagger$ & -0.158$^\dagger$ \\
E5-Omni-7B & Cars & -0.075$^\dagger$ & -0.069$^\dagger$ & -0.094$^\dagger$ & -0.098$^\dagger$ \\
\bottomrule
\end{tabular}

%% file: sections/en/C_reasoning_results.tex
\FloatBarrier
\subsection{Complete Reasoning Comparisons}
\label{app:reasoning-results-en}

Table~\ref{tab:reasoning-complete-en} extends the reasoning comparisons to all six main subsets and four Self/Same references. UME-R1 compares query-side Gen with Disc; Embed-RL-2B/4B compare two-sided +T-CoT with no CoT, holding each encoder's weights fixed. Holm correction is applied separately over the ten tasks in each column, consistent with the paired analysis in the main text.

\begin{papertable}
\small\setlength{\tabcolsep}{5pt}\renewcommand{\arraystretch}{1.12}
\caption{Reasoning-minus-non-reasoning $\Delta$ mAP, with 95\% intervals from 3,000 query-bootstrap resamples in brackets. $^{\dagger}$ marks $p<0.05$ for the two-sided Wilcoxon test after Holm correction over ten tasks per column. Each main subset and Self direction contains 1,000 queries; Same contains 500 per domain. B/C denote the constraints and state tracks.}
\label{tab:reasoning-complete-en}
\begin{center}
\input{tables/app_reasoning_complete.tex}
\end{center}
\end{papertable}

%% file: tables/app_reasoning_complete.tex
\begin{tabularx}{\linewidth}{@{}Xrrr@{}}
\toprule
Task & UME-R1-7B & Embed-RL-2B & Embed-RL-4B \\
\midrule
I2C & \shortstack[r]{\textcolor[HTML]{B85C43}{$-0.001$}\\{}[-0.006,+0.004]} & \shortstack[r]{\textcolor[HTML]{247A83}{$+0.056^{\dagger}$}\\{}[+0.045,+0.067]} & \shortstack[r]{\textcolor[HTML]{247A83}{$+0.020^{\dagger}$}\\{}[+0.010,+0.030]} \\
C2I & \shortstack[r]{\textcolor[HTML]{247A83}{$+0.011^{\dagger}$}\\{}[+0.007,+0.016]} & \shortstack[r]{\textcolor[HTML]{247A83}{$+0.053^{\dagger}$}\\{}[+0.045,+0.061]} & \shortstack[r]{\textcolor[HTML]{247A83}{$+0.049^{\dagger}$}\\{}[+0.041,+0.058]} \\
B--Fashion & \shortstack[r]{\textcolor[HTML]{B85C43}{$-0.077^{\dagger}$}\\{}[-0.090,-0.066]} & \shortstack[r]{\textcolor[HTML]{247A83}{$+0.048^{\dagger}$}\\{}[+0.035,+0.060]} & \shortstack[r]{\textcolor[HTML]{247A83}{$+0.004$}\\{}[-0.012,+0.020]} \\
B--Cars & \shortstack[r]{\textcolor[HTML]{B85C43}{$-0.075^{\dagger}$}\\{}[-0.084,-0.066]} & \shortstack[r]{\textcolor[HTML]{247A83}{$+0.023^{\dagger}$}\\{}[+0.015,+0.032]} & \shortstack[r]{\textcolor[HTML]{247A83}{$+0.055^{\dagger}$}\\{}[+0.043,+0.066]} \\
C--Fashion & \shortstack[r]{\textcolor[HTML]{B85C43}{$-0.086^{\dagger}$}\\{}[-0.098,-0.074]} & \shortstack[r]{\textcolor[HTML]{247A83}{$+0.065^{\dagger}$}\\{}[+0.054,+0.077]} & \shortstack[r]{\textcolor[HTML]{247A83}{$+0.055^{\dagger}$}\\{}[+0.040,+0.070]} \\
C--Cars & \shortstack[r]{\textcolor[HTML]{B85C43}{$-0.066^{\dagger}$}\\{}[-0.075,-0.057]} & \shortstack[r]{\textcolor[HTML]{247A83}{$+0.025^{\dagger}$}\\{}[+0.018,+0.034]} & \shortstack[r]{\textcolor[HTML]{247A83}{$+0.065^{\dagger}$}\\{}[+0.054,+0.075]} \\
Self--I2C & \shortstack[r]{\textcolor[HTML]{247A83}{$+0.010^{\dagger}$}\\{}[+0.003,+0.018]} & \shortstack[r]{\textcolor[HTML]{247A83}{$+0.086^{\dagger}$}\\{}[+0.070,+0.102]} & \shortstack[r]{\textcolor[HTML]{247A83}{$+0.034^{\dagger}$}\\{}[+0.021,+0.047]} \\
Self--C2I & \shortstack[r]{\textcolor[HTML]{247A83}{$+0.004$}\\{}[-0.004,+0.012]} & \shortstack[r]{\textcolor[HTML]{247A83}{$+0.061^{\dagger}$}\\{}[+0.046,+0.076]} & \shortstack[r]{\textcolor[HTML]{247A83}{$+0.058^{\dagger}$}\\{}[+0.043,+0.072]} \\
Same--Fashion & \shortstack[r]{\textcolor[HTML]{247A83}{$+0.009^{\dagger}$}\\{}[+0.001,+0.016]} & \shortstack[r]{\textcolor[HTML]{B85C43}{$-0.017^{\dagger}$}\\{}[-0.030,-0.005]} & \shortstack[r]{\textcolor[HTML]{B85C43}{$-0.019^{\dagger}$}\\{}[-0.033,-0.005]} \\
Same--Cars & \shortstack[r]{\textcolor[HTML]{247A83}{$+0.003$}\\{}[-0.002,+0.008]} & \shortstack[r]{\textcolor[HTML]{247A83}{$+0.015^{\dagger}$}\\{}[+0.004,+0.025]} & \shortstack[r]{\textcolor[HTML]{247A83}{$+0.046^{\dagger}$}\\{}[+0.035,+0.057]} \\
\bottomrule
\end{tabularx}

%% file: sections/en/C_hierarchy_sensitivity.tex
\FloatBarrier
\subsection{Sensitivity of Hierarchy Scoring}
\label{app:hierarchy-scoring-sensitivity-en}

The main task treats one- and two-level ancestors jointly as positives, whereas distance-specific analysis scores each group on the same ranking. Correct ancestors at the other distance can therefore occupy earlier positions. For each distance, we remove labeled positives at the other distance on the 711 shared queries, compress subsequent ranks, and preserve the relative order of all remaining candidates. The two-level-minus-one-level mAP difference remains negative for all 17 configurations; the median configuration-level difference narrows from $-0.314$ to $-0.246$. All 17 paired tests remain significant after Holm correction. Positive occupancy explains part, but not all, of the observed gap. No distance-selecting instruction is introduced, so this analysis does not test switching retrieval targets between one- and two-level requests. Table~\ref{tab:hierarchy-sensitivity-en} reports every configuration; intervals use 3,000 query-bootstrap resamples and differences are two-level minus one-level.

\begin{table}[!htbp]
\footnotesize\setlength{\tabcolsep}{2pt}\renewcommand{\arraystretch}{1.12}
\caption{Hierarchy-scoring sensitivity on the same 711 queries. Original scoring retains positives at the other distance in the ranking; adjusted scoring removes only those labeled positives and compresses subsequent ranks. $\Delta$ is two-level minus one-level mAP; brackets give the adjusted difference's 95\% bootstrap interval. All 17 paired Wilcoxon tests in each scoring family remain significant after Holm correction.}
\label{tab:hierarchy-sensitivity-en}
\begin{center}
\input{tables/app_hierarchy_sensitivity_en.tex}
\end{center}
\end{table}

%% file: tables/app_hierarchy_sensitivity_en.tex
\begin{tabular*}{\linewidth}{@{}l@{\extracolsep{\fill}}rrrrrl@{}}
\toprule
& \multicolumn{3}{c}{Original scoring} & \multicolumn{3}{c}{Other-distance positives removed} \\
\cmidrule(lr){2-4}\cmidrule(lr){5-7}
Configuration & Level 1 & Level 2 & $\Delta$ & Level 1 & Level 2 & $\Delta$ [95\% CI] \\
\midrule
CLIP ViT-L/14 & 0.392 & 0.222 & $-0.171$ & 0.415 & 0.268 & $-0.148\;[-0.181, -0.115]$ \\
SigLIP-2-so400m & 0.324 & 0.125 & $-0.199$ & 0.340 & 0.150 & $-0.191\;[-0.220, -0.161]$ \\
jina-clip-v2 & 0.477 & 0.200 & $-0.277$ & 0.498 & 0.252 & $-0.247\;[-0.277, -0.216]$ \\
WeMM-Embedding-9B & 0.644 & 0.297 & $-0.347$ & 0.684 & 0.415 & $-0.270\;[-0.297, -0.242]$ \\
E5-Omni-7B & 0.484 & 0.170 & $-0.314$ & 0.507 & 0.217 & $-0.290\;[-0.321, -0.259]$ \\
RzenEmbed-v2 & 0.655 & 0.331 & $-0.324$ & 0.699 & 0.463 & $-0.235\;[-0.264, -0.208]$ \\
Qwen3-VL-Emb.-8B & 0.648 & 0.251 & $-0.397$ & 0.672 & 0.363 & $-0.309\;[-0.337, -0.281]$ \\
Qwen3-VL-Emb.-2B & 0.603 & 0.254 & $-0.349$ & 0.636 & 0.348 & $-0.288\;[-0.317, -0.259]$ \\
OmniEmbed-3B & 0.546 & 0.193 & $-0.354$ & 0.569 & 0.252 & $-0.316\;[-0.347, -0.286]$ \\
VLM2Vec-V2.0-2B & 0.427 & 0.198 & $-0.229$ & 0.452 & 0.245 & $-0.208\;[-0.237, -0.179]$ \\
VLM2Vec-2B & 0.374 & 0.206 & $-0.168$ & 0.397 & 0.237 & $-0.160\;[-0.190, -0.129]$ \\
VLM2Vec-7B & 0.502 & 0.249 & $-0.254$ & 0.531 & 0.317 & $-0.214\;[-0.246, -0.184]$ \\
B-Tokens-2B & 0.561 & 0.256 & $-0.305$ & 0.600 & 0.333 & $-0.267\;[-0.298, -0.239]$ \\
UME-R1-7B (Gen) & 0.634 & 0.318 & $-0.316$ & 0.679 & 0.441 & $-0.237\;[-0.266, -0.208]$ \\
TWN-4B & 0.625 & 0.297 & $-0.328$ & 0.661 & 0.414 & $-0.247\;[-0.276, -0.218]$ \\
Embed-RL-2B + T-CoT & 0.608 & 0.302 & $-0.306$ & 0.648 & 0.406 & $-0.242\;[-0.273, -0.212]$ \\
Embed-RL-4B + T-CoT & 0.643 & 0.318 & $-0.325$ & 0.687 & 0.440 & $-0.246\;[-0.275, -0.218]$ \\
\bottomrule
\end{tabular*}

%% file: sections/en/D_supplementary_experiments.tex
\section{Supplementary Experiments}
\label{app:supplementary-experiments-en}

This appendix uses history truncation, single-modality inputs, and image-removal controls to examine the role of inputs in history integration, joint constraints, and cross-level retrieval, and Top-30 reranking to assess recoverable ranking performance and recall limitations. Additional checks examine statistical resolution, query formulation robustness, and training-data exposure. Input ablations use full input as the reference, whereas reranking uses first-stage retrieval; candidate settings and sample scope are specified in each subsection. Evaluation protocols and implementation details appear in Appendix~\ref{app:evaluation-details-en}; supplementary main-result analyses, model abbreviations, and statistical definitions appear in Appendix~\ref{app:main-results-details-en}.

\subsection{Construct-Validity Ablations}
\label{app:completed-validity-en}

We test history integration by retaining only the last turn, image--text contributions to joint constraints through single-modality inputs, and the role of images in cross-level retrieval through image removal. The three ablations use a fixed panel of 11 configurations listed in the corresponding figures and table, distinct from the 20-configuration main evaluation.

\subsubsection{History Integration and Revision Sessions}
\label{app:history-revisions-en}

\begin{figure}[!htbp]
  \centering
  \includegraphics[width=\linewidth]{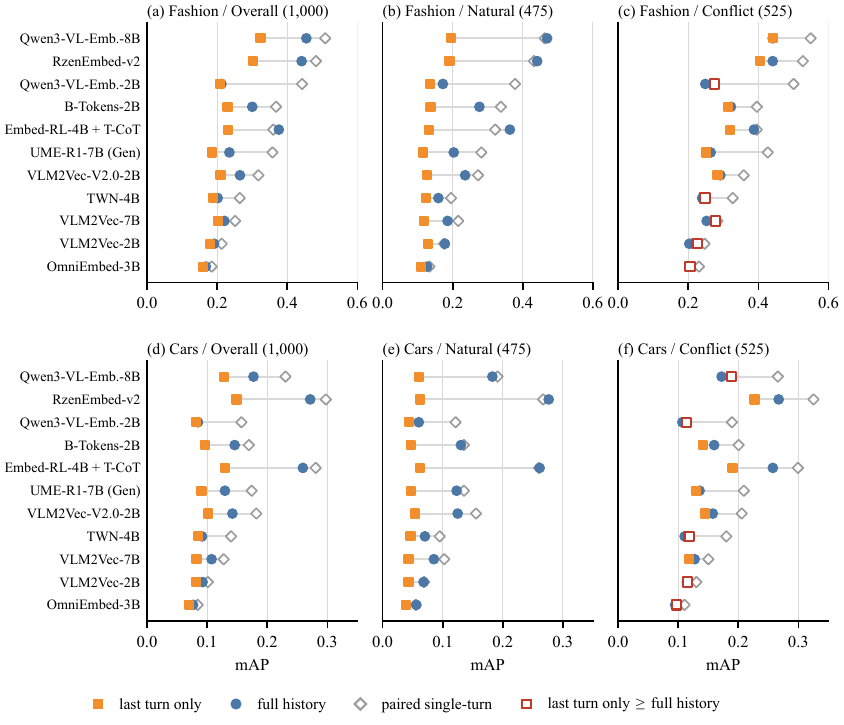}
  \caption{\textbf{Three-point ladder over last-turn-only, full-history, and paired single-turn inputs.} 11 ablation-panel configurations $\times$ 2 domains $\times$ 3 strata; every cell is mAP. The top row is Fashion and the bottom row Cars; the three columns are the overall, natural, and conflict strata, with the query count of each stratum in parentheses. Rows are ordered by the Fashion overall paired single-turn mAP and kept identical across all six panels. Last-turn-only and full-history inputs use exactly the same pool, whereas the paired single-turn reference uses a highly overlapping but non-identical constraints-track pool and is therefore a task-level reference only. Across all 22 cells of the overall and natural columns, last-turn-only is below full history; 10 cells in the conflict column reverse this and are marked with hollow red squares.}
  \label{fig:ladder-en}
\end{figure}

The last-turn-only control preserves the state-track candidate set and representations, replacing only the full query history with the final turn. Figure~\ref{fig:ladder-en} also includes the paired constraints-track single-turn results as a task-level reference. Both overall and within natural sessions, last-turn-only is below full history in all 22 configuration--domain combinations, showing the importance of integrating still-active conditions from history. Last-turn-only and full history use exactly the same pools; the paired single-turn pools overlap highly but are not identical.

Revision sessions show a different result: last-turn-only exceeds full history in 10 of the 22 combinations, marked with hollow red squares in Figure~\ref{fig:ladder-en}(c,f). Some configurations benefit from adding history, while others are hurt by history containing superseded values. History can thus supply information that remains valid while also allowing superseded values to influence ranking, offsetting part of the benefit from integrating history.

The trap-stratified results in Appendix~\ref{app:state-paired-changes-en} provide a complementary observation: in the three displayed configurations, queries degrade more when candidates matching obsolete values enter the Top 10. Together with the last-turn-only control, this suggests that history both supplies information needed for the current request and can interfere through superseded values; multi-turn retrieval requires information integration and state updating.

\FloatBarrier
\subsubsection{Image--Text Ablations}
\label{app:modal-ablation-en}

The single-modality ablation evaluates 11 configurations on the constraints and state tracks in both domains (44 cells, 1,000 queries). Text-only keeps the condition or dialogue text but removes the reference image and its metadata; \texttt{text+refmeta} restores textual reference metadata without the image; image-only keeps the reference image and drops the text. Models needing an image on the text path use the recorded gray-image control.

\begin{figure}[!htbp]
  \centering
  \includegraphics[width=\linewidth]{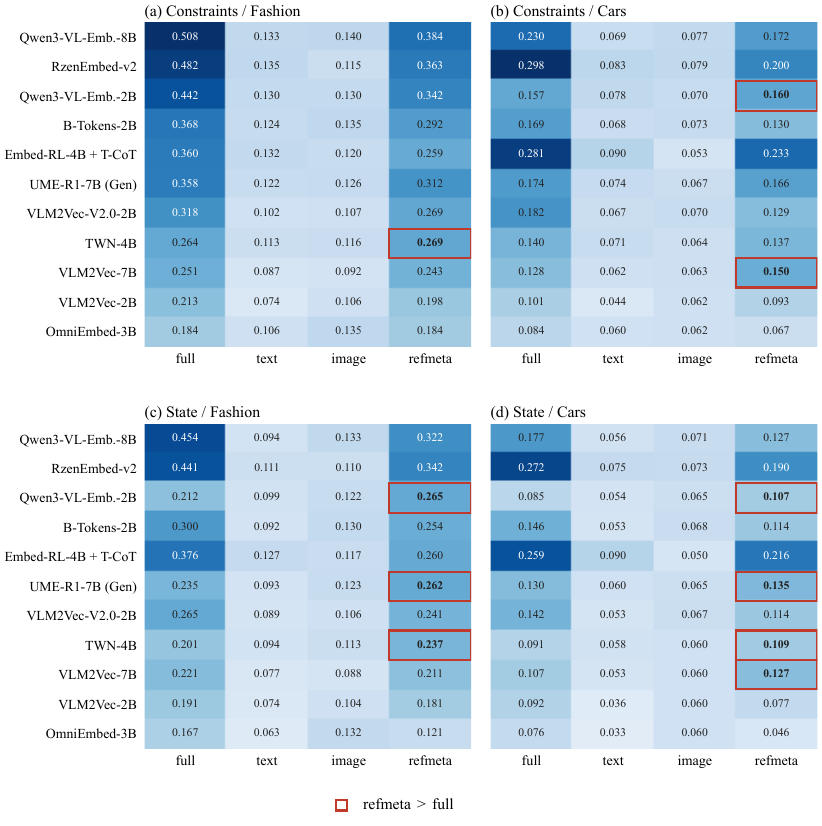}
  \caption{\textbf{Complete panel of the query-side single-modality ablation.} 11 ablation-panel configurations $\times$ 2 tracks $\times$ 2 domains $\times$ 4 input variants, i.e.\ 176 cells of 1,000 queries each. ``full'' is the complete input, ``text'' keeps only the condition/dialogue text, ``image'' keeps only the reference image, and ``refmeta'' removes the image but restores the reference product text attributes; candidate-side inputs are unchanged throughout. Cells show mAP; color scales are normalized within each domain and are not used for cross-domain comparisons. Red boxes mark the 10 cells where refmeta exceeds full. All four panels share the row order of (a), sorted by its full values, so a configuration's full rank within the other panels may differ. Models that must receive an image also have a gray-image control, which is not shown here.}
  \label{fig:e2-ablation-en}
\end{figure}

Figure~\ref{fig:e2-ablation-en} reports per-cell mAP for all 11 configurations under the four input variants, and three phenomena complement one another. First, no single modality sustains high-level retrieval: text-only and image-only are below full in all 44 cells; for example, Qwen3-VL-Embedding-8B, RzenEmbed-v2, and Embed-RL-4B + T-CoT lose 0.0433--0.1321 mAP across all 12 model--track--domain combinations when the image is removed while retaining the reference product text, and even the exceptional weaker configurations reach only 39.6\%--58.2\% of the corresponding subset's best full-input score. Second, the image's contribution is not always positive: \texttt{text+refmeta} exceeds full input in 10 cells, all of them in weaker configurations ranked 6--10 of 11 by full-input mAP on the corresponding subset (boxed in red). Third, the absolute loss is bounded by the baseline level: four single-modality comparisons on low-baseline Cars settings lose less than 0.026, simply because there is little room to fall. Taken together, high-level retrieval does require both modalities---no single-modality shortcut preserves it---but the gain from adding the image depends on a model's cross-modal integration: when image and text are poorly coordinated, adding the image can interfere with the representation, which is consistent with the localized gains of weaker configurations under \texttt{text+refmeta}. Note that pure text-only removes both the image and the reference product text, so it does not isolate image removal; six paired rank tests among the ten positive mean differences are significant after Holm correction, but mean differences and rank effects do not always share a direction, so these are not collectively described as significant mean-AP gains. This mechanism reading is consistent with the observations but is not a causal conclusion identified by this ablation alone.

\input{sections/en/D_leaf_image_ablation}

\subsection{Reranking Reference and Recall-Limited Upper Bound}
\label{app:e5-reranking-en}

We freeze Qwen3-VL-Embedding-8B as the first-stage retriever and use Qwen3-VL-Reranker-8B to jointly score retrieved query--candidate pairs, comparing vector retrieval, actual reranking, and ideal ordering. For every query in the six main subsets, the embedder retrieves Top-30 from the 1,000-candidate pool; the reranker then receives each query--candidate pair with the main protocol's task fields and serialization and reorders only those 30 items. The abstraction track retains its direction-specific I2C/C2I instruction; the constraints and state tracks use the official generic wrapper. The official implementation runs in bfloat16 with FlashAttention-2. This pass covers 1,000 queries and 30,000 scored pairs per subset, 180,000 in total.

Full-pool AP is recomputed after splicing the reranked Top-30 back above the unchanged rank-31--1,000 tail. Positive coverage@30 is the query-average fraction of all positives contained in the first-stage Top-30. Oracle@30 places positives before negatives within the same retrieved set while preserving ranks 31--1,000, giving the ideal upper bound when only the Top-30 can be reordered. Actual reranker performance measures how much ranking quality joint scoring recovers. We recomputed all 6,000 per-query AP values from the saved rankings. The main results sort candidates by descending reranker score, retaining the original first-stage order for ties. To examine numerical sensitivity, we consider two extreme orderings within each tied group only in the sensitivity analysis: positives first and positives last. Across the six subsets, the maximum mAP deviation from the main results is 0.0031, and every subset retains a positive gain over first-stage retrieval.

Table~\ref{tab:ceilings-en} reports first-stage and reranked mAP, paired gains and statistical tests, positive coverage@30 to describe recall, and oracle@30 as the ideal ordering upper bound within that retrieved set. The leading column lists the best configuration for each subset in Table~\ref{tab:main-results-en} as a descriptive reference; all other values come from the same full run.

\input{tables/tab07_ceilings_en}

\subsection{Resolution and Reliability Checks}
\label{app:reliability-checks-en}

This section examines discrimination between models, stability under query rewording, and training-data exposure. Human reviews of queries and visual attributes are reported separately in Appendix~\ref{app:e6a-audit-en}.

\paragraph{Statistical Resolution.}

A benchmark used for ranking must state its resolution: how large a difference lies outside the noise. We therefore resample the frozen query-level AP records at $n\in\{100,200,300,500,700,1000\}$. Figure~\ref{fig:power-en}a shows that the median 95\% bootstrap half-width over the six subsets falls from 0.060 at $n=100$ to 0.019 at $n=1000$ (subset range 0.012--0.022). Figure~\ref{fig:power-en}b uses one representative adjacent-ranked pair per subset and asks what constant paired improvement reaches 80\% power under a two-sided Wilcoxon test, with Holm correction over all six subsets. The median minimum detectable effect falls from 0.036 to 0.009; at 1,000 queries its full subset range is 0.004--0.019, with the median below the 0.012 median gap of the six selected adjacent pairs. Detectability is pair- and subset-dependent, however; this summary does not establish significance for every adjacent pair or for isolated sub-point differences. The wider marginal confidence intervals in panel (a) do not contradict panel (b): pairing removes query-level variation shared by the two configurations.

\begin{paperfigure}
  \input{figures/fig_power}
  \caption{\textbf{Resolution of the benchmark.} (a) Median 95\% query-bootstrap half-width for each subset, computed with 3,000 resamples; the black line is the median over subsets and the dotted boundaries show their range. All six subsets summarize the 20 configurations in the main table using current query-level records. (b) Minimum detectable constant paired mAP improvement at 80\% power, estimated with 1,000 Monte Carlo resamples of a representative adjacent-ranked pair per subset, two-sided Wilcoxon tests, and one six-subset Holm family. The red dashed line in both panels is the median gap (0.012) of the six selected adjacent pairs.}
  \label{fig:power-en}
\end{paperfigure}

\paragraph{Query Formulation Robustness.}
We compare three equivalent query formulations using Qwen3-VL-Embedding-8B, RzenEmbed-v2, and Embed-RL-4B + T-CoT. For the abstraction track, we re-encode all 1,000 queries in each retrieval direction using the original instruction and two human-written equivalent alternatives. For the constraints and state tracks, we fix the candidate side and change only the wording of requests or dialogues. Across all 18 model--subset/direction combinations, the mAP range over the three formulations is below 0.026, with maxima of 0.0240 for abstraction and 0.0247 for constraints and state. The largest rank shift is one, with Qwen3-VL and Rzen swapping only on the Fashion state subset under the second wording variant. The main capability conclusions remain stable under equivalent formulations, although absolute scores still vary with wording.

\paragraph{Training-Data Exposure.}
We audit official papers, model cards, and project pages for 17 independent systems, distinguishing source images (R1), source labels and ontology (R2), and TRIAD-specific query wording (R3), relation and violated-role assignments (R4), and candidate sets and protocol instances (R5; Table~\ref{tab:e14-exposure-en}). Ten systems have confirmed use of source image datasets, and eleven involve ImageNet-1K categories or WordNet relations; dataset-level exposure does not establish that every selected TRIAD image was used in training. Eight systems partially disclose deduplication methods and nine provide no verifiable disclosure; neither supports a blanket decontamination claim.

E5-Omni explicitly uses MMEB training data, which include ImageNet-1K classification training, establishing source-image and category-label exposure~\citep{chen2026e5omni,jiang2024vlm2vec}. WeMM describes public, weakly supervised web, synthetic, and internal data without an exhaustive source list; its within-batch near-duplicate handling is not benchmark-specific decontamination~\citep{zhou2026wemm}. Unlisted sources cannot therefore be excluded.

TRIAD query wording, positive/negative role assignments, and per-query candidate combinations are newly constructed in this study, remain unpublished before submission, and were not used to train the evaluated checkpoints. R3--R5 in Table~\ref{tab:e14-exposure-en} therefore indicate unpublished benchmark-specific content. The relevant overlap concerns pre-existing source images, labels, and conceptual knowledge: familiarity with these materials does not imply prior access to complete TRIAD evaluation instances.

\input{tables/tab_e14_exposure_en}

\subsection{Qualitative Examples}
\label{app:qualitative-cases-en}

\begin{figure}[p]
  \centering
  \includegraphics[width=.97\linewidth]{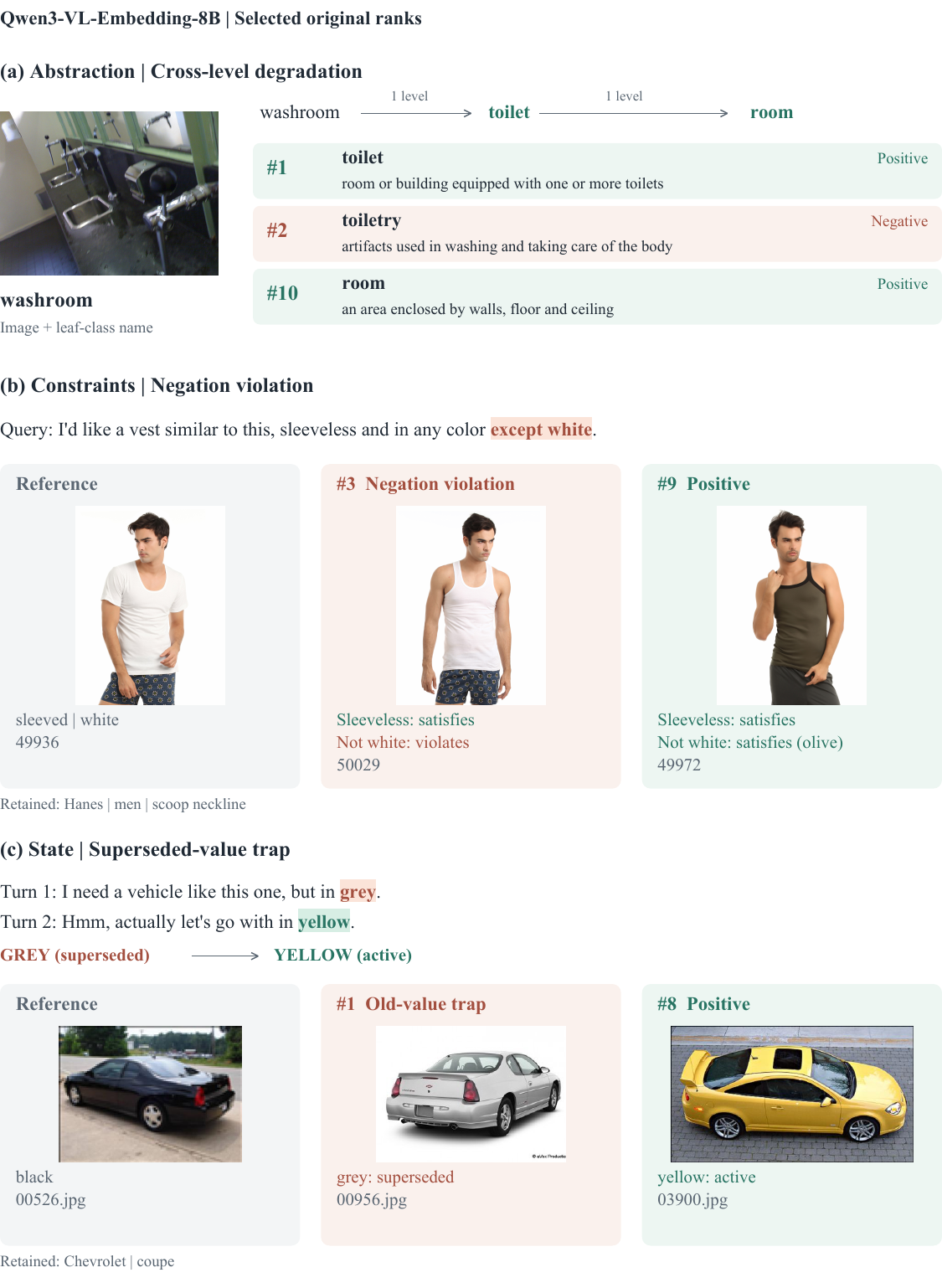}
  \caption{\textbf{Qualitative examples across the three tracks.} All panels show selected original candidate ranks from Qwen3-VL-Embedding-8B; green denotes annotated positives and red negatives. (a) Abstraction: the query contains an image and its leaf-class name. Toilet and room are ancestors one and two levels above the query; toiletry is an ordinary negative. Concept definitions are excerpted for display. (b) Constraints: the query requires sleeveless and not white. The white vest violates only the color exclusion, whereas the olive positive satisfies the full request; both retain brand, gender, and neckline. (c) State: the user revises grey to yellow. Both the old-value trap and the yellow positive retain Chevrolet and coupe. Queries and dialogue turns are reproduced verbatim; attributes follow the dataset annotations, and unshown ranks are omitted. Warm highlighting marks the exclusion phrase and superseded value; green highlighting marks the active value after revision.}
  \label{fig:qualitative-cases-en}
\end{figure}

Figure~\ref{fig:qualitative-cases-en} presents three examples from Qwen3-VL-Embedding-8B, illustrating ranking difficulties in cross-level retrieval, negative-constraint satisfaction, and dialogue-state updates.

\FloatBarrier
\noindent\begin{minipage}{\linewidth}
\paragraph{Cross-level degradation.}
For the washroom query, the model ranks the immediate ancestor toilet first but places the next ancestor, room, tenth. The contextually related yet non-ancestral concept toiletry appears second. Here, toilet denotes a room or building equipped with toilets, not a plumbing fixture. This example shows that successful matching at a nearby level need not extend to a broader valid concept.

\paragraph{Negation violation.}
The constraints-track query asks for a similar garment that is sleeveless and not white. The white vest at rank 3 satisfies the sleeve requirement but violates the color exclusion, whereas the olive positive satisfying both appears at rank 9. Both items also satisfy the retained brand, gender, and neckline conditions. This example shows a partially compliant candidate outranking a fully compliant one: satisfying the positive condition does not compensate for violating the negative condition.

\paragraph{Superseded-value trap.}
In the state-track example, the user first requests a grey car and then changes the color to yellow. The final intent retains Chevrolet and coupe while replacing grey with yellow. Nevertheless, a trap candidate matching the superseded grey value ranks first, while the highest-ranked positive matching the final yellow value appears eighth. The ranking thus favors a withdrawn value over the active requirement, illustrating the difficulty of distinguishing historical requests from the current intent.
\end{minipage}

%% file: sections/en/D_leaf_image_ablation.tex
\subsubsection{The Role of Images in Cross-Level Retrieval}
\label{app:leaf-image-ablation-en}

I2C queries provide both an image and its leaf-class name. To examine whether the class name can substitute for the image, we keep the query instruction, query-specific candidate pools, and positive labels fixed, remove only the original query image, and retain the leaf-class name. We compare this variant with the full image-and-name input across 11 configurations on the same 711 queries with positives at both one and two levels. We report overall mAP with positives at both distances scored jointly, rather than using distance-specific score differences as a measure of modality contribution.

Table~\ref{tab:leaf-image-ablation-en} shows lower overall mAP after image removal in all 11 configurations, with losses of 0.018--0.105; ten configurations show significant differences after Holm correction. Qwen3-VL-Embedding-8B, RzenEmbed-v2, and Embed-RL-4B + T-CoT lose 0.054, 0.089, and 0.070 mAP, respectively. VLM2Vec-2B has the smallest loss, 0.018, which is not statistically significant.

\begin{table}[!htbp]
\caption{I2C with the leaf-class name retained and the original query image removed. Both inputs use the same 711 queries, query-specific candidate pools, and positive labels. mAP scores positives at both distances jointly; $\Delta$ is image-removed minus full input. $^\dagger$ denotes paired Wilcoxon $p<0.05$ after Holm correction across the overall, one-level, and two-level contrasts within each configuration. The variant retains the I2C instruction; UME-R1 uses its text interface's blank-image placeholder, without the original query image, and Embed-RL generates external reasoning from the leaf-class name. Means and differences are reported separately at archived precision, so subtraction of displayed values may differ in the last digit.}
\label{tab:leaf-image-ablation-en}
\small
\setlength{\tabcolsep}{5pt}
\begin{center}
\begin{tabular}{@{}lrrr@{}}
\toprule
Configuration & \shortstack{Image +\\leaf name} & \shortstack{Leaf name\\only} & $\Delta$ mAP \\
\midrule
Qwen3-VL-Embedding-8B & 0.5523 & 0.4980 & $-0.0543^{\dagger}$ \\
Qwen3-VL-Embedding-2B & 0.5257 & 0.4699 & $-0.0558^{\dagger}$ \\
OmniEmbed-3B & 0.4407 & 0.3506 & $-0.0901^{\dagger}$ \\
RzenEmbed-v2 & 0.6178 & 0.5290 & $-0.0887^{\dagger}$ \\
B-Tokens-2B & 0.5002 & 0.3950 & $-0.1052^{\dagger}$ \\
VLM2Vec-2B & 0.3470 & 0.3287 & $-0.0182$ \\
VLM2Vec-7B & 0.4578 & 0.4133 & $-0.0445^{\dagger}$ \\
VLM2Vec-V2.0-2B & 0.3767 & 0.3137 & $-0.0630^{\dagger}$ \\
TWN-4B & 0.5705 & 0.4660 & $-0.1044^{\dagger}$ \\
Embed-RL-4B + T-CoT & 0.5993 & 0.5289 & $-0.0704^{\dagger}$ \\
UME-R1-7B (Gen) & 0.5973 & 0.5180 & $-0.0793^{\dagger}$ \\
\bottomrule
\end{tabular}
\end{center}
\end{table}

Class names thus provide important information for hierarchical retrieval but do not preserve full-input performance in these configurations: the original image improves I2C cross-level ranking even when the class name is given. This supports an incremental contribution from images in the evaluated I2C settings, rather than establishing images as necessary for every abstraction retrieval request.

%% file: tables/tab07_ceilings_en.tex
\begin{table}[!htbp]
\caption{Top-30 reranking results, paired analyses, and the recall-limited oracle upper bound on all 1,000 queries per subset. Qwen3-VL-Embedding-8B performs first-stage retrieval and Qwen3-VL-Reranker-8B reranks the Top-30. All six subsets show significant improvements under two-sided Wilcoxon signed-rank tests with Holm correction ($p_{\mathrm{Holm}}<10^{-11}$). Positive coverage@30 is the query-average fraction of positives recalled; oracle@30 ideally orders the same Top-30, providing a recall-limited upper bound. Values other than coverage are full-pool mAP or mAP differences. $^\dagger$ denotes the best configuration for that subset in Table~\ref{tab:main-results-en}, used only as a descriptive reference; paired tests compare Qwen first-stage retrieval with its own reranked output. Human relevance ratings on ten-candidate sets appear in Appendix~\ref{app:e6b-ratings-en} and do not represent a human ceiling for the full 1,000-candidate pool.}
\label{tab:ceilings-en}
\begin{center}
\begingroup\small
\setlength{\tabcolsep}{3.5pt}
\begin{tabular}{@{}lrrrrrr@{}}
\toprule
\textbf{Subset} & \shortstack{best pooled\\system$^\dagger$} & \shortstack{Qwen embed.\\base} & \shortstack{Top-30\\reranked} & \shortstack{reranked\\$-$ base} & \shortstack{positive\\coverage@30} & \shortstack{oracle\\@30} \\
\midrule
A--I2C     & 0.569 & 0.494 & 0.607 & +0.113 & 0.865 & 0.868 \\
A--C2I     & 0.638 & 0.637 & 0.674 & +0.037 & 0.869 & 0.877 \\
B--Fashion & 0.547 & 0.508 & 0.566 & +0.058 & 0.884 & 0.887 \\
B--Cars    & 0.367 & 0.230 & 0.327 & +0.097 & 0.657 & 0.663 \\
C--Fashion & 0.498 & 0.454 & 0.516 & +0.062 & 0.891 & 0.894 \\
C--Cars    & 0.291 & 0.177 & 0.258 & +0.081 & 0.616 & 0.623 \\
\bottomrule
\end{tabular}
\par\endgroup
\end{center}
\end{table}

%% file: figures/fig_power.tex
\input{figures/fig_power_data}
\begin{tikzpicture}
\begin{groupplot}[
  group style={group size=2 by 1, horizontal sep=1.35cm},
  width=0.39\linewidth, height=4.25cm,
  xmin=70, xmax=1030, xtick={100,300,500,700,1000},
  xlabel={queries per subset}, xlabel style={yshift=2pt},
  ymin=0, scaled y ticks=false,
  tick label style={font=\scriptsize},
  label style={font=\scriptsize},
  title style={font=\scriptsize},
  grid=major, grid style={black!8},
]
\nextgroupplot[
  title={(a) Bootstrap uncertainty},
  ylabel={95\% CI half-width (mAP)},
  ymax=0.076,
  ytick={0,0.02,0.04,0.06}, yticklabels={0,.02,.04,.06},
]
\addplot[\XOneAIToCColor, line width=0.45pt, mark=none] coordinates {\XOneAIToCCICoords};
\addplot[\XOneACToIColor, line width=0.45pt, mark=none] coordinates {\XOneACToICICoords};
\addplot[\XOneBFashionColor, line width=0.45pt, mark=none] coordinates {\XOneBFashionCICoords};
\addplot[\XOneBCarsColor, line width=0.45pt, mark=none] coordinates {\XOneBCarsCICoords};
\addplot[\XOneCFashionColor, line width=0.45pt, mark=none] coordinates {\XOneCFashionCICoords};
\addplot[\XOneCCarsColor, line width=0.45pt, mark=none] coordinates {\XOneCCarsCICoords};
\addplot[black!35, densely dotted, line width=0.45pt, mark=none] coordinates {\XOneCIMinCoords};
\addplot[black!35, densely dotted, line width=0.45pt, mark=none] coordinates {\XOneCIMaxCoords};
\addplot[black, line width=1.0pt, mark=*, mark size=1.15pt] coordinates {\XOneCIMedianCoords};
\addplot[red!75!black, dashed, line width=0.8pt, mark=none]
  coordinates {(70,\XOneGlobalAdjacentGap)(1030,\XOneGlobalAdjacentGap)};

\nextgroupplot[
  title={(b) Holm-adjusted 80\% power},
  ylabel={minimum detectable $\Delta$ (mAP)},
  ymax=0.080,
  ytick={0,0.02,0.04,0.06,0.08}, yticklabels={0,.02,.04,.06,.08},
]
\addplot[\XOneAIToCColor, line width=0.45pt, mark=none] coordinates {\XOneAIToCMDECoords};
\addplot[\XOneACToIColor, line width=0.45pt, mark=none] coordinates {\XOneACToIMDECoords};
\addplot[\XOneBFashionColor, line width=0.45pt, mark=none] coordinates {\XOneBFashionMDECoords};
\addplot[\XOneBCarsColor, line width=0.45pt, mark=none] coordinates {\XOneBCarsMDECoords};
\addplot[\XOneCFashionColor, line width=0.45pt, mark=none] coordinates {\XOneCFashionMDECoords};
\addplot[\XOneCCarsColor, line width=0.45pt, mark=none] coordinates {\XOneCCarsMDECoords};
\addplot[black!35, densely dotted, line width=0.45pt, mark=none] coordinates {\XOneMDEMinCoords};
\addplot[black!35, densely dotted, line width=0.45pt, mark=none] coordinates {\XOneMDEMaxCoords};
\addplot[black, line width=1.0pt, mark=square*, mark size=1.15pt] coordinates {\XOneMDEMedianCoords};
\addplot[red!75!black, dashed, line width=0.8pt, mark=none]
  coordinates {(70,\XOneGlobalAdjacentGap)(1030,\XOneGlobalAdjacentGap)};
\end{groupplot}
\path (group c1r1.south west) -- (group c2r1.south east)
  node[midway, below=0.78cm, font=\tiny] {\begin{tabular}{@{}cccc@{}}
  \textcolor{\XOneAIToCColor}{\rule{7pt}{0.7pt}} A--I2C &
  \textcolor{\XOneACToIColor}{\rule{7pt}{0.7pt}} A--C2I &
  \textcolor{\XOneBFashionColor}{\rule{7pt}{0.7pt}} B--Fashion &
  \textcolor{\XOneBCarsColor}{\rule{7pt}{0.7pt}} B--Cars \\
  \textcolor{\XOneCFashionColor}{\rule{7pt}{0.7pt}} C--Fashion &
  \textcolor{\XOneCCarsColor}{\rule{7pt}{0.7pt}} C--Cars &
  \textcolor{black}{\rule{7pt}{0.9pt}} subset median &
  \textcolor{red!75!black}{\rule{7pt}{0.7pt}} adjacent gap
  \end{tabular}};
\end{tikzpicture}

%% file: figures/fig_power_data.tex
\def\XOneGlobalAdjacentGap{0.012354}
\def\XOneCIMedianCoords{(100,0.059690) (200,0.042041) (300,0.033532) (500,0.026677) (700,0.022328) (1000,0.018530)}
\def\XOneCIMinCoords{(100,0.037608) (200,0.026794) (300,0.021471) (500,0.016673) (700,0.014070) (1000,0.011894)}
\def\XOneCIMaxCoords{(100,0.068257) (200,0.047526) (300,0.039401) (500,0.030409) (700,0.025803) (1000,0.021589)}
\def\XOneMDEMedianCoords{(100,0.036039) (200,0.023593) (300,0.018607) (500,0.014867) (700,0.011374) (1000,0.009197)}
\def\XOneMDEMinCoords{(100,0.012968) (200,0.009025) (300,0.008082) (500,0.005726) (700,0.004381) (1000,0.004068)}
\def\XOneMDEMaxCoords{(100,0.071905) (200,0.047500) (300,0.036111) (500,0.026869) (700,0.023315) (1000,0.018722)}
\def\XOneAIToCColor{purple!80!black}
\def\XOneAIToCCICoords{(100,0.068257) (200,0.047526) (300,0.039401) (500,0.030409) (700,0.025803) (1000,0.021589)}
\def\XOneAIToCMDECoords{(100,0.071905) (200,0.047500) (300,0.036111) (500,0.026869) (700,0.023315) (1000,0.018722)}
\def\XOneACToIColor{violet!65!black}
\def\XOneACToICICoords{(100,0.060350) (200,0.042831) (300,0.034044) (500,0.027457) (700,0.022569) (1000,0.018700)}
\def\XOneACToIMDECoords{(100,0.047603) (200,0.032540) (300,0.024900) (500,0.019671) (700,0.016774) (1000,0.013931)}
\def\XOneBFashionColor{teal!85!black}
\def\XOneBFashionCICoords{(100,0.067663) (200,0.047259) (300,0.038280) (500,0.029902) (700,0.025181) (1000,0.021216)}
\def\XOneBFashionMDECoords{(100,0.042879) (200,0.027863) (300,0.022672) (500,0.017831) (700,0.013187) (1000,0.009880)}
\def\XOneBCarsColor{cyan!60!black}
\def\XOneBCarsCICoords{(100,0.043946) (200,0.031127) (300,0.025380) (500,0.020277) (700,0.016660) (1000,0.013892)}
\def\XOneBCarsMDECoords{(100,0.014527) (200,0.009750) (300,0.008632) (500,0.006738) (700,0.004681) (1000,0.004111)}
\def\XOneCFashionColor{orange!90!black}
\def\XOneCFashionCICoords{(100,0.059030) (200,0.041251) (300,0.033020) (500,0.025897) (700,0.022088) (1000,0.018360)}
\def\XOneCFashionMDECoords{(100,0.029199) (200,0.019323) (300,0.014542) (500,0.011904) (700,0.009560) (1000,0.008514)}
\def\XOneCCarsColor{red!70!black}
\def\XOneCCarsCICoords{(100,0.037608) (200,0.026794) (300,0.021471) (500,0.016673) (700,0.014070) (1000,0.011894)}
\def\XOneCCarsMDECoords{(100,0.012968) (200,0.009025) (300,0.008082) (500,0.005726) (700,0.004381) (1000,0.004068)}

%% file: tables/tab_e14_exposure_en.tex
\begin{table}[!htbp]
\caption{Public-source exposure audit for 17 systems. $\bullet$: confirmed use at the source-dataset level; grey $\bullet$: cannot exclude; $\circ$: content constructed in this study, unpublished before submission and not used to train the evaluated models. R1/R2 concern source images and labels/ontology; R3--R5 concern benchmark-specific queries, role assignments, and candidate combinations. Deduplication disclosure does not establish removal of TRIAD overlaps. Source materials for WeMM and E5-Omni were rechecked on September 20, 2026.}
\label{tab:e14-exposure-en}
\begin{center}
\begingroup\scriptsize
\setlength{\tabcolsep}{3pt}
\renewcommand{\arraystretch}{1.0}
\begin{tabular}{@{}>{\raggedright\arraybackslash}p{.36\linewidth}ccccc>{\centering\arraybackslash}p{.15\linewidth}@{}}
\toprule
\textbf{System} & \textbf{R1} & \textbf{R2} & \textbf{R3} & \textbf{R4} & \textbf{R5} & \shortstack{Dedup.} \\
\midrule
CLIP ViT-L/14 & \textcolor{gray!68}{$\bullet$} & $\bullet$ & $\circ$ & $\circ$ & $\circ$ & None \\
SigLIP-2-so400m & \textcolor{gray!68}{$\bullet$} & \textcolor{gray!68}{$\bullet$} & $\circ$ & $\circ$ & $\circ$ & None \\
jina-clip-v2 & \textcolor{gray!68}{$\bullet$} & \textcolor{gray!68}{$\bullet$} & $\circ$ & $\circ$ & $\circ$ & None \\
RzenEmbed-v2 & $\bullet$ & $\bullet$ & $\circ$ & $\circ$ & $\circ$ & Partial \\
Qwen3-VL-Embedding-8B & \textcolor{gray!68}{$\bullet$} & \textcolor{gray!68}{$\bullet$} & $\circ$ & $\circ$ & $\circ$ & Partial \\
Qwen3-VL-Embedding-2B & \textcolor{gray!68}{$\bullet$} & \textcolor{gray!68}{$\bullet$} & $\circ$ & $\circ$ & $\circ$ & Partial \\
OmniEmbed-3B & \textcolor{gray!68}{$\bullet$} & \textcolor{gray!68}{$\bullet$} & $\circ$ & $\circ$ & $\circ$ & None \\
VLM2Vec-V2.0-2B & $\bullet$ & $\bullet$ & $\circ$ & $\circ$ & $\circ$ & None \\
VLM2Vec-2B & $\bullet$ & $\bullet$ & $\circ$ & $\circ$ & $\circ$ & None \\
VLM2Vec-7B & $\bullet$ & $\bullet$ & $\circ$ & $\circ$ & $\circ$ & None \\
B-Tokens-2B & $\bullet$ & $\bullet$ & $\circ$ & $\circ$ & $\circ$ & None \\
Embed-RL-2B & $\bullet$ & $\bullet$ & $\circ$ & $\circ$ & $\circ$ & Partial \\
Embed-RL-4B & $\bullet$ & $\bullet$ & $\circ$ & $\circ$ & $\circ$ & Partial \\
UME-R1-7B & $\bullet$ & $\bullet$ & $\circ$ & $\circ$ & $\circ$ & Partial \\
TWN-4B & $\bullet$ & $\bullet$ & $\circ$ & $\circ$ & $\circ$ & Partial \\
WeMM-9B & \textcolor{gray!68}{$\bullet$} & \textcolor{gray!68}{$\bullet$} & $\circ$ & $\circ$ & $\circ$ & Partial \\
E5-Omni-7B & $\bullet$ & $\bullet$ & $\circ$ & $\circ$ & $\circ$ & None \\
\midrule
\textbf{Confirmed inclusion} & 10 & 11 & 0 & 0 & 0 & \multicolumn{1}{r}{Partial 8 / None 9} \\
\bottomrule
\end{tabular}
\par\endgroup
\end{center}
\end{table}

%% file: sections/en/E_quality_audit.tex
\section{Human Quality Review and Relevance Ratings}
\label{app:quality-audit-en}

This appendix reports human sample-based reviews of query expression and visual attributes, as well as human relevance ratings on small candidate sets.

\subsection{Query Quality and Visual Attribute Accuracy}
\label{app:e6a-audit-en}

\paragraph{Sampling and scope.}
We sample queries and images separately with seed 2027. Query review covers B--Fashion, B--Cars, C--Fashion, and C--Cars, with 20 queries or sessions per subset and 80 in total. Sampling balances category coverage and an approximately even allocation across D1--D7; each state-track domain contains 10 natural and 10 conflict sessions. These samples therefore examine different query types rather than estimate corpus-wide quality in proportion to the population.

Visual-attribute review samples 40 images from each of the Fashion and Cars domain pools, without duplicate sampling across B/C. Fashion covers eight coarse categories with five images each, yielding 361 attributes; Cars covers seven body types with five or six images each, yielding 160 attributes. Visual verification excludes source-dataset textual metadata, and the abstraction track's concept taxonomy is also outside this sample review. A single human annotator performs the human reviews and relevance ratings; these activities do not require ethics approval under departmental requirements.

\paragraph{Query review and issue classification.}
We examine query expression using the query text, reference image, and construction conditions, and assess how detected issues affect understanding of the requirements. Among the 80 queries or sessions, no issue is found in 67; eight raise wording or intent concerns that, on inspection, do not affect understanding, mainly involving unnatural phrasing; five may affect understanding. Table~\ref{tab:e6a-query-en} summarizes the results by subset.

\begin{table}[!htbp]
\caption{Human query-quality review of 20 tasks per subset. The three mutually exclusive categories distinguish whether an issue is found and its impact on understanding the requirements.}
\label{tab:e6a-query-en}
\small
\setlength{\tabcolsep}{5pt}
\begin{center}
\input{tables/e6a_query_counts_en}
\end{center}
\end{table}

All five records that may affect understanding are from Fashion: two concern year or new/old-style wording, two concern product-category names and source metadata, and one concerns the attribute's target part. For example, a watch session uses \emph{not in blue} in both turns, but the construction conditions refer to the band and dial separately, without the query distinguishing these parts. In contrast, expressions such as \emph{running cap cap type} are unnatural but remain understandable. This classification distinguishes wording issues from issues that may affect task understanding; the original queries and data labels have not been modified on this basis.

\paragraph{Visual attributes.}
For each displayed attribute annotation, we use the image to judge it correct, incorrect, or not visible/undetermined. Determinate accuracy uses correct plus incorrect judgments as its denominator; the not-visible/undetermined rate uses all attributes. Among the sampled images, 507 of 513 determinate visual attributes are annotated correctly, with eight additional attributes marked not visible/undetermined, yielding 98.83\% accuracy. The not-visible/undetermined items account for 1.54\% of all 521 attributes (Table~\ref{tab:e6a-visual-en}). These results characterize visual-attribute annotation quality in the sampled images, not the accuracy of queries or candidate relevance labels.

\begin{table}[!htbp]
\caption{Human visual-attribute verification of 40 images per domain: 361 attribute judgments for Fashion and 160 for Cars. Determinate accuracy excludes not-visible/undetermined items; their rate uses all attributes as the denominator.}
\label{tab:e6a-visual-en}
\small
\setlength{\tabcolsep}{4pt}
\begin{center}
\input{tables/e6a_visual_counts_en}
\end{center}
\end{table}

\subsection{Human Relevance Rating on Small Candidate Sets}
\label{app:e6b-ratings-en}

\paragraph{Samples and rating setup.}
We select 20 tasks from each of the six main subsets, showing ten candidates per task. The annotator completes all 120 tasks, yielding 1,200 five-level ratings. Sampling uses difficulty-stratified quotas and requires archived reranker scores for all ten candidates. I2C includes 14 queries with positives both one and two levels away and six with positives only one level away. The C2I depth bins 5--7, 8--10, 11--13, and 14--16 contain 7, 9, 3, and 1 queries. Each constraints/state subset allocates 3, 3, 3, 2, 4, 3, and 2 tasks to D1--D7. Stratified quotas use seed 2027; within-stratum replacements satisfying score completeness use seed 20260909.

Each task retains all dataset gold candidates (at most five) and fills the remaining slots with the highest-ranked non-gold candidates in the reranker's natural Top 30, then shuffles their presentation. All gold candidates in the final comparison sample are recalled in the natural Top 30. The interface displays query images and text, concept names and definitions where applicable, product metadata, and full dialogues, with Chinese structured-condition aids. Model scores, original ranks, and gold labels are hidden.

\paragraph{Five grades and expected AP.}
Grades are 1 (definitely irrelevant), 2 (probably irrelevant), 3 (uncertain), 4 (probably relevant), and 5 (definitely relevant). Rankings sort grades in descending order and average uniformly over all permutations within tied grades. Grade 3 remains a middle tier rather than replacing the benchmark's binary relevance labels. For query $q$, let $P_q$ denote the original gold set. For each grade $g$, let $n_g$ and $m_g$ count its candidates and gold candidates, and let $N_g$ and $M_g$ denote the corresponding counts in higher grades. Then
\begin{equation}
\mathbb{E}[\mathrm{AP}_q]=\frac{1}{|P_q|}
\sum_{g:m_g>0}\sum_{j=1}^{n_g}
\frac{m_g}{n_g}\,
\frac{M_g+1+(j-1)(m_g-1)/(n_g-1)}{N_g+j},
\label{eq:e6b-expected-ap-en}
\end{equation}
where the within-grade correction is zero for $n_g=1$. Expected mAP averages the query-level expected AP. Across 31 tasks, 76 candidates receive grade 3 and are included according to the same ranking rule.

\paragraph{Same-pool results.}
The reranker sorts scores in descending order, breaking exact ties by base rank. Human and reranker AP use identical ten-candidate and gold sets, and all 20 tasks in each subset enter the comparison. Table~\ref{tab:e6b-same-pool-en} shows lower human mean AP on both abstraction subsets and higher human mean AP on all four constraints/state subsets. This measures assisted judgment on small candidate sets; gold-inclusive sampling conditioned on complete recall does not establish a human upper bound for the full 1,000-candidate pool.

\begin{table}[!htbp]
\caption{Human and reranker comparison on identical ten-candidate sets. One human annotator rates all six subsets, with 20 score-complete tasks per row. Differences are human expected mAP minus reranker mAP; entries are descriptive means.}
\label{tab:e6b-same-pool-en}
\label{tab:e6b-ratings-en}
\label{tab:e6b-continuation-en}
\small
\setlength{\tabcolsep}{5pt}
\begin{center}
\input{tables/e6b_same_pool_en}
\end{center}
\end{table}

%% file: tables/e6a_query_counts_en.tex
\begin{tabular}{@{}lrrr@{}}
\toprule
Subset & \shortstack{No issue\\found} & \shortstack{Concerns without impact\\on understanding} & \shortstack{May affect\\understanding} \\
\midrule
B--Fashion & 17 & 0 & 3 \\
B--Cars & 17 & 3 & 0 \\
C--Fashion & 16 & 2 & 2 \\
C--Cars & 17 & 3 & 0 \\
\midrule
Total & 67 & 8 & 5 \\
\bottomrule
\end{tabular}

%% file: tables/e6a_visual_counts_en.tex
\begin{tabular}{@{}lrrrrr@{}}
\toprule
Domain & Correct & Incorrect & \shortstack{Not visible/\\undetermined} & \shortstack{Determinate\\accuracy} & \shortstack{Not visible/\\undetermined rate} \\
\midrule
Fashion & 352 & 3 & 6 & 99.15\% & 1.66\% \\
Cars & 155 & 3 & 2 & 98.10\% & 1.25\% \\
Total & 507 & 6 & 8 & 98.83\% & 1.54\% \\
\bottomrule
\end{tabular}

%% file: tables/e6b_same_pool_en.tex
\begin{tabular}{lrrrr}
\toprule
Subset & $n$ & Human mAP & Reranker mAP & Difference \\
\midrule
A-I2C & 20 & 0.5512 & 0.6288 & $-0.0776$ \\
A-C2I & 20 & 0.7280 & 0.7657 & $-0.0377$ \\
B-Fashion & 20 & 0.7021 & 0.5725 & $+0.1297$ \\
B-Cars & 20 & 0.6610 & 0.4369 & $+0.2241$ \\
C-Fashion & 20 & 0.6750 & 0.5108 & $+0.1641$ \\
C-Cars & 20 & 0.7343 & 0.4160 & $+0.3183$ \\
\bottomrule
\end{tabular}

%% file: sections/en/F_benchmark_comparison.tex
\section{Supplementary Comparisons with Existing Benchmarks}

\label{app:bench-comparison-en}

\label{app:compare-en}

\subsection{Design Criteria and Extended Comparison}

\label{app:bench-criteria-en}

Table~\ref{tab:bench-extended-en} extends the main-text comparison along six dimensions. Partial indicates that an operation is covered but organized or scored differently from the criterion used here.

\begin{table}[!htbp]
\footnotesize\setlength{\tabcolsep}{3pt}\renewcommand{\arraystretch}{1.12}
\caption{Extended design comparison with related benchmarks.}
\label{tab:bench-extended-en}
\begin{center}
\begin{tabularx}{\linewidth}{@{}l*{6}{>{\centering\arraybackslash}X}@{}}
\toprule
Benchmark & Cross-level retrieval & Condition control & State updates & Multi-positive scoring & Typed negatives & Low-complexity reference \\
\midrule
M-BEIR & --- & Partial a & --- & Partial b & --- & --- \\
MMEB & --- & Partial a & Partial d & --- & --- & --- \\
MMEB-V2 & --- & Partial a & Partial d & Partial b & --- & --- \\
CIRCO & --- & Partial a & --- & $\checkmark$ & --- & --- \\
GeneCIS & --- & Partial a & --- & --- & $\checkmark$ & --- \\
NegBench & --- & $\checkmark$ & --- & --- & Partial c & $\checkmark$ \\
MA-CIR & --- & $\checkmark$ & --- & --- & Partial c & $\checkmark$ \\
ChatSearch & --- & Partial a & Partial d & --- & --- & --- \\
TRIAD & $\checkmark$ & $\checkmark$ & $\checkmark$ & $\checkmark$ & $\checkmark$ & $\checkmark$ \\
\bottomrule
\end{tabularx}
\end{center}
\end{table}

The six criteria examine cross-level retrieval with positives defined by explicit ancestor/descendant relations; explicit control of condition polarity or operation composition; retention, replacement, and interference of historical conditions; ranking of multiple positives; negatives typed by violated requirements; and low-complexity references for the target operation.

\paragraph{Cross-level criterion.} This column requires positives to be defined by ancestor/descendant relations in a category hierarchy; ordinary image--class-name matching alone does not qualify. The criterion does not require a query to select an exact hierarchical distance.

\textbf{a. Condition organization.} Editing subtasks in M-BEIR and the MMEB family, CIRCO, and ChatSearch may involve multiple conditions; GeneCIS distinguishes focusing on or changing objects/attributes. TRIAD further groups one to three conditions by positive/negative composition and uses preserved attributes and typed violations to analyze joint satisfaction~\citep{wei2024uniir,jiang2024vlm2vec,baldrati2023circo,vaze2023genecis,zhao2024chatsearch}.

\textbf{b. Multi-positive scoring.} M-BEIR allows multiple correct targets, but its main recall convention checks for any correct hit. Some MMEB-V2 document tasks use multiple relevant evidence items and NDCG@5, whereas image/video tasks use Hit@1. CIRCO and TRIAD both use multi-positive mAP; TRIAD applies it to complete candidate rankings across all six main subsets~\citep{wei2024uniir,meng2025vlm2vecv2,baldrati2023circo}.

\textbf{c. Hard negatives and references.} NegBench tests negation through affirmative/negative alternatives and object-present/absent images; MA-CIR constructs controlled negatives for negation, replacement, and addition. Both provide corresponding simplified references. GeneCIS distinguishes errors in reference content from errors in requested conditions. TRIAD organizes typed negatives and diagnostics around hierarchical relations, condition violations, and obsolete states~\citep{alhamoud2025negbench,byun2025macir,vaze2023genecis}.

\textbf{d. Multi-turn state.} MMEB's VisDial subtask uses dialogue context, and ChatSearch recovers retrieval intent from multi-turn image--text interactions. TRIAD distributes the same final requirements across turns and introduces subsequently replaced values to distinguish active-condition integration from stale-value interference~\citep{jiang2024vlm2vec,zhao2024chatsearch}; task coverage follows the \href{https://huggingface.co/spaces/TIGER-Lab/MMEB-Leaderboard/blob/8129607279d8e4eda8ed80c07b43f7cd45c6767d/datasets.py}{official image-task list}.

\subsection{Ranking Data, Configuration Mapping, and Computation}

We use official MMEB-V2 Image-Overall(V2) scores read on September 19, 2026. This aggregate covers the original 36 image tasks and excludes the subsequently added MCMR. We compare the more closely aligned image portion rather than Overall-V2, which also includes video/document tasks, and do not directly compare absolute scores under different metrics. \href{https://huggingface.co/spaces/TIGER-Lab/MMEB-Leaderboard}{Official leaderboard}, \href{https://huggingface.co/spaces/TIGER-Lab/MMEB-Leaderboard/blob/8129607279d8e4eda8ed80c07b43f7cd45c6767d/utils_v3.py}{scoring implementation}.

The shared set contains 13 systems with one representative configuration each. UME-R1 uses Gen; Embed-RL-2B and Embed-RL-4B correspond to our +T-CoT configurations. The latter use external reasoning from Qwen3-VL-8B-Instruct on TRIAD, so this is a descriptive comparison of public configurations and our implementations, not a claim that every component of the reasoning pipelines is identical. The main-text reversals use only direct-encoding models; the sensitivity analysis below also excludes the two external-reasoning configurations.

Table~\ref{tab:bench-all-ranks-en} combines configuration mapping with complete rankings; differing leaderboard names are listed in its caption.

Both benchmarks rank only these 13 systems from highest to lowest score. MMEB uses the displayed two-decimal scores; TRIAD uses mAP before display rounding, with average ranks for ties. The main text displays selected rows but retains all six subsets; ranks are not recomputed over a smaller displayed set.

For model m and subset t, define rank change as $\Delta$r(m,t)=r\_MMEB(m)-r\_TRIAD(m,t). A positive value denotes a higher position on TRIAD and a negative value a lower position; for example, Rzen's I2C change is 3-1=+2, displayed as 1 ($\uparrow$2). Spearman $\rho$ is the Pearson correlation of the two complete rank vectors. Changes describe point-estimate rankings and do not establish statistical significance for every pairwise rank difference.

\subsection{Complete Rankings and Changes}

\begin{table}[!htbp]
\footnotesize\setlength{\tabcolsep}{3pt}\renewcommand{\arraystretch}{1.12}
\caption{Complete rankings within the same 13-system set. Each cell gives rank and change relative to MMEB image ranking; B denotes constraints and C denotes state. Leaderboard mapping: RzenEmbed-v2 maps to RzenEmbed-v2-7B, B-Tokens-2B to Btoks, and E5-Omni-7B to e5-omni-7B. VLM2Vec-7B/2B map to VLM2Vec-V1-Qwen2VL-7B/2B, VLM2Vec-V2.0-2B to VLM2Vec-V2.0-Qwen2VL-2B, and OmniEmbed-3B to omni-embed-nemotron-3b. Other entries use the model names shown; UME-R1 uses Gen and Embed-RL uses +T-CoT.}
\label{tab:bench-all-ranks-en}
\label{tab:bench-mapping-en}
\begin{center}
\begin{tabularx}{\linewidth}{@{}Xrrrrrrr@{}}
\toprule
Model/configuration & \shortstack{MMEB\\image} & I2C & C2I & \shortstack{B\\Fashion} & \shortstack{B\\Cars} & \shortstack{C\\Fashion} & \shortstack{C\\Cars} \\
\midrule
WeMM-Embedding-9B & 1 & 4 ($\downarrow$3) & 1 (---) & 1 (---) & 1 (---) & 1 (---) & 1 (---) \\
Qwen3-VL-Embedding-8B & 2 & 6 ($\downarrow$4) & 2 (---) & 2 (---) & 4 ($\downarrow$2) & 2 (---) & 5 ($\downarrow$3) \\
RzenEmbed-v2 & 3 & 1 ($\uparrow$2) & 3 (---) & 3 (---) & 2 ($\uparrow$1) & 3 (---) & 2 ($\uparrow$1) \\
Qwen3-VL-Embedding-2B & 4 & 7 ($\downarrow$3) & 8 ($\downarrow$4) & 4 (---) & 10 ($\downarrow$6) & 11 ($\downarrow$7) & 12 ($\downarrow$8) \\
B-Tokens-2B & 5 & 8 ($\downarrow$3) & 10 ($\downarrow$5) & 6 ($\downarrow$1) & 9 ($\downarrow$4) & 6 ($\downarrow$1) & 6 ($\downarrow$1) \\
E5-Omni-7B & 6 & 11 ($\downarrow$5) & 6 (---) & 7 ($\downarrow$1) & 6 (---) & 9 ($\downarrow$3) & 10 ($\downarrow$4) \\
UME-R1-7B (Gen) & 7 & 3 ($\uparrow$4) & 4 ($\uparrow$3) & 9 ($\downarrow$2) & 8 ($\downarrow$1) & 8 ($\downarrow$1) & 8 ($\downarrow$1) \\
Embed-RL-4B + T-CoT & 8 & 2 ($\uparrow$6) & 5 ($\uparrow$3) & 8 (---) & 3 ($\uparrow$5) & 4 ($\uparrow$4) & 3 ($\uparrow$5) \\
Embed-RL-2B + T-CoT & 9 & 5 ($\uparrow$4) & 7 ($\uparrow$2) & 5 ($\uparrow$4) & 5 ($\uparrow$4) & 5 ($\uparrow$4) & 4 ($\uparrow$5) \\
VLM2Vec-7B & 10 & 9 ($\uparrow$1) & 11 ($\downarrow$1) & 11 ($\downarrow$1) & 11 ($\downarrow$1) & 10 (---) & 9 ($\uparrow$1) \\
VLM2Vec-V2.0-2B & 11 & 12 ($\downarrow$1) & 12 ($\downarrow$1) & 10 ($\uparrow$1) & 7 ($\uparrow$4) & 7 ($\uparrow$4) & 7 ($\uparrow$4) \\
VLM2Vec-2B & 12 & 13 ($\downarrow$1) & 13 ($\downarrow$1) & 12 (---) & 12 (---) & 12 (---) & 11 ($\uparrow$1) \\
OmniEmbed-3B & 13 & 10 ($\uparrow$3) & 9 ($\uparrow$4) & 13 (---) & 13 (---) & 13 (---) & 13 (---) \\
\bottomrule
\end{tabularx}
\end{center}
\end{table}

The full table preserves both agreement and reversals. WeMM remains first on five subsets, all except I2C; Qwen3-VL-Embedding-8B remains second on C2I, B-Fashion, and C-Fashion. TRIAD does not reverse the general ranking wholesale but changes relative order under some retrieval requirements. For example, Qwen3-VL-Embedding-2B leads B-Tokens on MMEB but trails it on both state tasks; E5-Omni-7B likewise leads VLM2Vec-V2.0-2B on MMEB but trails it on both state tasks. This is consistent with the need to integrate complete active requirements in multi-turn expression.

\subsection{Rank Correlations and Configuration Sensitivity}

\label{app:bench-sensitivity-en}

\begin{table}[!htbp]
\footnotesize\setlength{\tabcolsep}{3pt}\renewcommand{\arraystretch}{1.12}
\caption{Spearman correlations between MMEB-V2 image rankings and TRIAD rankings.}
\label{tab:bench-correlations-en}
\begin{center}
\begin{tabularx}{\linewidth}{@{}Xcc@{}}
\toprule
TRIAD subset & Shared systems (n=13) & Without external T-CoT (n=11) \\
\midrule
I2C & 0.582 & 0.727 \\
C2I & 0.775 & 0.809 \\
B-Fashion & 0.934 & 0.991 \\
B-Cars & 0.681 & 0.800 \\
C-Fashion & 0.703 & 0.800 \\
C-Cars & 0.560 & 0.718 \\
\bottomrule
\end{tabularx}
\end{center}
\end{table}

Each column recomputes ranks within its own shared set rather than retaining ranks from the other column. Removing the two external-reasoning configurations raises correlations overall, but the main-text I2C reversal of Rzen against WeMM/Qwen3-VL-Embedding-8B and the single-/multi-turn reversal of Qwen3-VL-Embedding-2B against B-Tokens remain. Correlation magnitudes therefore depend on the evaluated configurations, whereas these pairwise reversals do not depend on including external reasoning.

These results support related overall ability with complementary differences under specific requirements, not universally low correlation or fully independent abilities. We report descriptive correlations for this updated comparison and do not transfer the previous ten-system analysis's confidence intervals, p-values, or significance conclusions to the 13-system set. Explanations involving joint constraints and state updates also draw on the main diagnostics rather than treating rank shifts between two leaderboards as causal evidence.